\documentclass[ijds,nonblindrev]{informs4_hide}

\RequirePackage{tgtermes}
\RequirePackage{newtxtext}
\RequirePackage{newtxmath}
\RequirePackage{bm}
\RequirePackage{endnotes}

\OneAndAHalfSpacedXI

\usepackage{enumitem}
\usepackage{bbding}
\usepackage{pifont} 
\usepackage{booktabs}
\usepackage{subcaption}
\usepackage[hidelinks]{hyperref}
\usepackage{fvextra}
\usepackage{csquotes}
\usepackage{tabularx}
\usepackage{threeparttable}

\usepackage{makecell}
\usepackage{siunitx}
\usepackage[table]{xcolor}

\newcolumntype{B}{>{}c}

\usepackage{natbib}
 \bibpunct[, ]{(}{)}{,}{a}{}{,}
 \def\bibfont{\small}

\EquationsNumberedThrough

\TheoremsNumberedThrough

\ECRepeatTheorems

\MANUSCRIPTNO{IJDS-2025-0124.R1}

\begin{document}

\RUNAUTHOR{Zhang et al.}

\RUNTITLE{Evaluating LLMs as Human Behavior Simulators in Operations Management}

\TITLE{Predicting Effects, Missing Distributions: Evaluating LLMs as Human Behavior Simulators in Operations Management}

\ARTICLEAUTHORS{

\AUTHOR{Runze Zhang, Xiaowei Zhang, Mingyang Zhao}
\AFF{Department of Industrial Engineering and Decision Analytics, Hong Kong University of Science and Technology, Clear Water Bay, Hong Kong SAR, \EMAIL{rzhangdl@connect.ust.hk, xiaowei@ust.hk, mingyang.zhao@connect.ust.hk}}

}

\ABSTRACT{
Large language models (LLMs) are increasingly used to simulate human behavior in business, economics, and the social sciences, offering a low-cost complement to laboratory experiments, field studies, and surveys. This paper evaluates how well LLMs replicate human behavior in operations management. Using nine published behavioral-operations experiments, we assess LLM performance along two dimensions: whether LLM-generated data reproduce the original hypothesis-test outcomes, and whether their full response distributions align with human data, measured by Wasserstein distance. We find that LLMs often replicate hypothesis-level effects, suggesting that they can capture salient decision biases and behavioral regularities. However, their response distributions frequently diverge from human data, even for strong proprietary models, with dispersion mismatch playing an important role. We also examine two lightweight mitigation strategies: chain-of-thought prompting and hyperparameter tuning. Both can reduce distributional misalignment, and appropriate tuning can sometimes allow smaller or open-source models to match or outperform larger proprietary systems.
}


\maketitle

\section{Introduction}\label{sec:Intro}

The design and performance of operations systems depend on how managers, workers, and customers behave and make decisions. Rather than being fully rational, these individuals often display decision biases when they face limited time and cognitive resources in operational settings. Inventory orders, forecasts from historical data, customer purchases in response to queue lengths, contract choices in supply chains, and bids in auctions all reflect behavioral regularities (e.g., mental accounting, satisficing, framing, reference dependence) that shape system outcomes. When these behaviors are misspecified, policies that appear optimal on paper can underperform in practice. Laboratory experiments are therefore a cornerstone of empirical research in behavioral operations management (OM): they allow clean causal identification by tightly controlling information flows, incentives, and the environment \citep{donohue2018handbook}. However, conventional human-subject experiments face two constraints—relatively high costs (recruitment and fixed lab overhead) and limited throughput due to scheduling and monitoring—which restrict experimental scale and the amount of data that can be collected.

Large language models (LLMs), such as ChatGPT, LLaMA, Qwen, and DeepSeek, have advanced rapidly in language understanding, reasoning, and generation. 
Building on this progress, researchers are increasingly using LLMs to run agent-based simulations that model agents' decision processes, communication, and adaptation in controlled environments \citep{GaoLanLiYuanDingZhouXuLi24}. 
This approach enables analysis of large-scale system performance under more realistic behavioral assumptions. 
Applications span economic activities \citep{li2023econagent}, transportation \citep{jin2024surrealdriver}, and epidemic modeling \citep{williams2023epidemic}. 
Within OM, recent studies document that LLM outputs often exhibit human-like decision biases \citep{ChenKirshnerOvchinnikovAndiappanJenkin25}, which suggests, but does not establish, that LLMs might serve as behavioral surrogates, that is, \emph{LLMs capable of mimicking human behavioral patterns} in OM experimental settings.
Given this evidence, we ask: 

\smallskip
\begin{center}
\emph{Can LLMs accurately simulate human behavior in OM?}    
\end{center}
\smallskip

A positive answer would matter for two reasons. First, LLMs could serve as a low-cost surrogate for expensive lab experiments—a major empirical method in behavioral OM—allowing fast theory prototyping and idea testing. Second, beyond replacing lab subjects, LLMs could drive richer simulations of operations systems with boundedly rational individuals, supporting policy design and stress tests under more realistic behavioral assumptions. This stands in contrast to many current system designs that rely on stylized, fully rational agents and can miss important behavioral frictions. 
To avoid presupposing the answer, we go beyond documenting decision biases in LLM responses, as shown by \cite{ChenKirshnerOvchinnikovAndiappanJenkin25}, and conduct task-level validation by comparing LLM behavior with human data across canonical OM problems.

We study this question by testing a range of LLMs, including proprietary (GPT‑3.5, GPT‑4, GPT‑4o mini) and open‑source (LLaMA, Qwen, Deepseek), on open datasets from recent lab experiments across OM domains: inventory management, supply chains, sourcing, queueing and forecasting. We give LLMs the same tasks and instructions as human participants of the lab experiments, collect responses under a standardized protocol, and run parallel analyses that mirror the original studies.

Our answer is both ``Yes'' and ``No''---and the distinction matters. 

Yes for \emph{predicting effects}: 
Consistent with recent findings, LLMs largely reproduce the outcomes of hypothesis tests from a recent large-scale replication study by \cite{DavisFlickerHyndmanKatokKepplerLeiderLongTong23}, often matching both effect direction and statistical significance. For example, LLM responses reflect mental accounting in inventory decisions \citep{doi:10.1287/mnsc.1120.1638} and the positive relationship between the number of bidders and the reservation price \citep{doi:10.1287/mnsc.2015.2264}.

No for \emph{distributional alignment}: Because LLM generation is probabilistic, as models estimate a next‑token probability distribution from learned patterns and then, under stochastic decoding, sample from it, outputs vary across runs, loosely analogous to human heterogeneity. 
When comparing the empirical distributions of LLM and human responses using the Wasserstein distance, a common metric for discrepancies between probability distributions \citep{KuhnEsfahaniNguyenShafieezadeh19}, we find sizable gaps across settings: LLMs often get the sign right but the shape wrong, missing the dispersion, tails, and multi-modality that characterize human decisions.

We then examine how to narrow this gap. 
While fine‑tuning is a standard way to improve alignment for domain-specific applications of pre‑trained LLMs \citep{wei2021finetuned, chung2024scaling}, it typically requires substantial task‑relevant data to preserve transferability—data often scarce in behavioral OM due to the cost of large lab experiments. We therefore test lightweight strategies that require no retraining. Specifically, we use chain-of-thought (CoT) prompting to make reasoning explicit \citep{wei2022chain} and systematically vary hyperparameters, focusing on temperature and configurations of sampling methods such as top-$p$, min-$p$, and top-$k$. Tuning these settings can substantially improve distributional alignment, as measured by the Wasserstein distance, and in some cases tuned open-source LLMs match or even exceed proprietary models with default settings.

\subsection{Main Contributions}

Our contributions are threefold. First, we introduce a reproducible framework for head-to-head evaluation of LLMs as human behavior simulators in behavioral OM. 
We implement lab-style protocols by assigning LLMs the identical tasks and instructions as human subjects and conducting parallel analyses that replicate lab experiments.
We pair hypothesis-testing replication with a distributional assessment that formalizes human–LLM comparisons using the Wasserstein distance.

Second, we present evidence across multiple OM domains that LLMs replicate human behavioral effects but exhibit weak distributional alignment with human responses. This has direct implications for behavioral OM. On the one hand, effect-level (i.e., hypothesis-level) replication supports using LLMs for theory prototyping, hypothesis generation, and as a complement to resource‑intensive lab data collection. 
On the other hand, the observed distributional gaps warrant caution when using LLM outputs as drop‑in replacements for human behavior in simulation‑based evaluation, system design, and operational policy development, where outcomes depend on full distributional characteristics.

Third, we identify two lightweight strategies for mitigating distributional misalignment: CoT prompting and hyperparameter tuning. CoT prompting is easy to implement and requires little additional computational overhead, but its effect is unstable: it improves alignment in some experiments while worsening it in others. Hyperparameter tuning, by contrast, yields more consistent improvements by adjusting the model's sampling behavior through temperature and sampling-method choices, although it requires additional simulation runs and hence greater computation. Importantly, this cost remains substantially lower than model retraining or fine-tuning. For OM researchers and practitioners, these findings suggest a practical path toward higher-fidelity behavioral simulation when large behavioral datasets or substantial computational resources are unavailable.

\subsection{Related Works}

LLMs have been widely studied as substitutes for human participants in agent-based simulation across domains such as transportation \citep{jin2024surrealdriver} and epidemic modeling \citep{williams2023epidemic}; see \cite{GaoLanLiYuanDingZhouXuLi24} for a recent survey. 
In business and economics, a growing body of work reports encouraging similarity between LLM and human responses. 
In economics, \cite{NBERw31122} finds that LLMs qualitatively match results from several classical behavioral experiments \citep{kahneman1986fairness, samuelson1988status,charness2002understanding}, allowing low-cost exploration of new hypotheses. 
In marketing, \cite{LiCasteloKatonaSarvary24} use LLMs for perceptual analysis and report over 75\% agreement with human survey results, while \cite{arora2024revolutionizing} show that LLMs can assist across both qualitative and quantitative stages of the research pipeline, serving as analyst--consultant tools. Beyond these areas, \cite{xie2024can} show that LLMs can reproduce human trust behavior. 
In behavioral OM, \cite{ChenKirshnerOvchinnikovAndiappanJenkin25} evaluate GPT-3.5 and GPT-4 on 18 common human decision biases and find that the LLMs exhibit human biases in about half of the cases.
Our paper advances this line of research by conducting comprehensive experiments across multiple OM domains to assess how accurately LLMs simulate human behavior in OM applications.

Meanwhile, despite the promise of LLMs, several studies show that LLM-generated data can diverge from human data, especially under less informative prompts. For example, \citet{aher2023using} find that larger models often produce unnaturally accurate responses that depart from typical human patterns, and \citet{santurkar2023whose} develop a quantitative framework showing substantial misalignment between LLM-generated opinions and those of human respondents across demographic groups, indicating that current LLMs do not fully capture the diversity of human views. \cite{turing_test_ai} engage GPT-4 in classical behavioral games and report behavior that is largely indistinguishable from humans, although GPT-4 shows higher altruism and cooperation. \cite{doi:10.1287/mksc.2023.0306} show that GPT-3.5 prefers smaller immediate rewards to larger delayed rewards, and that GPT-4 exhibits higher discount rates than humans in intertemporal choice. \cite{yang2024large} further suggest that, without shortcut features, LLMs underperform on social prediction relative to \cite{argyle2023out}. Complementing these findings, \cite{wang2025large} provide evidence from studies with 3,200 participants across 16 demographic identities that LLMs misportray demographic groups and flatten within-group diversity, underscoring the need for mitigation strategies. \cite{Zhang25comparing} focus on the exploration-exploitation trade-off of canonical multi-armed bandit experiments and find that in more complex, non-stationary environments, LLMs struggle to match human adaptability.
Taken together, our paper helps reconcile these lines of evidence: in OM applications, LLMs largely reproduce the behavioral effects observed in lab experiments with human participants, yet fail to match the full distributions of human responses.

There are several ways to align LLMs with human or other criteria, including fine-tuning \citep{wei2021finetuned}, prompt engineering \citep{chung2024scaling}, and in-context learning \citep{wei2022emergent}. Fine-tuning trains pretrained models on new datasets to adapt them to specific tasks. For example, \cite{suh2025language} and \cite{cao2025specializing} fine-tune LLMs to generate group-level survey data and report notable improvements relative to human benchmarks. However, such methods typically require tens of thousands of data points, which are difficult to obtain in behavioral OM experiments. Beyond fine-tuning and prompting, techniques tailored to LLM-based human simulation have also emerged. \cite{wang2025largelanguagemodelsmarket} use data augmentation and transfer learning to debias LLM-generated data with a small amount of human data, while \cite{leng2024reduce} reweight LLM responses by demographic personas to improve distributional alignment.
By contrast, our paper investigates lightweight strategies, specifically adjusting generation hyperparameters such as temperature and sampling, without large datasets or retraining, to mitigate distributional misalignment between LLM and human responses.

\smallskip 

The rest of the paper is organized as follows. Section~\ref{sec:Method} introduces our framework for evaluating LLMs as simulators of human behavior in OM applications. Section~\ref{llm_qualitative} shows that LLMs largely replicate the hypothesis-test outcomes of behavioral OM experiments. Section~\ref{sec:dist_gap} documents a sizable distributional gap between LLM and human responses. Section~\ref{COT}  and Section~\ref{temperature_tuning} examine whether CoT prompting and hyperparameter tuning can reduce this misalignment, respectively. We conclude in Section~\ref{sec:Conclusion} and report additional results in the e-companion.

\section{Methodology}\label{sec:Method}

In this section, we present a framework for evaluating LLMs as simulators of human behavior in OM. First, we describe the dataset, which comes from a recent large-scale replication of behavioral OM experiments. Second, we design LLM simulations that mirror the lab protocols by aligning the tasks, instructions, and response formats. Third, we analyze the outputs using statistical tests to assess whether LLMs reproduce the experiments’ hypothesized effects. We also use the Wasserstein distance to compare the full response distributions with human data.

\subsection{Human Response Data from Lab Experiments}\label{llm_qualitative_hyp}

\emph{Management Science}, a flagship journal in  operations research and management science, recently organized the first large-scale replication study of behavioral OM lab experiments. 
The team selected ten influential experimental papers published in the journal since 2000, spanning domains such as inventory, supply chains, queueing, sourcing and forecasting. 
Each experiment was replicated with high statistical power by multiple independent research groups at different sites, using human participants recruited from different geographic regions. The results were consistent across teams and locations and aligned with the original findings reported over the past two decades, indicating that this dataset is highly representative of human behavior in OM. 
The study results are documented in \cite{DavisFlickerHyndmanKatokKepplerLeiderLongTong23}.

In all of the lab experiments originally designed by the ten OM papers, except for \cite{doi:10.1287/mnsc.2016.2610}\footnote{This experiment tests human reaction speed and requires physically sliding modules to adjust values on a graphical interface, which cannot be simulated by LLMs. We therefore exclude it from our study.}, human participants acted as managers and completed the tasks by making managerial decisions. Their responses, expressed as text and numbers and often collected over multiple rounds of interaction, were recorded.
Each experiment was designed to test a hypothesis about systematic patterns in human decision-making in OM contexts. 
Below, we summarize the hypotheses examined in the nine experimental OM papers (listed in alphabetical order by the first author's last name).
More details can be found in Section~\ref{experimental_settings} of the e-companion.

\begin{enumerate}
\item 
\cite{doi:10.1287/mnsc.1120.1638} study how payment schemes affect inventory order decisions. They find that order quantities under own financing (scheme O) are significantly higher than under customer financing (scheme C). Mental accounting is proposed as a behavioral explanation for this pattern.

\item 
\cite{doi:10.1287/mnsc.1050.0436} examine the bullwhip effect in a four-tier supply chain (retailer, wholesaler, distributor, factory) with ordering and delivery lags. The hypothesis is that sharing dynamic inventory information across the supply chain reduces order oscillations.

\item 
\cite{doi:10.1287/mnsc.1100.1258} investigate how sellers set reservation prices in second-price auctions with $n$ potential buyers, where buyers’ private values are drawn from a cube-root distribution. The main hypothesis is that the seller's chosen reservation price increases with the number of bidders ($n = 1, 4, 7, 10$).

\item 
\cite{doi:10.1287/mnsc.1070.0806} study the role of regret and feedback in first-price sealed-bid auctions. They consider two feedback conditions. Under Loser's Regret, subjects see the winning price and the size of their missed opportunity (resale value minus winning bid, or zero if they won). Under Winner's Regret, subjects see the second-highest bid and how much they left on the table (their bid minus the second-highest bid, or zero if they did not win). The hypothesis is that providing both types of feedback leads to lower average bids than providing only Loser's Regret.

\item 
\cite{doi:10.1287/mnsc.1070.0788} test two-part tariffs as a mechanism to improve supply chain efficiency, varying the framing as either a fixed fee or a quantity discount. The hypothesis is that efficiency is higher under the quantity-discount framing than under the fixed-fee framing.

\item 
\cite{doi:10.1287/mnsc.2015.2264} study purchasing decisions as a function of waiting time when some consumers are informed about product quality and others are not. In a simulated queueing setting, the hypothesis is that uninformed consumers can infer quality from observed waits when there are enough informed consumers: for short waits, uninformed consumers are less likely to purchase in the presence of informed consumers (the ``empty restaurant'' effect), while for longer waits, they are more likely to purchase relative to settings with no informed consumers.

\item 
\cite{doi:10.1287/mnsc.1110.1382} analyze forecasting behavior from time-series data under two types of shocks: temporary and permanent. They hypothesize that forecasters overreact to errors in more stable environments and underreact in less stable ones.

\item 
\cite{doi:10.1287/mnsc.46.3.404.12070} investigate newsvendor order quantities for high- versus low-profit products. 
They hypothesize and find that participants order less than the optimal benchmark for high-profit products and more than optimal for low-profit products.

\item 
\cite{doi:10.1287/mnsc.1110.1334} consider a supply chain in which a supplier chooses capacity and a manufacturer holds private demand forecasts and can send a cheap-talk message. While standard theory predicts uninformative messages and that suppliers should ignore them, the hypothesis is that manufacturers' messages are positively correlated with their private forecasts and that suppliers' capacity choices are positively correlated with the messages they receive.

\end{enumerate}

{
To evaluate LLM performance under different levels of strategic complexity, we classify the selected behavioral OM experiments into two categories: single-agent and multi-agent experiments; see Table~\ref{tab:llms}. In single-agent experiments, participants choose actions to maximize utility based on observed data and learned patterns. For example, in \citet{doi:10.1287/mnsc.1120.1638} and \citet{doi:10.1287/mnsc.46.3.404.12070}, participants select order quantities by inferring demand, either under a fixed distributional assumption or from past data used to form forecasts.
In multi-agent experiments, participants must reason strategically about other agents’ motivations and decisions. In \citet{doi:10.1287/mnsc.1110.1334}, for example, manufacturers decide how much private information to report to suppliers, knowing that higher deliveries increase the chance of meeting downstream demand. Suppliers, in turn, decide how much to trust the reported information when placing their own orders.}

\subsection{LLM Response Data from Simulation}

To evaluate LLMs' ability to simulate human responses, we test a range of widely used {instruction-tuned models}: GPT-3.5, GPT-4, GPT-4o mini, LLaMA, Qwen, and { DeepSeek-V3}. The GPT models are proprietary, whereas the remaining models are open source. { We focus on instruction-tuned models\footnote{{All models evaluated in this study are standard instruction-tuned models. We do not use dedicated reasoning models, such as DeepSeek-R1, OpenAI o1, or Qwen reasoning/thinking variants. When models provide optional reasoning functionalities through an API, these reasoning modes are not invoked in our experiments. This design ensures that comparisons across models are not confounded by model-specific hidden reasoning mechanisms or reasoning-token budgets.}} because their training incorporates human-provided instructions, feedback, or preference data, which may make them better suited than base pretrained models for generating human-like judgments and responses.} To examine scaling effects, we include multiple model sizes. For LLaMA, we use 3B, 8B, and 70B parameter models; for Qwen, we use 7B, 14B, 32B, and 72B parameter models.

We adopt the original lab designs, including sample sizes and data-collection protocols as documented on \url{www.aspredicted.org}.
Guided by these designs, we build a simulation framework that maps all nine behavioral OM experiments into prompts and interaction protocols. 
For each question posed to human subjects, we encode the experimental context into an LLM prompt. 
Mirroring the human data collection, LLM ``participants'' advance through the same multi‑round sequences specified by each lab experiment.

We adapt the setup for single‑agent and multi‑agent designs. 
In single‑agent experiments, when an LLM continues as the same participant across rounds, it retains information from earlier rounds (including prior questions and answers). To keep simulated participants independent, a new simulated participant does not receive information from others' rounds. 
In multi‑agent experiments, multiple LLM agents interact within the same round. Although they represent distinct participants, they can access shared information within that round to support coordination or other interactions. For each experiment, we generate the same number of simulated observations as in the corresponding human study.

For each LLM simulation, we adapt the prompt from the original instructions given to human participants. 
Each prompt had three parts:
\begin{enumerate}[label=(\roman*)]
    \item Role assignment and objective. At the beginning, we assign the model a role, typically a firm's manager, and state that the decision objective as maximizing profit. 
    \item Experiment instructions. We preserve the original instructions as much as possible. 
    When information cannot be shown directly (e.g., figures or tables), we convert it to text. We are careful about what information to include: we use zero-shot prompts to limit bias and avoid cues that might steer LLMs toward specific human behavioral patterns. For example, in the experiment of \cite{doi:10.1287/mnsc.1120.1638}, where human subjects often display mental accounting, we do not endow LLMs with behavioral utility; instead, we ask them to make decisions in terms of standard profit maximization.
    \item Historical and peer information. Because all experiments are multi-round, and some involve interactions among multiple agents, we save and feed back relevant history and any shared information from other agents to the LLM when required by the lab experiment's design.
\end{enumerate}

\begin{figure}
    \FIGURE{
    \includegraphics[width=\textwidth]{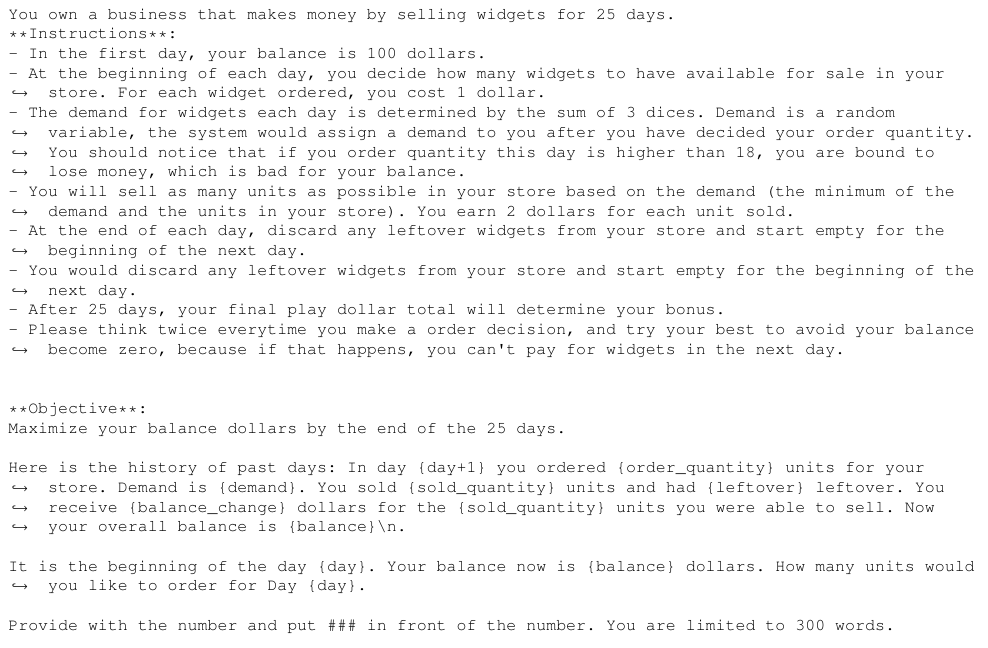}
    }
    {Prompt to Simulate Human Responses in the Experiment of \cite{doi:10.1287/mnsc.1120.1638} \label{fig:prompt-example}}
    {}
\end{figure}

In addition, to speed up generation and simplify data processing, we cap response length and require structured outputs so numerical values are easy to extract. For example, we append: \textit{Provide with the number and put \#\#\# in front of the number. You are limited to 300 words.}
See Figure~\ref{fig:prompt-example} for an example prompt used to simulate human responses in the lab experiment by \cite{doi:10.1287/mnsc.1120.1638}.

\subsection{Data Analysis}
We investigate our central question---``How accurately can LLMs simulate human behavior in OM?''---using two evaluation criteria, one weaker and one stronger.

First, we examine whether LLM-generated responses reproduce the hypothesis-test outcomes obtained from human data. Specifically, we replace human responses with LLM responses in the tests described in Section~\ref{llm_qualitative_hyp} and ask whether the resulting conclusions remain the same. We follow the data-analysis procedures of \citet{DavisFlickerHyndmanKatokKepplerLeiderLongTong23}. For example, in the auction experiment of \citet{doi:10.1287/mnsc.1100.1258}, we replicate the original analysis by regressing the seller's reservation price on the number of bidders $n$, demeaned by its sample average, and conducting Student's $t$-tests for pairwise comparisons across values of $n$ (e.g., $n=1$ versus $n=4$) to assess whether reservation prices increase with the number of bidders. We adopt a stricter replication standard than \citet{DavisFlickerHyndmanKatokKepplerLeiderLongTong23}: to count as a replicated hypothesis-test outcome, the LLM-generated result must be statistically significant in the same direction with a $p$-value below 0.01, whereas their criterion uses $p<0.05$.

Second, we assess distributional alignment between LLM and human responses. This provides a more demanding test of whether LLMs can replicate human behavior, because it compares the full response distribution rather than selected summary statistics or hypothesis-test outcomes. We measure discrepancy using the Wasserstein distance between LLM-generated samples and human responses. 
{ To diagnose the sources of any distributional gap, we also compare the means and variances of the LLM and human data.}

Each lab experiment includes at least two human-replicated datasets, collected asynchronously across sites, with some experiments also involving in-person data. We use these datasets to construct a human benchmark by computing the average Wasserstein distance across pairs of human-generated datasets. When applicable, we further compare Wasserstein distances for conditional distributions. For example, in the supply chain experiment of \citet{doi:10.1287/mnsc.1110.1334}, we analyze reported private information and capacity decisions separately in the LLM-generated data.

\section{Replicating Hypothesis-Test Outcomes of Lab Experiments} \label{llm_qualitative} 

In this section, we evaluate 11 LLMs of varying sizes and versions: {DeepSeek-V3}; GPT‑3.5, GPT‑4, and GPT‑4o mini; {LLaMA 3.2 3B, LLaMA 3.1 8B, LLaMA 3.3 70B}; and Qwen 2.5 (7B, 14B, 32B, 72B) on their ability to replicate hypothesis-test outcomes of the nine lab experiments outlined in Section~\ref{llm_qualitative_hyp}, using a 0.01 significance level as the replication standard.

\begin{table}[ht]
    \TABLE{LLM Replication Results for Lab Experiments \label{tab:llms}}
    {
    \resizebox{\columnwidth}{0.2\textheight}{
    \setlength{\tabcolsep}{1.2mm}{
    \begin{tabular}{lcccccccccccccccccc}
    \toprule
& & \textbf{Deepseek}  & \multicolumn{3}{c}{\textbf{GPT}} & & \multicolumn{3}{c}{\textbf{LLaMA}} & & \multicolumn{4}{c}{\textbf{Qwen}}&  \\
\cline{4-6} \cline{8-10} \cline{12-16}
 Paper/hypothesis & Single/Multi-Agent & &  3.5 & 4 & 4o mini & & 3B & 8B & 70B  & & 7B & 14B & 32B & 72B \\
    \midrule
    \cite{doi:10.1287/mnsc.1120.1638} & S & \Checkmark & \Checkmark & \Checkmark & \ding{55} &  & \ding{55} & \ding{55} & \Checkmark &  & \ding{55} & \Checkmark & \Checkmark & \Checkmark \\
    \cite{doi:10.1287/mnsc.1050.0436} & M & \ding{55} & \Checkmark & \ding{55} & \Checkmark &  & \ding{55} & \ding{55} & \ding{55} &  & \ding{55} & \ding{55} & \ding{55} & \ding{55} \\
    \cite{doi:10.1287/mnsc.1100.1258} & S & \ding{55} & \ding{55} & \Checkmark & \ding{55} &  & \ding{55} & \Checkmark & \Checkmark &  & \ding{55} & \ding{55} & \Checkmark & \Checkmark \\
    \cite{doi:10.1287/mnsc.1070.0806} & S & \Checkmark & \ding{55} & \ding{55} & \ding{55} &  & \ding{55} & \ding{55} & \Checkmark &  & \ding{55} & \Checkmark & \Checkmark & \Checkmark \\
    \cite{doi:10.1287/mnsc.1070.0788} & M & \Checkmark & \ding{55} & \Checkmark & \ding{55} &  & \ding{55} & \Checkmark & \ding{55} &  & \ding{55} & \ding{55} & \Checkmark & \Checkmark \\
    \cite{doi:10.1287/mnsc.2015.2264} & M  &  &  &  &   &   &  &  &  &  &  \\ $\quad$ Hyp KD$_a$ &  & \ding{55} & \Checkmark & \Checkmark & \Checkmark &  & \ding{55} & \ding{55} & \ding{55} &  & \ding{55} & \ding{55} & \ding{55} & \ding{55} \\
   $\quad$ Hyp KD$_b$  &   & \ding{55} & \ding{55} & \ding{55} & \ding{55} &  & \ding{55} & \Checkmark & \ding{55} &  & \Checkmark & \ding{55} & \Checkmark & \Checkmark \\
    \cite{doi:10.1287/mnsc.1110.1382} & S  &  &  &  &   &   &  &  &  &  &  \\ $\quad$ Hyp KR$_a$ &  & \Checkmark & \Checkmark & \Checkmark & \Checkmark &  & \Checkmark & \Checkmark & \Checkmark &  & \Checkmark & \Checkmark & \Checkmark & \Checkmark \\
   $\quad$ Hyp KR$_b$ &   & \Checkmark & \Checkmark & \Checkmark & \Checkmark &  & \Checkmark & \Checkmark & \Checkmark &  & \Checkmark & \Checkmark & \Checkmark & \Checkmark \\
    \cite{doi:10.1287/mnsc.46.3.404.12070} & S  &  &  &  &   &   &  &  &  &  &  \\ 
    $\quad$ Hyp  SC$_a$ &  & \Checkmark & \Checkmark & \Checkmark & \Checkmark &  & \Checkmark & \Checkmark & \Checkmark &  & \Checkmark & \ding{55} & \Checkmark & \Checkmark \\
   $\quad$ Hyp SC$_b$  &  & \Checkmark & \Checkmark & \Checkmark & \Checkmark &  & \ding{55} & \Checkmark & \Checkmark &  & \Checkmark & \Checkmark & \Checkmark & \Checkmark \\
    \cite{doi:10.1287/mnsc.1110.1334} & M  &  &  &  &   &   &  &  &  &  &  \\ $\quad$ Hyp OZ$_a$ &  & \Checkmark & \Checkmark & \Checkmark & \Checkmark &  & \Checkmark & \Checkmark & \Checkmark &  & \Checkmark & \Checkmark & \Checkmark & \Checkmark \\
    $\quad$ Hyp OZ$_b$ &   & \Checkmark & \Checkmark & \Checkmark & \Checkmark &  & \ding{55} & \Checkmark & \Checkmark &  & \Checkmark & \Checkmark & \Checkmark & \Checkmark \\
    \bottomrule    
    \end{tabular}}}}
    {
    \emph{Note.}
    The last four papers each evaluate two hypotheses (e.g., Hyp KD$_a$ and Hyp KD$_b$ in \citet{doi:10.1287/mnsc.2015.2264}). { We classify an LLM as successfully replicating an experiment only if it validates both hypotheses.} A checkmark (\Checkmark) indicates that the LLM result aligns with the original hypothesis-test outcome, whereas a cross (\ding{55}) indicates a failure to replicate. Details of the hypotheses are provided in Section~\ref{experimental_settings} of the e-companion.}
\end{table}

Table~\ref{tab:llms} reports the hypothesis‑level replication results.
Overall, newer versions and larger models are more likely to replicate the experiments than earlier, smaller models. For example, Qwen‑7B failed to replicate 6 of 9 experiments, while Qwen‑72B replicated 7 of 9.  Still, nearly all models fail on at least one study which is the most often the experiment on how waiting time affects consumers' purchase behavior in \cite{doi:10.1287/mnsc.2015.2264} indicating that some tasks remain difficult even as scale increases. 
Among the 11 LLMs, Qwen-72B and Qwen-32B performs best, following with LLaMA-72B and GPT-4, replicating nearly all experiments.

{  
A key factor behind this hypothesis-level replication is instruction tuning. All LLMs evaluated in our study have undergone instruction fine-tuning, such as supervised fine-tuning or reinforcement learning from human feedback. By incorporating human preferences, judgments, and response patterns into the training process, instruction-tuned models are better positioned than base pretrained models to generate responses that reflect human judgment and behavioral regularities.
}

Nevertheless, instruction tuning does not make LLM performance uniform across experimental settings. Some experiments, such as the newsvendor study on how product profits affect order quantities \citep{doi:10.1287/mnsc.46.3.404.12070}, are relatively easy to replicate, with nearly all LLMs succeeding in Table~\ref{tab}. By contrast, other experiments \citep{doi:10.1287/mnsc.1050.0436,doi:10.1287/mnsc.2015.2264} remain difficult even for the strongest models. A consistent pattern is that the easier studies involve a single decision maker, whereas the harder studies involve multiple interacting agents, including the four-agent settings in \citet{doi:10.1287/mnsc.1050.0436} and \citet{doi:10.1287/mnsc.2015.2264}.

\subsection{Single-Agent Experiments}

In single‑agent experiments, LLMs replicate hypothesis‑level behavioral effects but differ numerically across models and relative to humans. On average, they appear more rational than human participants, as indicated by smaller discrepancies  from ``optimal'' decisions: whether profit‑maximizing order quantities or objectively optimal forecasts. Among models, Qwen‑72B and GPT‑4 are closest to human responses.

To simplify comparisons, we focus on two single‑agent experiments: \citet{doi:10.1287/mnsc.1120.1638} and \citet{doi:10.1287/mnsc.1110.1382}.
Each has one participant per round, avoiding interactions among simulated LLMs. 
Their designs are similar: both study decision‑making under uncertainty with known distributional parameters, where agents can optimize by computing expected values. In \citet{doi:10.1287/mnsc.1120.1638}, GPT‑4 aligns most closely with human responses, with the smallest mean difference (0.0395). 

In \citet{doi:10.1287/mnsc.1110.1382}, participants forecast next‑day demand from past time‑series data. Their forecasts are evaluated against fully rational benchmarks to detect over‑ or underreaction in stable and unstable environments, where stability is measured by the variance of temporal shocks relative to permanent shocks. 
To quantify reaction magnitude, the study uses a smoothing factor $\alpha$ with rational benchmarks of $0$ (stable) and $0.94$ (unstable). 
Values above the benchmark indicate overreaction; values below indicate underreaction.

LLMs reproduce the human-like biases: overreaction in stable environments ($\alpha_{\text{stable}} > 0$) and underreaction in unstable environments ($\alpha_{\text{unstable}} < 0.94$)—but differ in degree. 
LLaMA‑70B ($\alpha_{\text{stable}} = 0.358$, $\alpha_{\text{unstable}} = 0.709$) and Qwen‑72B ($\alpha_{\text{stable}} = 0.403$, $\alpha_{\text{unstable}} = 0.515$) react more strongly, whereas DeepSeek ($\alpha_{\text{stable}} = 0.152$, $\alpha_{\text{unstable}} = 0.261$) and GPT‑4 ($\alpha_{\text{stable}} = 0.164$, $\alpha_{\text{unstable}} = 0.331$) react less. All models react less than humans ($\alpha_{\text{stable}} = 0.59$, $\alpha_{\text{unstable}} = 0.89$).

\subsection{Multi-Agent Experiments}

{   

In multi-agent experiments, outcomes depend not only on whether each simulated agent understands the payoff structure, but also on whether it correctly models how other agents' actions convey information. This distinction is important because multi-agent settings introduce an additional layer of strategic inference: an agent's optimal action depends on whether it interprets others' behavior as informative, noisy, or irrelevant. 

Motivated by our finding that multi-agent experiments are less likely to replicate the hypothesized effects, we further examine behavior at the individual-agent level. This decomposition allows us to identify whether replication failures arise because some agents fail to generate the appropriate social signals, because other agents fail to interpret those signals, or both. 
We focus on the queueing experiment of \citet{doi:10.1287/mnsc.2015.2264}, the only study in our sample for which no LLM replicates all of the original hypothesis-test outcomes. The corresponding analyses for the other three multi-agent experiments from \citet{doi:10.1287/mnsc.1050.0436}, \citet{doi:10.1287/mnsc.1070.0788}, and \citet{doi:10.1287/mnsc.1110.1334} are reported in Section~\ref{ec:multi-agent} of the e-companion.

\subsubsection{Experiment of \citet{doi:10.1287/mnsc.2015.2264}}

In this experiment, four consumers sequentially decide whether to purchase a product subject to a delivery delay. The key treatment manipulation concerns the composition of the consumer population. Under the $q00$ condition, no consumers are informed about the product's true value. Under the $q50$ condition, each consumer has a 50\% probability of being informed. Informed consumers observe the true product value before making their purchase decisions, whereas uninformed consumers must infer product quality from observable information, including the waiting time generated by earlier consumers' decisions.

The original hypotheses distinguish short waits from longer waits. When the wait is short, the presence of informed consumers can reduce the purchase probability of uninformed consumers: if informed consumers are present but no queue has formed, the absence of demand is a negative signal about quality. This is the ``empty restaurant'' logic. By contrast, when the wait is longer, the presence of informed consumers can increase the purchase probability of uninformed consumers: a longer queue may indicate that earlier informed consumers found the product valuable enough to buy despite the waiting cost. Therefore, successful replication requires LLM-simulated uninformed consumers to reason about waiting time not only as a direct cost, but also as an endogenous signal generated by the decisions of preceding consumers.

\begin{table}[t]
\TABLE{{Rational Benchmark for Informed Consumers in  \citet{doi:10.1287/mnsc.2015.2264}} \label{tab:informed_rational_decision}}
{{
\begin{tabular}{l c c c c}
\toprule
& \multicolumn{2}{c}{\textbf{Low Value ($V=\$0$)}} & \multicolumn{2}{c}{\textbf{High Value ($V=\$3.50$)}} \\
\cmidrule(lr){2-3} \cmidrule(lr){4-5}
\textbf{Delivery Time ($w$)} & \textbf{Payoff if Buy} & \textbf{Decision} & \textbf{Payoff if Buy} & \textbf{Decision} \\
\midrule
1 week  & \$3.00 & Reject & \$6.50 & Accept \\
2 weeks & \$2.00 & Reject & \$5.50 & Accept \\
3 weeks & \$1.00 & Reject & \$4.50 & Accept \\
4 weeks & \$0.00 & Reject & \$3.50 & Reject \\
\bottomrule
\end{tabular}}
}
{{
\emph{Note.} The payoff from rejecting the order, i.e., keeping the endowment, is \$4.00. The rational benchmark is to accept if and only if the payoff from buying strictly exceeds \$4.00.}
}
\end{table}

We first examine simulated informed consumers. This role is useful diagnostically because informed consumers do not need to make social inferences: they observe the true product value and need only compare the payoff from buying with the payoff from rejecting. Their rational decisions are therefore fixed by the payoff structure, as shown in Table~\ref{tab:informed_rational_decision}. We define the \emph{rational rate} as the fraction of informed-consumer decisions that coincide with this benchmark.

The results show that some models fail at this relatively simple signal-generation task. LLaMA-3B, LLaMA-8B, and Qwen-7B have rational rates of only 0.39, 0.69, and 0.70, respectively. In these cases, informed consumers become noisy signal generators: their purchase decisions no longer reliably reveal whether the product is high value. Consequently, even if uninformed consumers were capable of observational learning, the social information available to them would be distorted. This finding highlights a distinctive source of failure in multi-agent simulations. The experiment may fail to replicate not because the focal uninformed agent lacks the relevant behavioral mechanism, but because other simulated agents fail to produce the behavioral signals required for that mechanism to operate.

Qwen-14B displays a different form of inconsistency. Under the $q00$ condition, its simulated uninformed consumers almost never purchase, regardless of waiting time. Under the $q50$ condition, however, their purchase probabilities become high. This pattern is inconsistent with the first hypothesis, which predicts a lower purchase probability at short waits when informed consumers are present. Thus, even when a model responds to the presence of informed consumers, the direction of its response need not match the intended economic logic.

We next compare the purchase rates of simulated uninformed consumers across waiting times and information conditions. Figure~\ref{LLMs_purchase_rate} reports the average purchase probability under $q00$ and $q50$ for each waiting time. Among the more capable models that generate relatively reliable informed-consumer signals, such as DeepSeek, Qwen-32B, and Qwen-72B, purchase rates decline much less steeply under $q50$ than under the $q00$ baseline when $w \ge 2$. For example, at a wait time of three weeks, Qwen-32B's purchase rate increases from 1.0\% under $q00$ to 51.4\% under $q50$, while DeepSeek's purchase rate increases from 0.7\% to 89.1\%. This divergence suggests that these models partially engage in observational learning. They appear to recognize that, when informed consumers may be present, a longer queue can be an endogenous signal of high product quality rather than merely an exogenous waiting cost.

However, the same models fail to reproduce the short-wait prediction. At $w=1$, their simulated uninformed consumers maintain purchase rates close to 100\% under both $q00$ and $q50$. In other words, they interpret a short wait primarily as a low-cost purchase opportunity, not as a potentially negative signal when informed consumers are present. These models therefore capture the positive social-inference effect at longer waits but miss the negative social-inference effect at short waits. This asymmetry suggests that LLMs may find it easier to reason that ``a long queue is good news'' than to reason that ``no queue is bad news'' in environments where informed agents could have revealed product quality through their choices.

\begin{figure}[t]
\FIGURE{
    \includegraphics[width=\textwidth]{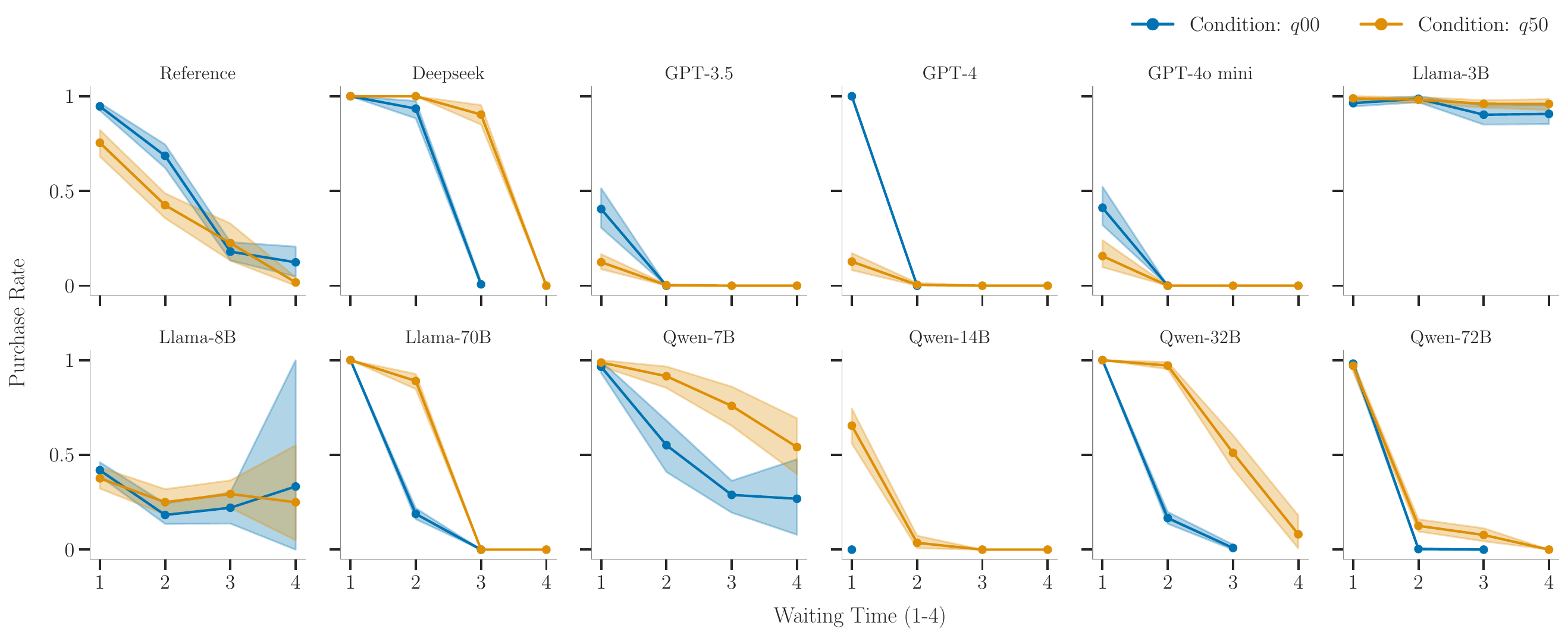}
}
{Uninformed LLM Consumers' Purchase Rates by Waiting Time \label{LLMs_purchase_rate}}
{The LLM simulation is adapted from \citet{doi:10.1287/mnsc.2015.2264}. Each subplot reports uninformed consumers' average purchase rates under $q00$ and $q50$ across four delivery times. Condition $q00$ has no informed consumers; under $q50$, each consumer has a 50\% probability of being informed. Shaded regions show 95\% confidence intervals. { Missing points indicate that no simulated uninformed consumer faced the corresponding waiting time, because waiting time is endogenously determined by earlier purchase decisions; they do not reflect incomplete experiments, processing errors, or invalid LLM outputs.}
}
\end{figure}

A contrasting pattern appears for GPT-3.5, GPT-4, and GPT-4o mini. These models sharply reduce purchase rates once the wait reaches $w \ge 2$, under both $q00$ and $q50$. Their behavior therefore treats waiting time primarily as a direct cost penalty, largely independent of whether informed consumers are present. This response is consistent with a cost-based heuristic but inconsistent with the observational-learning mechanism in the original experiment. At longer waits, these models fail to infer that a queue may be informative about product quality.

Interestingly, the GPT models perform better on the short-wait prediction. When informed consumers are introduced, their purchase rates at $w=1$ fall substantially relative to the $q00$ baseline. For GPT-3.5, the purchase rate decreases from 36.6\% to 11.6\%; for GPT-4o mini, it decreases from 38.1\% to 13.0\%. GPT-4 exhibits an even stronger response, with its purchase rate falling from 100\% under $q00$ to 11.7\% under $q50$. These models therefore reproduce the empty-restaurant logic at short waits, but they fail to sustain the complementary inference that longer queues can be positive quality signals. Put differently, they appear to understand the absence of demand as bad news, but not the presence of demand as sufficiently good news to offset waiting costs.

LLaMA-70B exhibits an intermediate, threshold-like pattern. At $w=2$, it appears to engage in observational learning: its purchase rate increases from 18.6\% under $q00$ to 88.5\% under $q50$. Yet this behavior disappears when the wait becomes moderately longer. At $w \ge 3$, purchase rates fall to nearly zero. This suggests that LLaMA-70B's social inference is fragile. The model recognizes the informational content of a queue when the waiting cost is moderate, but once the wait becomes sufficiently long, the direct cost of waiting dominates its decision rule.

\subsubsection{Common Implications Across the Multi-Agent Experiments}

Across the multi-agent experiments, replication failures rarely arise from a uniform inability of LLMs to understand the task. Instead, they typically reflect role-specific failures that become amplified through interaction. In the queueing experiment of \citet{doi:10.1287/mnsc.2015.2264}, some models fail because informed consumers do not generate reliable quality signals, while others fail because uninformed consumers misinterpret waiting time either as a pure cost or as a uniformly positive signal. In the Beer Game experiment of \cite{doi:10.1287/mnsc.1050.0436}, upstream agents can destabilize the supply chain even when downstream agents respond appropriately to shared inventory information. In the contracting experiment of \cite{doi:10.1287/mnsc.1070.0788}, poor performance can originate from either Retailer-side rejection behavior or Manufacturer-side pricing choices. In the asymmetric-information experiment of \cite{doi:10.1287/mnsc.1110.1334}, LLaMA-3B communicates some information as the Manufacturer but fails to use communicated information as the Supplier. These results show that aggregate treatment effects can mask substantially different behavioral mechanisms.

A common theme is that LLMs are more reliable when a role requires a direct response to payoff-relevant information, and less reliable when the role requires transforming, filtering, or strategically interpreting information generated by other agents. Waiting times, shared inventory signals, nonlinear contract representations, and cheap-talk messages all require agents to map observed actions or messages into latent economic states. Some models perform this mapping only in certain roles or regions of the state space. Others overreact, underreact, ignore the signal, or generate noisy signals that make subsequent inference unreliable.

These findings have two implications for using LLMs as behavioral simulators in operations management. First, multi-agent simulations should be evaluated with role-level diagnostics, not only aggregate hypothesis tests. Researchers should examine whether each agent generates the intended signals, interprets others' signals correctly, and makes decisions consistent with the economic mechanism being studied. Second, successful replication should be interpreted cautiously. An LLM may match a human treatment effect for the wrong reason, or fail to match it because one role corrupts the information environment for others. Multi-agent LLM experiments are therefore most informative when they are paired with mechanism-level checks that identify which agents drive the simulated outcome.

}

\section{Distributional Misalignment} \label{sec:dist_gap}

The hypothesis-level results in Section~\ref{llm_qualitative} show whether LLM simulations reproduce the direction and statistical significance of the original experimental findings. This criterion, however, does not require the simulated response distribution to match the human response distribution. Because LLM outputs are generated by sampling from a learned distribution over tokens, repeated simulations can produce different decisions even under the same prompt. We therefore evaluate distributional alignment directly.

\begin{figure}[t] 
 \FIGURE{
    \includegraphics[width=\textwidth]{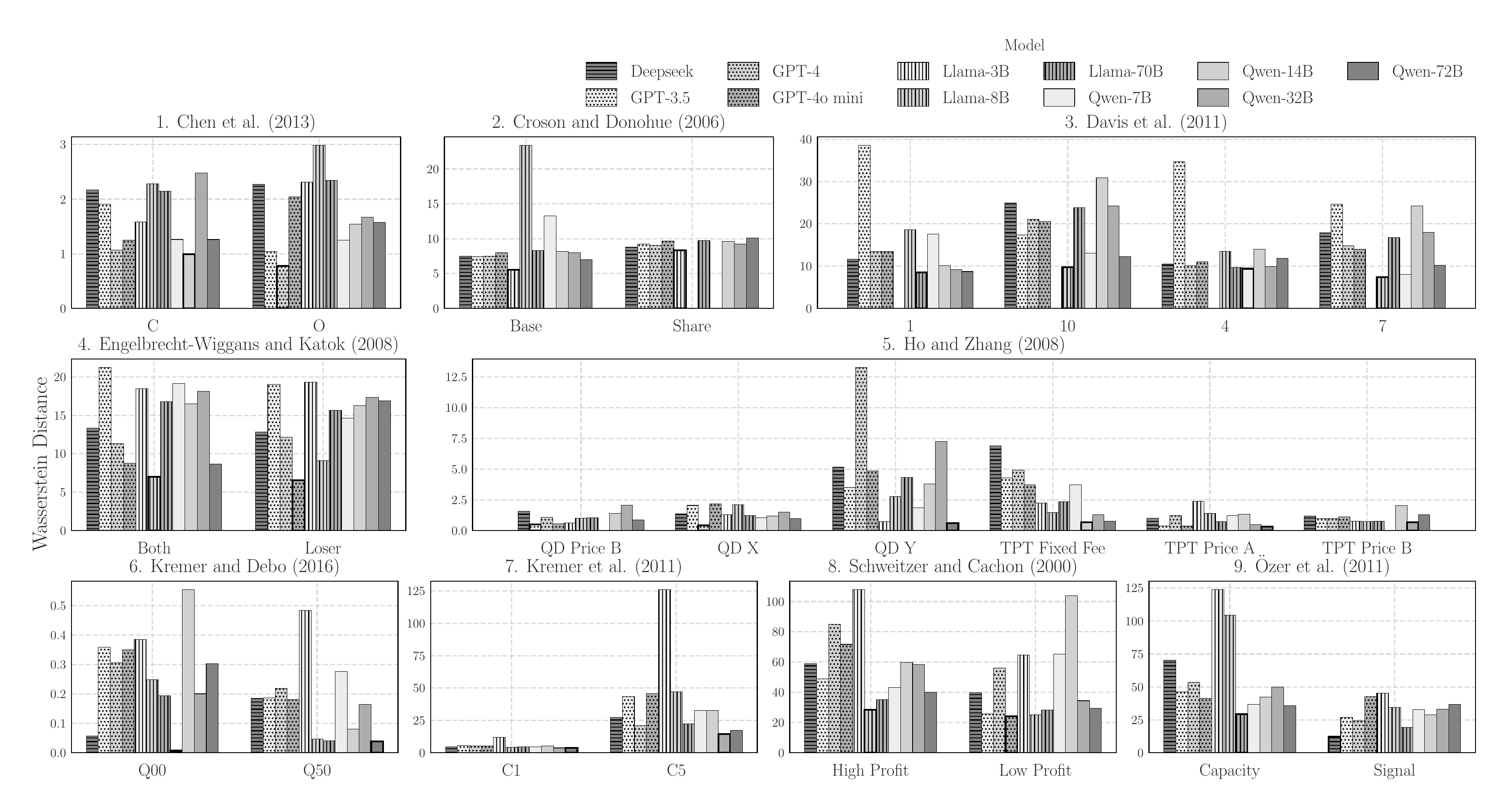}
    }
     {Wasserstein Distance Across LLMs \label{LLMs_wass_ratio}}
     {The x-axis lists the output categories for each experiment, using abbreviated labels for brevity. For example, O and C in the experiment of \citet{doi:10.1287/mnsc.1120.1638} denote the own-financing and customer-financing conditions. Detailed descriptions appear in Section~\ref{experimental_settings} of the e-companion, and numerical results are reported in Section~\ref{subsec:sim-diff-llms}.}
\end{figure}

For each experiment and output category, we compute the Wasserstein distance between the distribution of LLM-generated responses and the corresponding distribution of human responses. Some experiments contain multiple output categories, such as different treatment conditions or different decision variables; the same metric is applied to each category. Figure~\ref{LLMs_wass_ratio} reports the resulting Wasserstein distances across the 11 LLMs.

The results show substantial distributional misalignment across all models. This finding contrasts with the hypothesis-level analysis, where most models successfully replicate many of the original hypothesis-test outcomes. Thus, matching the qualitative treatment effect does not imply that the model reproduces the full distribution of human behavior.

Model scale is associated with better distributional alignment, although the improvement is not uniform. Larger models generally produce lower Wasserstein distances and less variation across simulations. LLaMA-70B and Qwen-72B are among the strongest performers, consistent with their relatively strong performance in the hypothesis-level analysis. Nevertheless, even these models remain visibly separated from human response distributions in several experiments.

Performance also varies by task. In the single-agent experiments on inventory management, auctions, and forecasting, Qwen-72B and GPT-4 perform best overall, but their relative advantages depend on the experimental setting. GPT-4 performs best in the inventory experiment of \citet{doi:10.1287/mnsc.1120.1638}, whereas Qwen-72B performs best in the inventory experiment of \citet{doi:10.1287/mnsc.46.3.404.12070}. Similarly, GPT-4 leads in the auction experiment of \citet{doi:10.1287/mnsc.1100.1258}, while Qwen-72B performs best in the auction experiment of \citet{doi:10.1287/mnsc.1070.0806}. For the demand-forecasting experiment of \citet{doi:10.1287/mnsc.1110.1382}, Qwen-72B most consistently generates responses closest to human judgments.

Multi-agent experiments are more challenging. This is especially true for the four-agent experiments of \citet{doi:10.1287/mnsc.1050.0436} and \citet{doi:10.1287/mnsc.2015.2264}, where all models display sizable distributional gaps and DeepSeek performs best among the tested models. In the simpler two-agent supply-chain experiments of \citet{doi:10.1287/mnsc.1070.0788} and \citet{doi:10.1287/mnsc.1110.1334}, Qwen-72B achieves the strongest overall alignment. These patterns suggest that distributional fidelity depends not only on model capability, but also on the strategic and informational complexity of the experimental environment.

{ 
To diagnose the sources of distributional misalignment, we further examine the gap between LLM-generated and human data via \emph{mean mismatch} and \emph{dispersion mismatch}:
\begin{align*}
\text{Mean Mismatch} ={}& | \mu_{\text{LLM}} - \mu_{\text{human}}|, \\
\text{Dispersion Mismatch} ={}& |\sigma_{\text{LLM}} - \sigma_{\text{human}}|,
\end{align*}
where $\mu$ and $\sigma$ denote the mean and standard deviation of the corresponding data, respectively.

We focus on LLaMA-70B as a representative high-performing model.The analysis shows that dispersion mismatch is the dominant source of error in most experimental conditions. In particular, dispersion accounts for more than 70\% of the Wasserstein distance in the experiments of \citet{doi:10.1287/mnsc.1050.0436}, \citet{doi:10.1287/mnsc.1100.1258}, \citet{doi:10.1287/mnsc.1110.1382}, and \citet{doi:10.1287/mnsc.46.3.404.12070}. By contrast, the experiments of \citet{doi:10.1287/mnsc.1070.0806} and \citet{doi:10.1287/mnsc.2015.2264} are more strongly dominated by bias. The remaining experiments exhibit mixed sources of error across conditions.

}

\begin{figure}[t] 
 \FIGURE{
    \includegraphics[width=\textwidth]{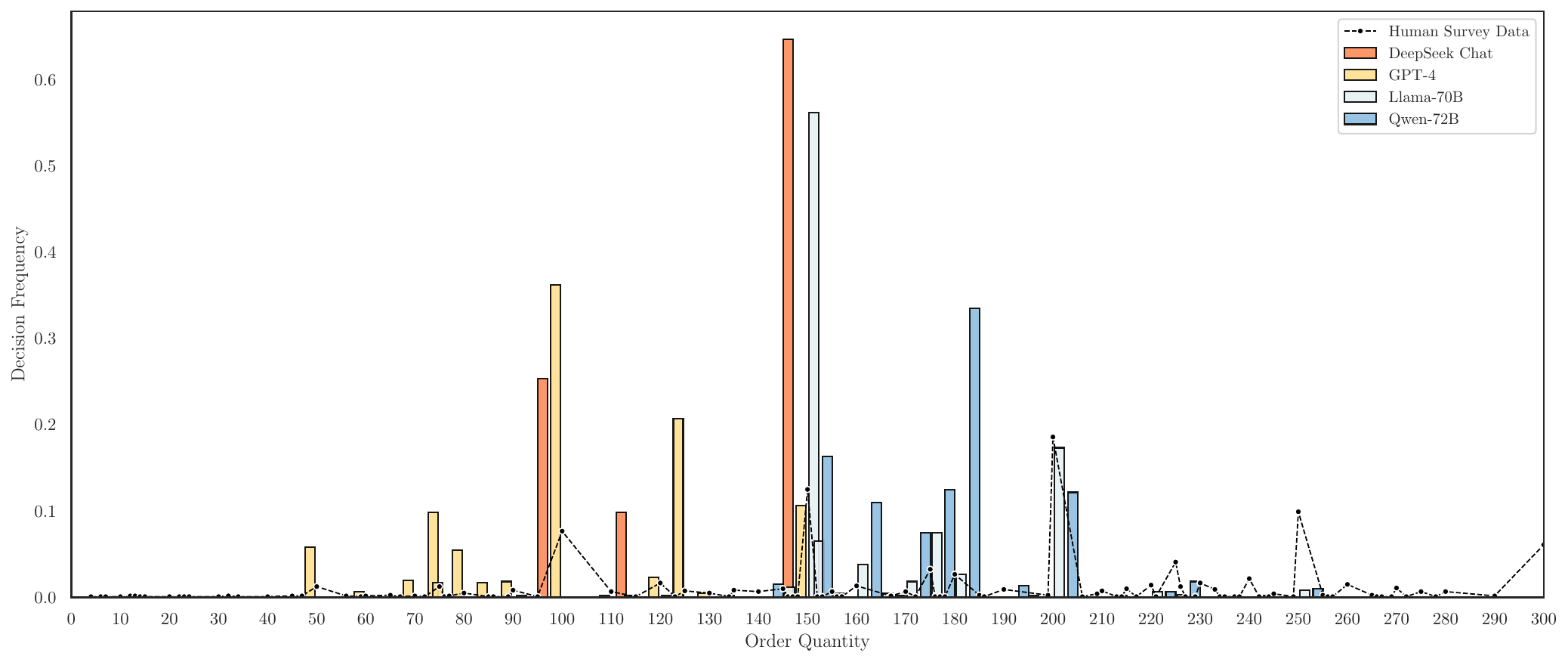}
    }
     {Distributions of LLM and Human Responses in the Experiment of \citet{doi:10.1287/mnsc.46.3.404.12070}\label{LLMs_output_frequency}}
     {The results correspond to the high-profit condition. Line plots show the density of human responses, while bar plots show the empirical distribution of LLM responses.}
\end{figure}

The high-profit condition in \citet{doi:10.1287/mnsc.46.3.404.12070} illustrates the dispersion problem. Figure~\ref{LLMs_output_frequency} compares human responses with responses from four strong models: DeepSeek, Qwen-72B, LLaMA-70B, and GPT-4. Even these models exhibit substantial distributional misalignment. Human responses are spread over a wider range, whereas LLM responses concentrate on a small number of values. In this case, the behavioral gap is driven less by an incorrect average response than by variance collapse: the models generate responses that are too concentrated relative to human behavior.

In short, these results qualify the interpretation of successful hypothesis-level replication. LLMs can often reproduce the direction of an experimental effect while failing to reproduce the heterogeneity of human decisions. For operations-management applications, this distinction matters because many managerial conclusions depend on the full distribution of behavior, not only on average treatment effects. Demand forecasts, order quantities, bids, and capacity choices are all sensitive to behavioral dispersion. Evaluating LLMs as behavioral simulators therefore requires distribution-level diagnostics in addition to hypothesis-level tests.

\section{Chain-of-thought Prompting}\label{COT} 

{
One natural way to reduce the distributional gap between LLM-generated and human behavioral data is to fine-tune the model on task-specific human responses or human-provided instructions. 
In the behavioral OM settings we study, however, fine-tuning is often impractical. First, fine-tuning typically requires a sufficiently large and representative training dataset \citep{zhang2021revisiting}, whereas behavioral OM data are commonly obtained from lab experiments that are costly to design, recruit, and administer. As a result, the available human samples are often too limited to support reliable task-specific adaptation. Second, fine-tuning can require substantial computational resources and model access, which may not be available to many researchers or practitioners, especially when working with large proprietary or open-weight models \citep{hu2022lora}. Third, when the training data are small or narrowly drawn from a specific experimental condition, fine-tuning may overfit to idiosyncratic features of the sample rather than improve general behavioral fidelity \citep{howard-ruder-2018-universal}. These limitations motivate the use of lightweight mitigation approaches that can be applied without retraining the model

In this section, we one such approach:  
CoT prompting,\footnote{{CoT is an external prompt-level intervention. It asks the model to explicitly generate intermediate reasoning text before producing the final decision. This differs from built-in reasoning mechanisms, which are model-specific post-training or inference-time features and may involve hidden reasoning steps not directly controlled by a generic prompt. Thus, our CoT analysis measures the effect of adding an explicit reasoning instruction to standard instruction-tuned models, rather than the effect of activating a model-specific reasoning mode.}} which is a general prompting strategy designed to improve the reasoning ability of LLMs \citep{wei2022chain}. Prior work suggests that CoT can improve performance on reasoning-intensive tasks and reduce certain forms of hallucination \citep{Hallucination-survey23}. 
In our setting, however, the effect of CoT is ex ante ambiguous. Simulating human behavior in OM experiments requires not only logical reasoning, but also behavioral calibration: the model must generate responses that match human choices, including human biases, heuristics, and dispersion. A prompt that induces more explicit reasoning may therefore either improve alignment with human behavior or move the model closer to a normative decision rule that humans do not follow.

We examine two CoT-based prompting approaches: simple CoT and structured CoT. Simple CoT is an unstructured prompting intervention that adds a generic reasoning instruction to the baseline prompt. Structured CoT, analyzed separately, imposes a stage-based reasoning framework tailored to the sequential structure of the experimental task. This distinction is useful because many of our simulations involve repeated or sequential decisions. For example, in the inventory-management experiment of \citet{doi:10.1287/mnsc.1120.1638}, participants make ordering decisions over 25 days. In such settings, performance may depend not only on whether the model reasons before each decision, but also on whether it applies a consistent decision framework over time.}

{

\subsection{Simple CoT}

In the simple CoT condition, we append the instruction ``Explain your reasoning process first'' to the original simulation prompt before the model generates its decision. This intervention is deliberately generic: it does not provide task-specific guidance, identify relevant economic variables, or specify a decision algorithm. It therefore allows us to test whether a low-cost, general-purpose reasoning prompt improves the behavioral fidelity of LLM simulations.

\begin{figure}[t]
    \centering
    \FIGURE{
        \includegraphics[width=0.7\textwidth]{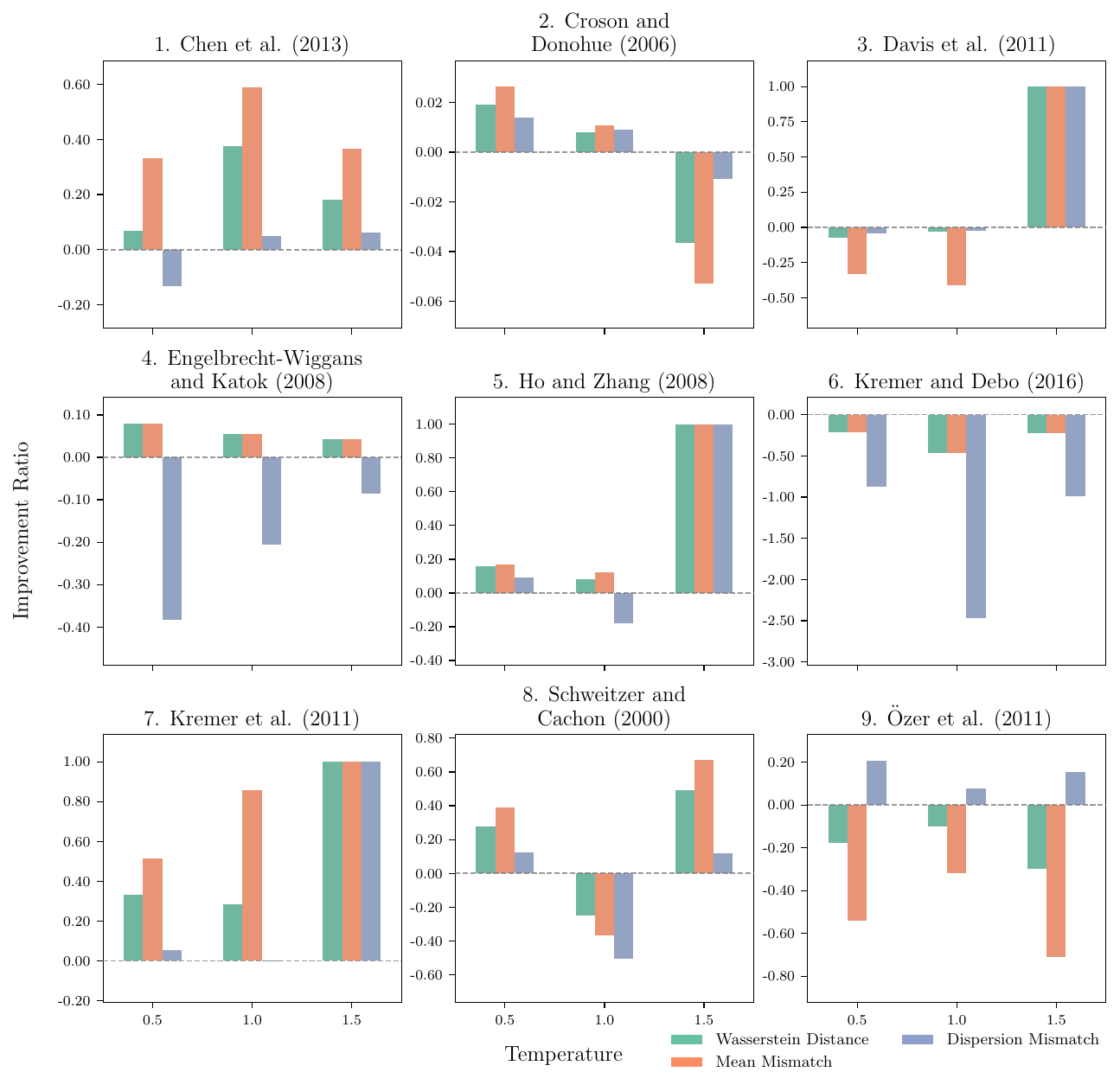}
    }
    {Simple CoT Improvement with LLaMA-70B \label{LLMs_cot_improve}}
    {Improvement ratio is defined as the relative change in Wasserstein distance, mean mismatch, or dispersion mismatch between LLM and human responses after CoT prompting. Positive values indicate improvement, while negative values indicate deterioration.}
\end{figure}

Figure~\ref{LLMs_cot_improve} reports the effect of simple CoT for LLaMA-70B across the nine lab experiments and a range of \emph{temperature}\footnote{The temperature hyperparameter controls output dispersion. Higher values produce more variable generations.} settings. Simple CoT often improves performance, as measured by a reduction in the Wasserstein distance between simulated and human response distributions. In some cases, the improvement ratio approaches 100\%.\footnote{Near-100\% improvements typically occur at high temperatures, where the no-CoT baseline is more prone to extreme outputs and therefore has a very large Wasserstein distance from the human distribution.} However, the effect is far from uniform. CoT worsens performance in some experiments, including those based on \citet{doi:10.1287/mnsc.2015.2264} and \citet{doi:10.1287/mnsc.1110.1382}. These results suggest that simple CoT is not a universally beneficial correction. Although it is easy to implement, its effect should be evaluated empirically rather than assumed to improve behavioral alignment.

The decomposition of distributional mismatch provides additional insight. 
As discussed in Section~\ref{sec:dist_gap}, variance (i.e., dispersion) mismatch is the dominant source of error in many experiments. 
Simple CoT, however, primarily affects mean mismatch. In Figure~\ref{LLMs_cot_improve}, changes in Wasserstein distance closely track changes in bias: when CoT reduces bias, overall mismatch tends to fall; when CoT shifts the mean in the wrong direction, performance deteriorates. By contrast, improvements in dispersion mismatch are limited and appear only in a subset of experiments. Thus, simple CoT appears more effective at changing the central tendency of model responses than at reproducing the dispersion of human behavior.

\begin{figure}[t]
    \centering
    \FIGURE{
        \includegraphics[width=0.5\textwidth]{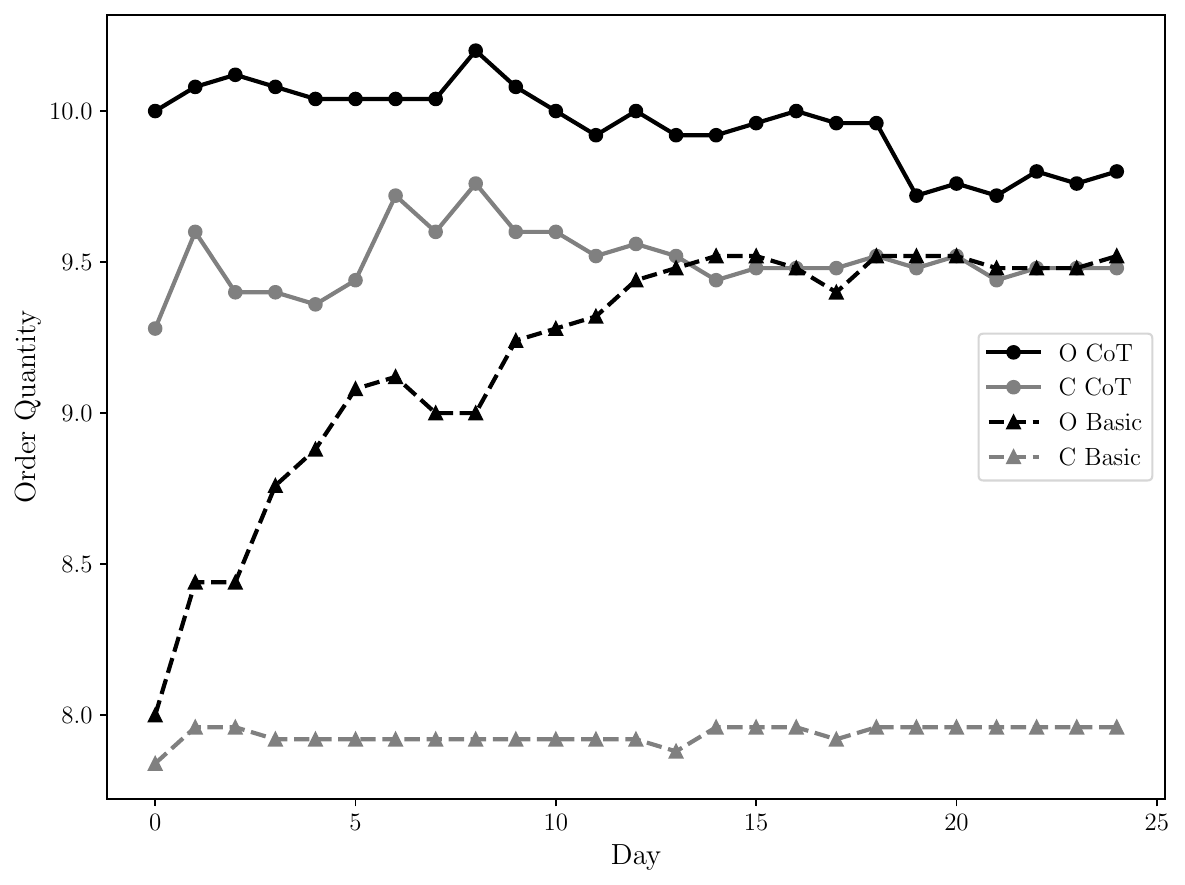}
    }
    {Performance of Simple CoT in the Experiment of \citet{doi:10.1287/mnsc.1120.1638} \label{LLMs_change_r1}}
    {Circle markers denote the average order quantity for simulated participants on a given day with CoT prompting; triangle markers denote the baseline condition without CoT.}
\end{figure}

The inventory-management experiment of \citet{doi:10.1287/mnsc.1120.1638} illustrates how simple CoT changes model behavior. As shown in Figure~\ref{LLMs_change_r1}, both the baseline and CoT conditions preserve the qualitative treatment effect: average order quantities are higher under the newsvendor's own-financing scheme (scheme O) than under the customer-financing scheme (scheme C). CoT nevertheless shifts the level of decisions closer to the human data. Under scheme O, the average order quantity increases from 9.197 without CoT to 9.957 with CoT. Under scheme C, it increases from 7.934 to 9.505. The reasoning traces also become more explicit and more closely tied to the stated trade-offs in the task.

Without CoT, the model often relies on a simple history-driven heuristic, especially under scheme O, adjusting orders upward across rounds in response to recent outcomes. With CoT, the model more consistently articulates the financing condition, demand uncertainty, and payoff consequences before choosing an order quantity. This pattern helps explain why CoT reduces mean mismatch in this experiment. At the same time, the broader results show that such improvements are context-dependent. Simple CoT can make decisions more deliberate, but deliberation does not necessarily make simulated behavior more human-like when the human benchmark reflects framing effects, noisy responses, or other behavioral regularities that explicit reasoning may attenuate.

Additional results on simple CoT are reported in Section~\ref{appendix_cot} of the e-companion.

}

{

\subsection{Structured CoT}

To better understand how CoT affects simulation performance, we implement a structured stage-ablation analysis. The goal is not only to evaluate whether CoT improves alignment with human behavior, but also to identify which components of the reasoning process contribute to the final decision. We decompose each structured reasoning trace into a set of semantic stages,
\begin{equation*}
    R = \{S_1, S_2, \dots, S_n\},
\end{equation*}
where each stage corresponds to a distinct analytical component, such as cost-benefit reasoning, demand analysis, or lead-time analysis.

For each stage \(S_i\), we construct a counterfactual reasoning trace
\begin{equation*}
    R_{-i} = \{S_1, \dots, S_{i-1}, \phi(S_i), S_{i+1}, \dots, S_n\},
\end{equation*}
where \(\phi(S_i)=S_i^{cf}\) replaces the original stage with a counterfactual instruction that blocks the information or reasoning associated with that stage. For example, when ablating a demand-analysis stage, we replace the corresponding reasoning with a statement such as ``Demand and forecast information from this stage are unknown.'' We then ask the model to regenerate its final decision using the modified reasoning trace and measure how the decision distribution changes relative to the empirical human benchmark.

\begin{figure}[t] 
    \FIGURE{
    \includegraphics[width=0.85\textwidth]{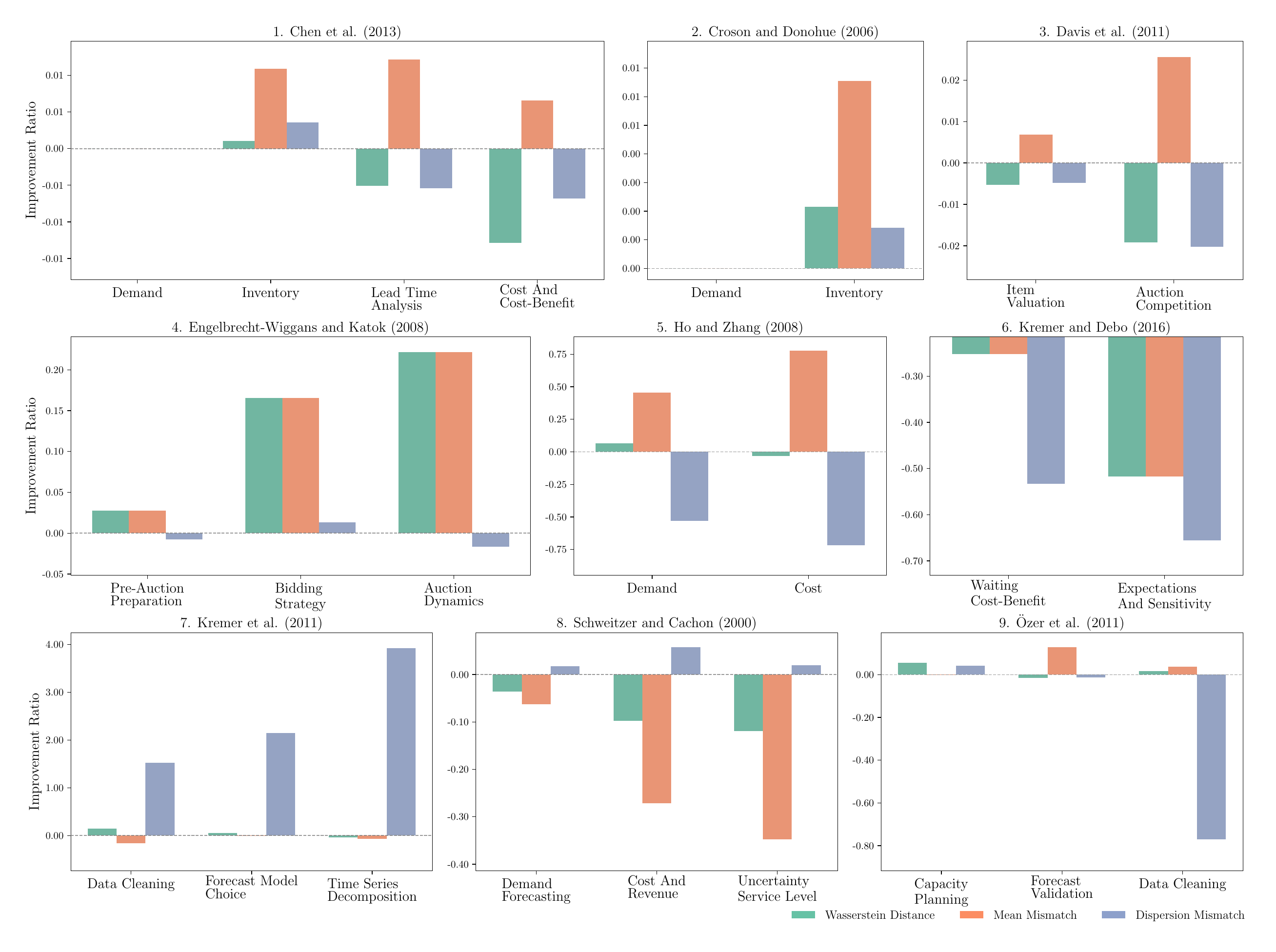}}     
    {{ Effects of Stage Ablation}  \label{LLMs_cot_stage_effect}}
    {{ Improvement ratio is defined as the relative change in Wasserstein distance, mean mismatch, or dispersion mismatch when a reasoning stage is retained rather than removed. Positive values indicate that retaining the stage improves alignment with human responses, whereas negative values indicate that retaining the stage worsens alignment.}}
\end{figure}

Figure~\ref{LLMs_cot_stage_effect} summarizes the stage-ablation results. To make the results interpretable, we group substantively similar stages into broader stage families and omit stage families with fewer than five observations. The results show that individual reasoning stages often have a larger effect on alignment with human behavior than on the mean decision itself. Thus, the stage-ablation analysis provides a mechanism-level account of why CoT helps in some experiments but not in others.

For experiments in which CoT improves alignment overall, including \citet{doi:10.1287/mnsc.1120.1638}, \citet{doi:10.1287/mnsc.1050.0436}, \citet{doi:10.1287/mnsc.1070.0806}, \citet{doi:10.1287/mnsc.1070.0788}, and \citet{doi:10.1287/mnsc.1110.1334}, the Wasserstein panel shows that many isolated stage families either worsen alignment or do not clearly improve it on their own. This pattern suggests that CoT is beneficial in these experiments mainly because the full reasoning chain coordinates multiple analytical components, rather than because each individual stage is independently alignment-enhancing.

By contrast, for experiments in which CoT does not improve alignment overall---including \citet{doi:10.1287/mnsc.1100.1258}, \citet{doi:10.1287/mnsc.2015.2264}, \citet{doi:10.1287/mnsc.1110.1382}, and \citet{doi:10.1287/mnsc.46.3.404.12070}---some visible stage families have positive effects. In these cases, certain reasoning components are locally helpful, even though the complete CoT prompt does not improve the final distribution. The failure of CoT therefore does not imply that all reasoning stages are harmful. Rather, helpful local effects may be offset by other parts of the reasoning chain that move the model away from human behavior.

The most influential stages also differ across experiments. In settings where CoT improves alignment, important stages are mainly operational or strategic: cost and cost-benefit reasoning in \citet{doi:10.1287/mnsc.1120.1638}, auction dynamics in \citet{doi:10.1287/mnsc.1070.0806}, demand and cost reasoning in \citet{doi:10.1287/mnsc.1070.0788}, and data cleaning or time-series decomposition in \citet{doi:10.1287/mnsc.1110.1334}. In settings where CoT does not improve alignment, influential stages include auction competition in \citet{doi:10.1287/mnsc.1100.1258}, expectation and sensitivity analysis together with waiting-cost reasoning in \citet{doi:10.1287/mnsc.2015.2264}, model choice and time-series decomposition in \citet{doi:10.1287/mnsc.1110.1382}, and demand forecasting, cost-revenue reasoning, or uncertainty-service-level reasoning in \citet{doi:10.1287/mnsc.46.3.404.12070}. These patterns indicate that structured CoT changes LLM behavior through economically meaningful reasoning channels, but the direction of the effect depends on how those channels interact within the complete reasoning trace.

Overall, the structured CoT results reinforce the main lesson from the simple CoT analysis. CoT is not merely a generic accuracy booster. It reshapes the model's internal decision process by emphasizing particular analytical stages. This can improve behavioral alignment when the stages jointly reconstruct the human decision logic, but it can reduce alignment when the reasoning chain overemphasizes normative optimization, suppresses human-like heuristics, or combines locally useful stages in a way that distorts the final decision distribution.

}

{

\subsection{Semantic Analysis of CoT Reasoning Traces}

The preceding results show that CoT prompting can improve distributional alignment in some experiments but worsen it in others. To understand why, we analyze the reasoning traces produced by LLaMA-70B and ask which semantic patterns in the traces are associated with smaller or larger distributional gaps. This analysis treats CoT not only as an intervention, but also as a source of diagnostic information about how the model makes sequential decisions. 

We focus on the structured CoT. 
and use an automated LLM-as-a-Judge pipeline to score the textual justifications that precede each simulated decision. The judge is implemented using Qwen2.5-32B-Instruct and assigns continuous scores from 0 to 1 to each round-level reasoning trace. We focus on three semantic heuristics:
\begin{enumerate}[leftmargin=*, noitemsep, topsep=2pt]
    \item \textbf{Historical Signal Averaging:} the extent to which the reasoning uses past observed signals to form an average-like reference for the current decision.
    \item \textbf{Outcome Feedback Adjustment:} the extent to which the reasoning adjusts the current decision based on outcomes from earlier rounds.
    \item \textbf{Constant Action Freezing:} the extent to which the reasoning justifies repeating the same action across rounds because of uncertainty, caution, or environmental volatility.
\end{enumerate}

For each simulated subject in each condition, we aggregate the round-level semantic scores across the full decision sequence. We then relate these subject-level semantic traits to subject-level measures of distributional misalignment, including Wasserstein distance, mean mismatch, and dispersion mismatch relative to the corresponding human benchmark. This aggregation is necessary because mean and dispersion mismatch are properties of an entire multi-round decision sequence rather than of a single round.

\begin{figure}[t]
 \FIGURE{
    \includegraphics[width=\textwidth]{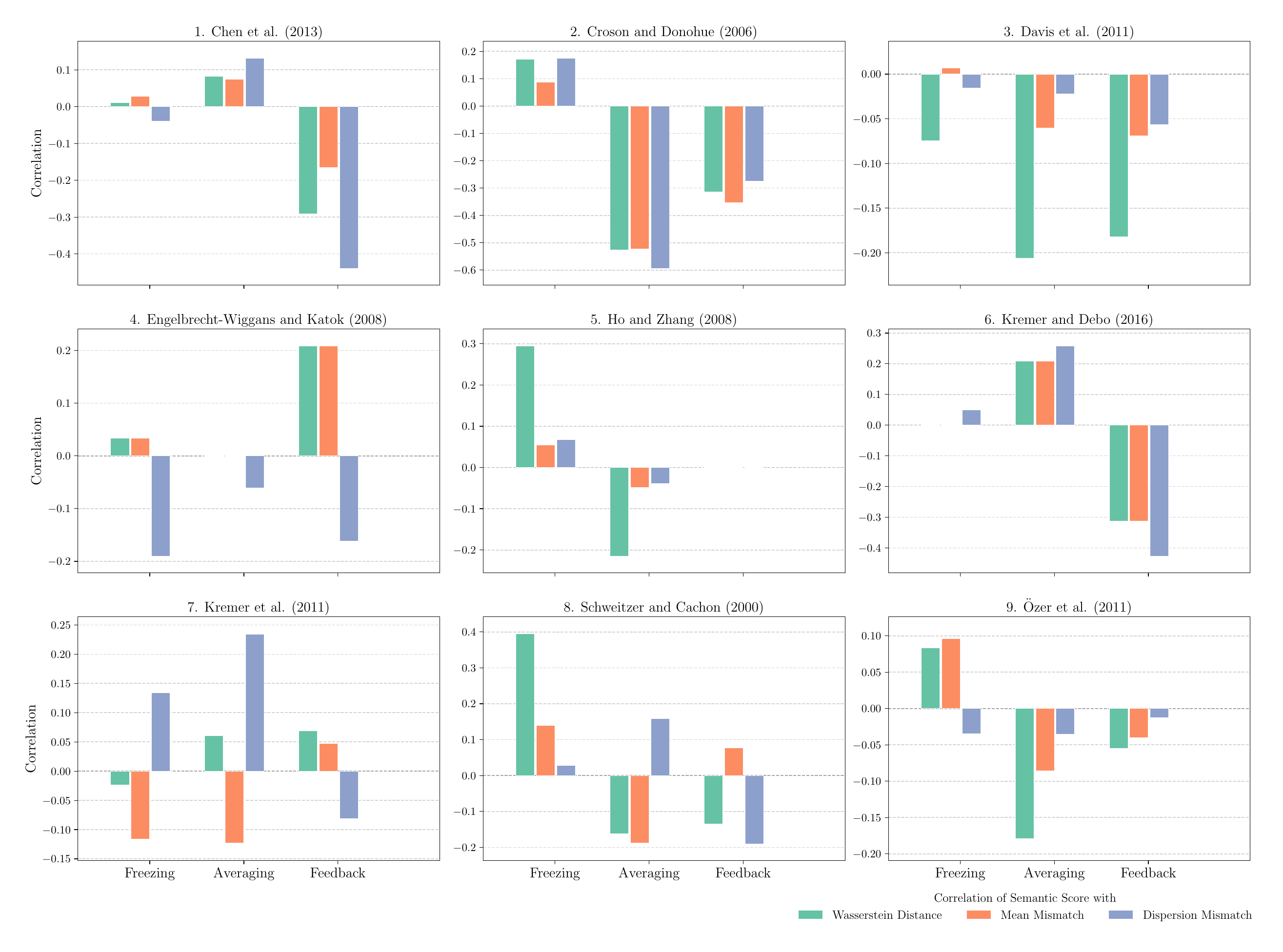}
    }
     {{ Average Correlation Between Semantic Heuristics and Distributional Misalignment Metrics} 
    \label{LLMs_semantic_bias_disp_correlation_facets}}
     {{  The figure reports, for each experiment, the average within-condition correlation between semantic heuristic scores---freezing, averaging, and feedback adjustment---and three misalignment metrics: Wasserstein distance, mean mismatch, and dispersion mismatch. The plotted values are averaged across conditions within each experiment.}}
\end{figure}

Figure~\ref{LLMs_semantic_bias_disp_correlation_facets} shows that \emph{constant action freezing} is the semantic pattern most consistently associated with poor distributional alignment. In many experiments, freezing is positively correlated with Wasserstein distance and, more specifically, with dispersion mismatch. This pattern is especially pronounced in dynamic inventory, auction, and capacity decisions under uncertainty, such as \citet{doi:10.1287/mnsc.1110.1334} and \citet{doi:10.1287/mnsc.1100.1258}. In these settings, the LLM often responds to uncertainty by locking onto a stable quantity rather than continuing to adjust its decisions over time. As a result, the simulated trajectories become too smooth and too concentrated relative to human behavior.

This result helps explain why CoT does not always improve simulation performance. Human subjects often exhibit boundedly rational, pattern-seeking behavior: they respond to recent shocks, adjust actions after unfavorable outcomes, and sometimes chase demand or prices even when such adjustments are not normatively optimal. These behaviors generate substantial within-subject variation over time. By contrast, CoT can encourage the LLM to produce more stable and internally consistent reasoning. Such consistency may improve the mean decision level, but it can also suppress the sequential variability that characterizes human data. In this sense, CoT may reduce first-moment error while worsening, or failing to correct, second-moment error.

Freezing can also contribute to mean mismatch. When the model commits early to a nearly constant decision rule, it may anchor on a suboptimal level and stop updating. This produces persistent deviations from the human mean. Consistent with this mechanism, freezing is positively associated with mean mismatch in several experiments, including \citet{doi:10.1287/mnsc.1070.0788}, \citet{doi:10.1287/mnsc.46.3.404.12070}, and \citet{doi:10.1287/mnsc.1050.0436}. Thus, the same semantic tendency that reduces variability can also prevent the model from correcting an initially miscalibrated decision level.

By contrast, historical averaging and outcome feedback adjustment are generally associated with better alignment. Averaging tends to reduce mean mismatch by recentering decisions around observed signals, while feedback adjustment helps the model correct prior-period errors. These heuristics act as sequential stabilizers: they allow the model to update decisions without relying entirely on a fixed policy. However, their benefits are incomplete. Although averaging and feedback can improve mean alignment and introduce some round-to-round movement, they do not fully reproduce the magnitude or irregularity of human sequential variation.

The semantic evidence suggests a reason why CoT has mixed effects. CoT can help the model articulate relevant information, average signals, and correct prior errors, which often improves mean alignment. At the same time, CoT can also induce overly disciplined and static reasoning, especially when the model treats uncertainty as a reason to freeze rather than explore or adjust. The resulting behavior differs from human bounded rationality: humans often display dynamic suboptimality through noisy adjustment, whereas the LLM frequently displays static suboptimality through excessive consistency. This difference in sequential variability is a central reason why CoT alone cannot eliminate the distributional gap between LLM-generated and human behavioral data.

}

\section{Hyperparameter Tuning}\label{temperature_tuning}

{
In this section, we examine a second lightweight strategy for improving distributional alignment: hyperparameter tuning. Unlike fine-tuning, hyperparameter tuning does not modify the architecture or weights of the underlying neural network. Instead, it changes how the model samples from its learned output distribution. This makes it a practical intervention for behavioral simulation, especially when task-specific human data are limited or model retraining is infeasible. 
In Section~\ref{intervention_process} of the e-companion,
we provide a systematic framework for evaluating and improving LLMs as human-behavior simulators, including the two lightweight strategies discussed in Section~\ref{COT}  and this section.
}

Beyond temperature, an LLM's output distribution also depends on the sampling rule used during generation. These sampling hyperparameters can substantially affect the dispersion, extremity, and stability of generated responses, even though the model itself remains unchanged. We focus on three common sampling methods:

\begin{enumerate}[label=(\roman*)]
\item \textbf{Top-$p$ sampling} \citep{holtzman2020curiouscaseneuraltext} ranks tokens by probability and retains the smallest set of tokens whose cumulative probability exceeds a threshold $p$. Larger values of $p$ allow a broader set of possible tokens, while $p=1$ imposes no truncation.

\item \textbf{Min-$p$ sampling} \citep{nguyen2024turningheatminpsampling} retains tokens whose probability exceeds a specified fraction of the most likely token's probability. Higher min-$p$ thresholds remove more low-probability tokens and therefore make the sampling distribution more selective.

\item \textbf{Top-$k$ sampling} restricts generation to the $k$ most probable tokens at each step. Smaller values of $k$ produce more concentrated outputs, whereas larger values allow greater diversity.
\end{enumerate}

\subsection{Effect of Temperature Tuning} \label{sec:temp-tuning}

We consider nine temperature values: $0.5$, $1.0$, $1.5$, $1.7$, $1.9$, $2.5$, $3.0$, $3.5$, and $4.0$. For each temperature, we vary one sampling method at a time across five threshold settings. For example, when varying top-$p$, we keep min-$p$ and top-$k$ at their default values. This design gives 15 specifications per temperature, based on three sampling methods and five thresholds for each method. We also include a default setting with no sampling constraints. This gives 16 specifications per temperature and 144 hyperparameter combinations in total. We apply all nine studies to each combination, resulting in more than four million LLM interactions. To keep the computation tractable while preserving response quality, we use Qwen-32B.

{
High temperature settings often lead to hallucinations, making single-pass outputs unreliable for decision-making. To address this issue, we allow up to three generation attempts per LLM instruction. If all three attempts fail to produce usable data, we mark that simulation path as incomplete and exclude it from comparisons. 
Consequently, the LLM responses used for LLM-human comparisons are conditioned on successful, non-hallucinated generation.
}
Temperature may also affect hypothesis-level replication, that is, whether the LLM reproduces the hypothesis-test outcomes of the lab experiments.

\begin{figure}[t] 
 \FIGURE{
     \includegraphics[width=\textwidth]{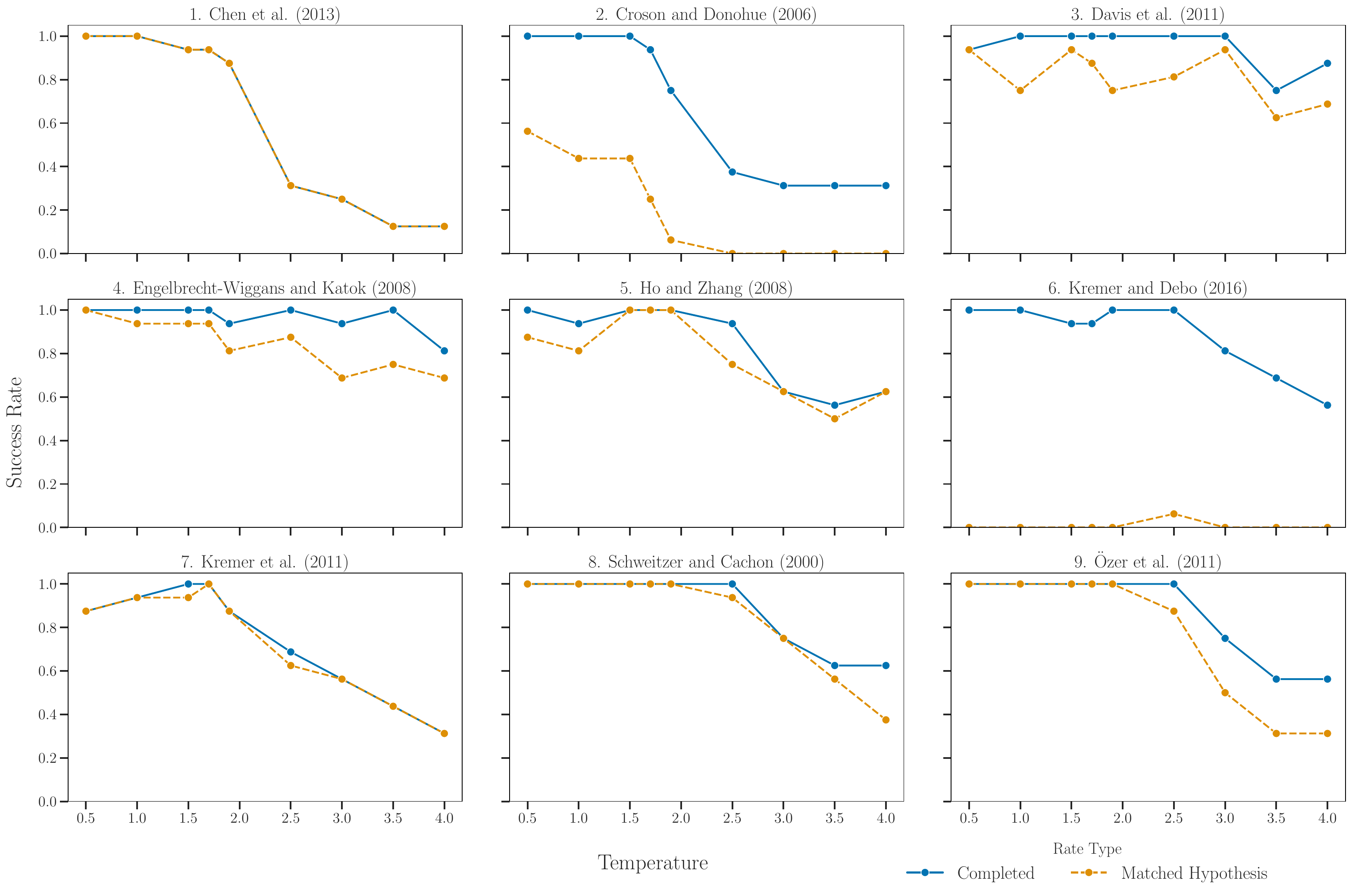}     
     }
     {Completion Rate and Hypothesis-Level Replication Rate by Temperature \label{temperature_success_rate}}
     {The blue line shows the completion rate: the percentage of runs in which the LLM returns valid numerical outputs for all required conditions of an experiment. The orange line shows the hypothesis-level replication rate: the percentage of runs whose results match the study's hypothesis-test outcomes.}

    \end{figure}

Figure \ref{temperature_success_rate} reports both completion rates and hypothesis-level replication rates across the nine lab experiments and different temperatures. 
For most completed simulations, except for three supply-chain studies \citep{doi:10.1287/mnsc.1050.0436, doi:10.1287/mnsc.1110.1334, doi:10.1287/mnsc.1070.0806}, the hypothesis-level replication rate is similar across temperatures. 
In some cases, higher temperatures even improve alignment with hypothesis-level outcomes. For instance, in the experiment of \cite{doi:10.1287/mnsc.2015.2264}, LLMs struggle to match the hypothesis-test outcomes at the default temperature, but increasing the temperature produces samples consistent with the reported outcomes.

\begin{figure}[t] 
  \FIGURE{
    \includegraphics[width=0.9\textwidth]{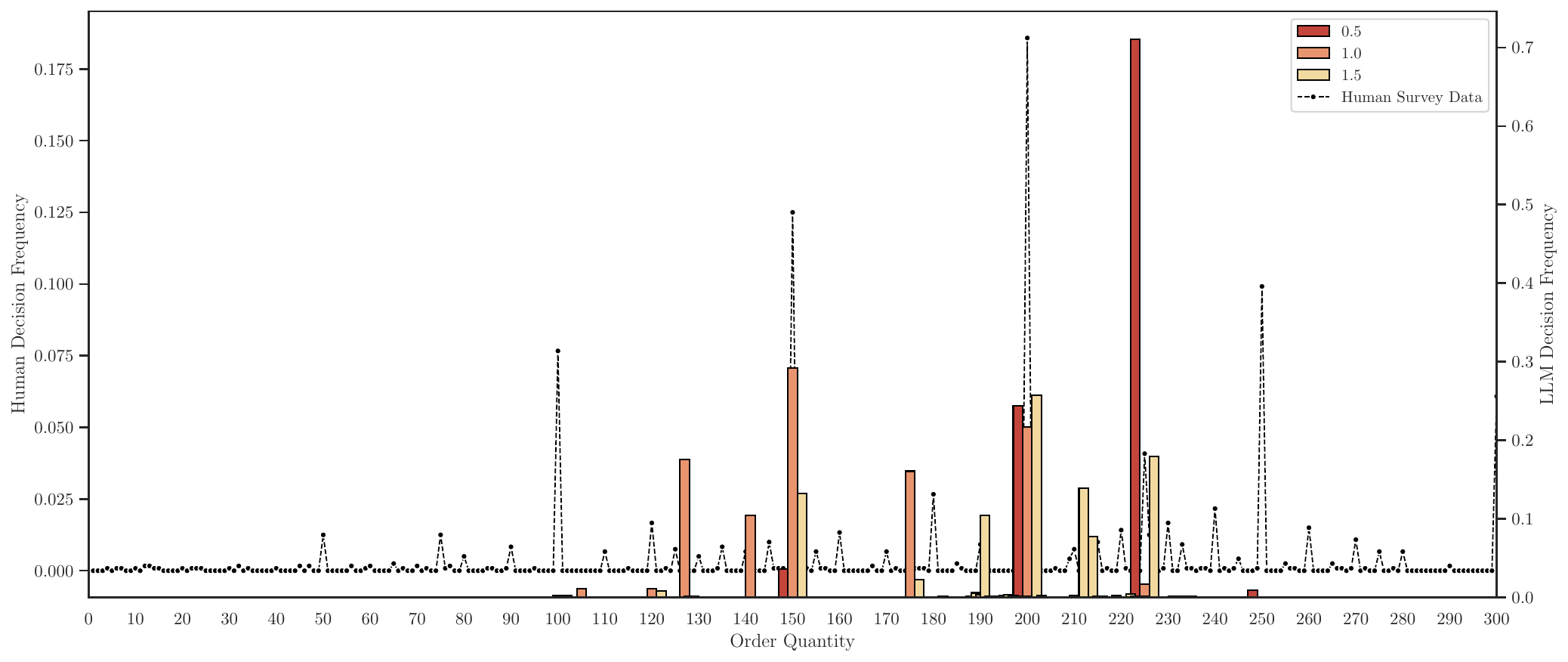}
     }
    {Temperature's Effect on LLM Responses in the Experiment of \cite{doi:10.1287/mnsc.46.3.404.12070} \label{distribution_difference} }
    {The results correspond to the ``high profit'' condition of the experiment. 
    The line plots show the probability density of human responses, while the bar plots show the probability distribution of LLM responses.}
    \end{figure}

Figure~\ref{distribution_difference} shows how the LLM response distribution changes with temperature using the experiment of \cite{doi:10.1287/mnsc.46.3.404.12070} as an example.\footnote{As shown in Figure~\ref{temperature_success_rate}, completion rates often drop sharply once the temperature exceeds 1.5, driven by more frequent hallucinations. We therefore focus on temperatures 0.5, 1.0, and 1.5 in Figure~\ref{distribution_difference}.} 
LLM outputs tend to cluster on a limited set of options, whereas human responses are much more dispersed, covering a wider range with few gaps. 
As temperature increases, the LLM outputs become more random and variable, which broadens the response distribution and reduces the distributional gap between LLM and human responses.

{ This pattern can be explained by entropy-based sampling. At low temperatures, the model relies more heavily on its learned probability distribution. It often collapses toward the most likely responses and misses the diversity found in human decision-making. As temperature increases, the entropy of the output probability distribution also increases. Higher sampling entropy adds more stochasticity and allows the model to select lower-probability tokens more often. This produces greater behavioral variation, which helps the model approximate the dispersion and multi-modality observed among human participants. As a result, the Wasserstein distance decreases. }

{
\subsection{Tuning via Bayesian Optimization}

Tuning multiple hyperparameters to improve distributional alignment is computationally costly. In Section~\ref{sec:temp-tuning}, for example, we evaluate 144 hyperparameter combinations. This is far smaller than a full grid over the multi-dimensional hyperparameter space. Even so, it requires more than 1,000 GPU hours on twenty-eight RTX 4090D GPUs.

To reduce this cost, we use Bayesian optimization (BO), a standard sequential sampling method for hyperparameter tuning \citep{ShahriariSwerskyWangAdamsdeFreitas16}. 
BO builds a surrogate model, typically a Gaussian process, to approximate the relationship between the hyperparameter combination and the optimization objective. Here, the objective is the Wasserstein distance between LLM and human responses, which we aim to minimize.
At each iteration, BO selects a hyperparameter combination using an acquisition function. This function balances exploration, which searches less-tested regions, and exploitation, which focuses on regions that already appear promising. We use Expected Improvement as the acquisition function, which is a standard choice and often performs well. After each evaluation, BO observes the resulting Wasserstein distance and updates the surrogate model with this new information. 
We implement BO using the \texttt{BoTorch} library \citep{BalandatKarrerJiangLethamWilsonBakshy20}. For an introduction to BO, see \cite{Garnett_2023}.

We define the search space as follows: 
temperature takes one of nine values, $0.5$, $1.0$, $1.5$, $1.7$, $1.9$, $2.5$, $3.0$, $3.5$, or $4.0$; 
$\text{top-}p \in [0.5, 1.0]$; 
$\text{min-}p \in [0.05, 0.25]$; 
and $\text{top-}k \in [10, 90]$. 
We assign the first 10 evaluations to an initial exploration phase. In this phase, we quasi-randomly sample hyperparameter combinations to obtain broad coverage of the search space. We then run 5 BO iterations.

Because LLM outputs are stochastic, we do not rely on a single evaluation for each proposed hyperparameter combination. Instead, we run three independent replications for every proposed point. We compute the sample mean and sample variance of the Wasserstein distance across the three runs. These values are then passed to the surrogate model as heteroscedastic noise.

Even though we allow up to three attempts per instruction, high-temperature settings can still fail to produce structured outputs in all three attempts. In such cases, the experiment for that hyperparameter combination fails. Simply discarding these failed runs could bias the surrogate model and lead it to keep exploring unstable regions of the search space. To avoid this, we use a pessimistic imputation rule. When a combination fails, we impute its target value using the largest Wasserstein distance observed among all valid combinations up to that point. This penalizes regions with high hallucination rates and guides the surrogate model toward more stable and reliable hyperparameter combinations.

}

\subsection{Results}

By tuning the temperature and the sampling method (including its thresholds), we achieve substantial reductions in Wasserstein distance relative to default settings. Across experiments, Qwen-7B improved by 3.8\%--50.69\%, and Qwen-32B by 11.98\%--76.09\%. These results suggest that hyperparameter tuning is a fast, effective, and resource‑efficient way to improve distributional alignment.

In rough comparisons of Wasserstein distance improvement ratios, hyperparameter tuning outperforms two prior studies that used fine-tuning for distributional alignment: 38\%–54\% across subpopulations in \cite{suh2025language} and an average 5.8\% in \cite{cao2025specializing}. The positions hyperparameter tuning as a practical approach for leveraging LLMs as human behavior simulators.

\begin{figure}[t] 
\centering
 \FIGURE{
     \includegraphics[width=\textwidth]{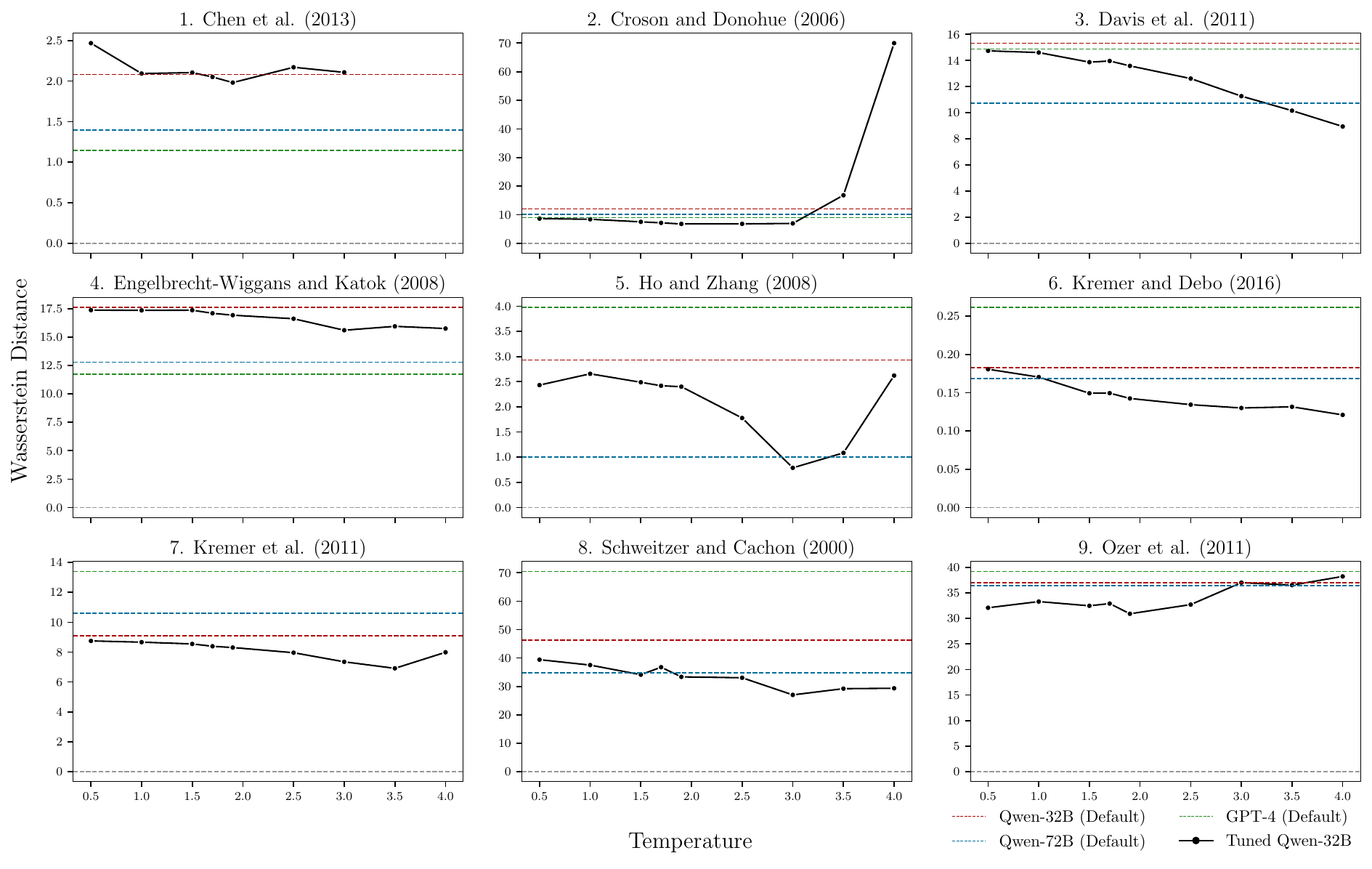}     
     }
     {{ Wasserstein Distance: Tuned Qwen-32B versus Default Qwen-72B and GPT-4} \label{temperature_improvement_rate}}
     {{ The black dot marks the sampling configuration with the lowest average Wasserstein distance at that temperature. Each hyperparameter combination is evaluated across three independent runs, and the black dot reports the best average Wasserstein distance. We exclude any combination for which at least one of the three runs fails to complete. If all combinations at a given temperature have at least one failed run, the best average Wasserstein distance for that temperature is treated as missing and omitted from the results. This occurs, for example, at temperatures 3.5 and 4.0 in the experiment of \cite{doi:10.1287/mnsc.1120.1638}.}}
    \end{figure}

Hyperparameter tuning can also help smaller LLMs match larger ones. In some cases, it also allows open-source models to outperform proprietary models on our behavioral OM tasks. Figure~\ref{temperature_improvement_rate} compares tuned Qwen-32B with default Qwen-72B and GPT-4 across the nine experiments. In most cases, except for the experiments of \citet{doi:10.1287/mnsc.1120.1638} and \citet{doi:10.1287/mnsc.1070.0806}, appropriate hyperparameter tuning allows the smaller model to match the performance of the larger or proprietary models.

These results suggest a practical path for simulating behavioral OM experiments with smaller-scale LLMs, especially when computational resources are limited. 
At the same time, the optimal hyperparameters are highly problem-dependent and difficult to identify in advance. {For example, in \cite{doi:10.1287/mnsc.1070.0788}, Qwen-32B matches Qwen-72B at a temperature of 3.0. In contrast, in \cite{doi:10.1287/mnsc.1100.1258}, Qwen-32B falls short at 3.0 but reaches comparable performance at temperatures of 3.5 or 4.0.} These cases show that hyperparameter choices are problem-specific and that a single setting is unlikely to generalize across experiments.

\begin{figure}[t]
 \FIGURE{
 \includegraphics[width=\textwidth]{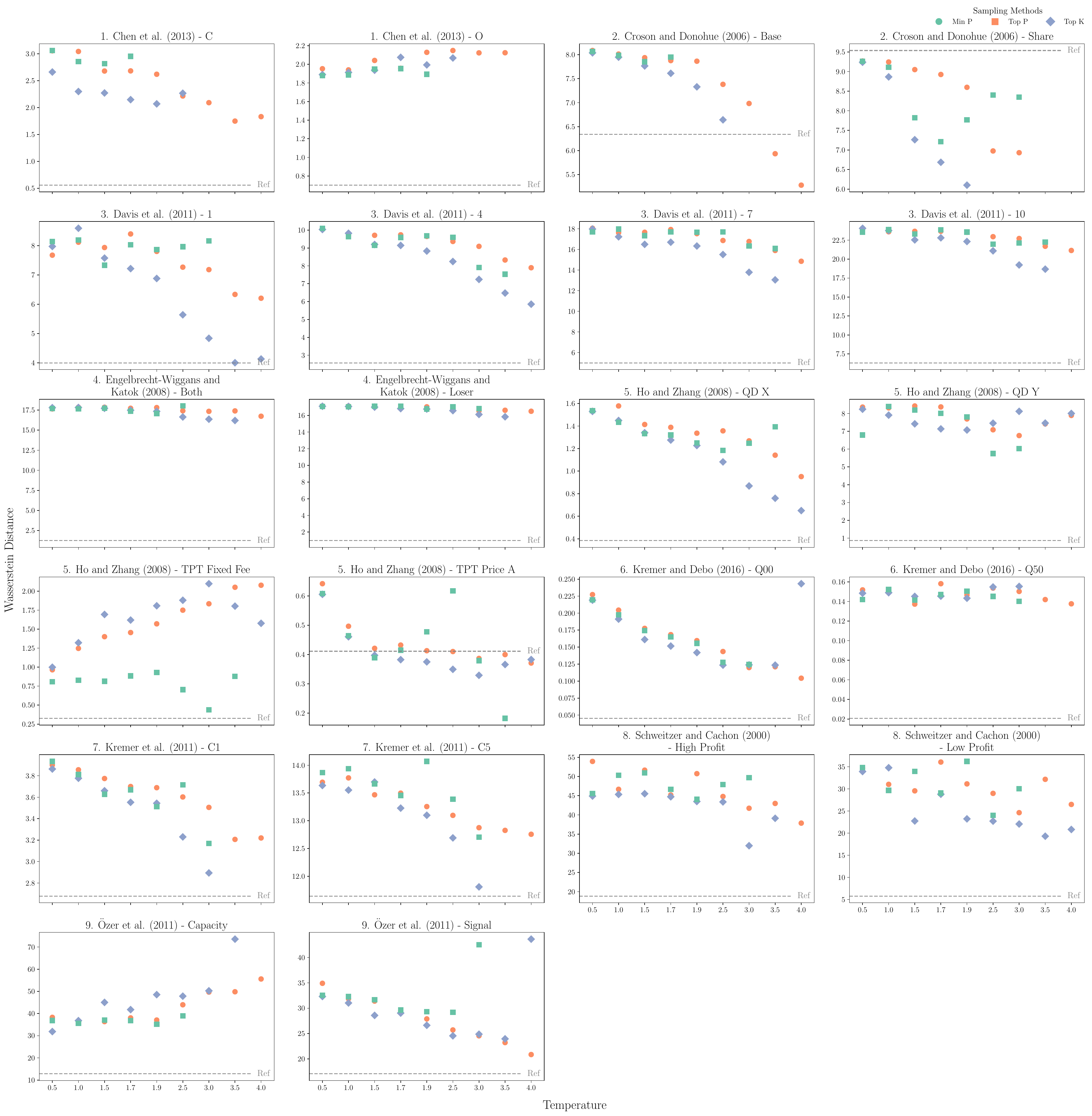}
     }
     {{Wasserstein Distance with Different Temperature and Sampling Methods by Qwen-32B } \label{fig:wass_distance_qwen32}}
     {The different sampling methods are distinguished in different colors. For each sampling method, we use five threshold levels. Since the close amount of survey data are collected from two places, the reference line (labeled “Ref”) represents the mean distance between the reference data collected from different sources.
     }
    \end{figure}

Figure~\ref{fig:wass_distance_qwen32} shows how temperature, sampling method, and threshold level affect the Wasserstein distance across all nine experiments. Section~\ref{subsec:sim-diff-temp} of the e-companion also reports the smallest Wasserstein distances and the corresponding sampling methods at each temperature for both Qwen-7B and Qwen-32B. In general, higher temperatures reduce the Wasserstein distance. This indicates closer alignment between LLM-generated and human-generated data, consistent with Figure~\ref{temperature_improvement_rate}.

However, as discussed in Section~\ref{sec:temp-tuning}, especially Figure~\ref{temperature_success_rate}, temperature creates a trade-off between completion rate and Wasserstein distance. Higher temperature lowers the Wasserstein distance but also reduces completion rates, which increases experimental costs.

\section{Conclusion}\label{sec:Conclusion}
In this paper, we examine whether a range of proprietary and open-source LLMs can replicate human behavior in OM lab experiments from the large-scale replication study of \citet{DavisFlickerHyndmanKatokKepplerLeiderLongTong23}. We evaluate behavioral fidelity using two complementary criteria. First, we compare hypothesis-test outcomes to assess whether LLM-generated data reproduce the original behavioral effects with statistical significance. Second, we measure distributional alignment using Wasserstein distances between LLM and human responses. Together, these criteria distinguish agreement on selected summary effects from alignment with the full distribution of human behavior.

Our findings are mixed. LLMs often replicate hypothesis-test outcomes, suggesting that they can capture some behavioral regularities, decision biases, and forms of bounded rationality, consistent with recent evidence \citep{ChenKirshnerOvchinnikovAndiappanJenkin25}. However, their generated responses frequently diverge from human data at the distributional level. This gap appears across both open-source models and strong proprietary systems such as GPT-4.

We also evaluate two lightweight strategies for reducing distributional misalignment: CoT prompting and hyperparameter tuning. Neither requires large new datasets or model retraining. Although neither fully eliminates the gap, both can meaningfully reduce Wasserstein distances between LLM and human responses. { CoT prompting often mitigates mean mismatch, and our analysis suggests that operational and strategic reasoning stages can help move LLM decisions closer to human behavior.} Hyperparameter tuning can also improve alignment; with appropriate temperature and sampling settings, smaller or open-source models can sometimes match or outperform larger proprietary systems in distributional fidelity.

In summary, our results suggest that LLMs can be useful for rapid theory prototyping and preliminary exploration of behavioral mechanisms in OM, especially when the goal is to examine hypothesis-level patterns. However, caution is needed when the distribution of simulated behavior matters. In applications such as evaluating service systems, testing operational policies, or predicting heterogeneous customer and decision-maker responses, performance depends on the full response distribution rather than only average tendencies or hypothesis-level effects. In such settings, current LLMs are not yet reliable substitutes for human data. Lightweight interventions such as CoT prompting and targeted sampling-parameter tuning can reduce the gap, but their effects are context-dependent and should be evaluated explicitly. As LLMs continue to evolve, continued benchmarking against human baselines will be essential for determining when, and for which OM tasks, they can serve as trustworthy behavioral simulators.

\bibliographystyle{informs2014}
\bibliography{sample}

\ECSwitch

\ECHead{Experimental Settings and Additional Simulation Results}
{
\section{Experimental Settings}\label{experimental_settings}
This section provides the experimental settings for the nine experiments in Table~\ref{tab:llms}, including game introduction, hypotheses and simulated scales of experiments.

\subsection{\cite{doi:10.1287/mnsc.1120.1638} }
\subsubsection{Core Hypothesis}
\cite{doi:10.1287/mnsc.1120.1638} investigate the role of payment scheme in the newsvendor setting. They define two types of payment schemes (Table \ref{tab:payment_schemes}). Under the O payment scheme, participants pay for the items at the point they are ordered and then get a per-unit payment if they are sold. Under the C payment scheme, participants get the margin at the point items are ordered and then are refunded the per unit price for items unsold. In summary, the two payment schemes are: O-payment scheme ($-c$ per unit ordered, $+p$ per unit sold) and C payment scheme (+$(p-c)$ per unit ordered, $-p$ per unit leftover). The two are mathematically equivalent, but differ in the timing of the payments.

\begin{table}[ht]
\begin{center}
\caption{Payment schemes in  \cite{doi:10.1287/mnsc.1120.1638}}
\label{tab:payment_schemes}
\begin{tabular}{lcccccc}
\hline & \multicolumn{2}{c}{ Order-time payments } & & \multicolumn{2}{c}{ Demand-time payments } \\
\cline { 2 - 3 } \cline { 5 - 6 } 
Payment scheme & Per unit ordered & && Per unit sold & Per unit leftover \\
\hline O & $-c$ & && $+p$ & 0 \\
C & $+(p-c)$ && & 0 & $-p$ \\
\hline
\end{tabular}
\end{center}
\end{table}

In this study, we replicate the O and C treatments. The hypothesis is whether subjects order higher quantities under O payment scheme compared to the C payment scheme.

\subsubsection{Task Structure} Consistent with the replication study by \cite{DavisFlickerHyndmanKatokKepplerLeiderLongTong23}, we set $p=\$2$, $c=\$1$ and use an LLM to simulate 50 subjects (25 per condition). Each simulated subject acts as a business manager making order decisions across 25 rounds.

The experiment is framed as a multi-round Newsvendor problem repeated for $T=25$ periods. In each period $t$, the LLM agent received prompt includes the current balance $B_t$ (with initial endowment $B_1=100$) and historical outcomes (decision and balance in the past rounds), and output the order quantity $q_t$. The demand $D_t$ is drawn from a discrete distribution generated by the sum of three independent six-sided dice (i.e., $D_t \in \{3, 4, \dots, 18\}$ with an expected value of $\mu=10.5$).

 Mathematically, given the order quantity $q_t$ and the realized demand $D_t$, the single-period profit is formulated as $\pi_t = 2\min(q_t, D_t) - q_t$. Unsold inventories are discarded at the end of each period, and the agent's objective is to maximize the final wealth $B_{26}$.

\subsubsection{Data Analysis} We conducted Mann-Whitney test to validate whether data under two conditions show significant difference. \textit{p}-value from the data generated by all the LLMs, except for GPT-4o mini, is much lower than 0.01, whereas GPT-4o mini has a \textit{p}-value of 0.806.

 In addition, we analyze the average order quantity under aforementioned two conditions. Aside from the hypothesis, \cite{doi:10.1287/mnsc.1120.1638} also demonstrated that order quantity under scheme O is higher than optimal value derived by expected profit maximization model (10.5), while order quantity under scheme C is lower than 10.5. Our generated data showed that average order quantity under scheme O by different LLMs are LLaMA-70B (9.95), Qwen-72B (9.98), Qwen-32B (9.97), GPT-3.5 (10.78), GPT-4 (11.448). Average order quantity under scheme C are LLaMA (9.50), Qwen-72B (9.06), Qwen-32B (7.49), GPT 3.5 (10.25), GPT-4 (9.87). Thus, GPT-4 showcases more similar average order quantity compared with original data.

\subsection{\cite{doi:10.1287/mnsc.1050.0436}}

\subsubsection{Core Hypothesis}
\cite{doi:10.1287/mnsc.1050.0436} study the bullwhip effect in a multi-tier supply chain consisting of a factory, distributor, wholesaler and retailer. In two studies, they vary whether inventory information is or is not shared and show that inventory information sharing helps to alleviate (but not eliminate) the bullwhip effect.

Our core hypothesis is sharing dynamic inventory information across the supply chain will decrease the level of order oscillation. Specifically, we will conduct the same two treatments in their original study: Beer Game with and without information sharing.

\subsubsection{Task Structure} 
We simulate LLMs as 11 groups of managers to conduct the beer distribution game for each treatment. Each group consists of 4 roles, Retailer, Wholesaler, Distributor, and Factory, operating sequentially from downstream to upstream.

The supply chain operates over $T=48$ weeks. Each member of the chain is responsible for maintaining beer inventory and fulfilling orders from their immediate downstream partner. Crucially, the system features significant delays: a 2-week order transmission delay and a 2-week shipment delay (with the factory subject to a 3-week overall production lead time). We initialize the system in a steady state, where each role starts with 12 kegs of beer in on-hand inventory, alongside orders and shipments of 4 kegs in each pipeline position.

In each period $t$, the LLM agent managing echelon $i$ observes the newly arrived shipment $S^{in}_{i,t}$ and the incoming order $D_{i,t}$ from its downstream partner. The agent fulfills the order from its available inventory. Unfulfilled orders are not lost but kept as backlogs. The agent then makes a single decision: the order quantity $O_{i,t}$ to be sent to its upstream supplier (or the production quantity, if factory). The ending inventory (or backlog) state updates recursively as $I_{i,t} = I_{i,t-1} + S^{in}_{i,t} - S^{out}_{i,t}$.

Under ``Base" treatment, agents operate under decentralized information; they can access their own local history (past orders, shipments, and inventory levels) but cannot observe the states of other echelons or the true consumer demand (except the Retailer); Under ``Share" treatment, except for their own information, each agent can observe the \textit{inventory position} of all team members. The inventory position is defined as the sum of the on-hand inventory and the pipeline inventory (i.e., beer that has been ordered but not yet received). Consumer demand is simulated by the computer and drawn uniformly from an integer set $\{0, 1, \dots, 8\}$. 

The objective of each agent is to minimize the cumulative costs incurred by the team over the 48 weeks. The single-period cost $C_{i,t}$ consists of a $\$0.50$ per-unit holding cost for positive inventory ($I_{i,t} > 0$) and a $\$1.00$ per-unit backlog penalty for negative inventory ($I_{i,t} < 0$).
\subsubsection{Data Analysis} Focusing on the hypothesis first, We conduct one-sided Mann-Whitney test towards variance of order quantity between base condition (benchmark) and share condition (inventory information is revealed to four different roles every time they make decisions). Amongst all the LLMs, LLaMA-70B ($U$ = 1208.5, \textit{p}-value = 0.0042), GPT-3.5 ($U$=1372, \textit{p}-value = 6.97e-05), GPT-4o mini ($U$= 1689.5, \textit{p}-value = 9.24e-11) showed that variance under base condition are significantly greater than variance under share condition, while GPT-4 ($U$ = 7.48, \textit{p}-value = 0.96), Qwen-32B ($U$ = 479, \textit{p}-value = 0.99) and Qwen-72B ($U$=1099, \textit{p}-value = 0.13) could not conclude that variances of two conditions are different.

\subsection{\cite{doi:10.1287/mnsc.1100.1258}}
\subsubsection{Core Hypothesis}
\cite{doi:10.1287/mnsc.1100.1258} investigate how auctioneers set reserve prices in auctions. Subjects play the role of a seller in a second-price auction with $n$ potential buyers. The authors vary the number of bidders as well as the distribution from which the buyers' private values are drawn.

The core hypothesis is that the seller's chosen reserve price is increasing in the number of bidders, specifically when the number of bidders is 1, 4, 7 and 10. The Drop-Out price for bidders are sampled from  cumulative distribution given by $F(v) = (v/100)^{1/3}$, which we refer to as the cube-root distribution.
\subsubsection{Task Structure}
We simulate an auction environment using LLMs to represent 40 independent subjects acting as sellers. Each simulated seller participates in a sequence of 60 independent auctions. In each round, the number of prospective buyers is randomly and uniformly drawn from the set $\{1, 4, 7, 10\}$. 

The auction mechanism is a second-price sealed-bid auction with a seller-defined reserve price. The buyer with the highest drop-out price (i.e., valuation) wins the auction, provided their valuation meets or exceeds the seller's reserve price. The seller's profit in a given round is determined as follows:
\begin{itemize}
    \item \textbf{Profit = Second-Highest Drop-Out Price}, if the second-highest drop-out price is greater than or equal to the reserve price.
    \item \textbf{Profit = Reserve Price}, if the highest drop-out price exceeds the reserve price, but the second-highest drop-out price falls below it.
    \item \textbf{Profit = 0}, if all drop-out prices are strictly below the reserve price (i.e., the item remains unsold).
\end{itemize}

At the end of each period, the LLM agent receives historical feedback, including their chosen reserve price, the realized number of buyers, the winning bid (if applicable), and the profit earned in past rounds. 

Although determining the optimal reserve price requires complex calculations involving order statistics and probability distributions, we do not provide the LLM agents with any auxiliary statistical information. Specifically, agents are not given the explicit probabilities of the item remaining unsold, selling exactly at the reserve price, or selling above the reserve price.
\subsubsection{Data Analysis}
We conduct regression analysis and pairwise t-test to verify hypothesis that the seller's chosen reservation price is increasing in the number of bidders. The dependent variable is the observed reserve price in a period, and the independent variables are (a) the number of bidders minus the average number of bidders and (b) the number of decision period minus the average number of periods, where we could interpret constant as average reservation price and can be compared to risk-neutral theoretical optimal reserve levels (42 for Cube-root). Among all the LLMs, coefficients on $n-\bar{n}$ conducted by LLaMA-70B, Qwen-72B, Qwen-32B, GPT-3.5, GPT-4 and GPT-4o mini are 1.888, 0.935, 1.026, 0.57, 0.599, 0.479, indicating the amount of price LLM simulated sellers increased for each additional bidder. Constants are respectively 19.049, 24,253, 17.736, 27.379, 24.725, 26.695, which are all significantly lower than optimal level: 42. However, by conducting t-test, GPT 3.5 shows no significant different between reservation price under $n=7$ (28.85) and $n=10$ (29.11); GPT-4o mini shows no significant different among $n=4$ (27.33), $n=7$ (27.49) and $n=10$ (27.71). 

\subsection{\cite{doi:10.1287/mnsc.1070.0806} }
\subsubsection{Core Hypothesis}
\cite{doi:10.1287/mnsc.1070.0806} investigate the role of regret and feedback in bidding behavior in first-price sealed-bid auctions. They define two types of feedback. Under ``Loser's Regret" subjects receive feedback on the winning price and how large their missed opportunity to win was (their resale value – winning bid, or zero if they won). Under ``Winner's regret" subjects receive feedback on the second highest bid and how much money was left on the table (their bid – second highest bid, or zero if they did not win).

We replicate 2 treatments: ``Loser's Regret" vs. ``Both" (Both loser's and winner's regret feedback). The core hypothesis is that ``Both" will lead to lower average bids than ``Loser's Regret."

\subsubsection{Task Structure}
We simulate an auction environment where LLMs act as 20 independent buyers (bidders) competing against two computerized opponents to purchase a fictitious asset. The auction mechanism is a first-price sealed-bid auction: the bidder who submits the highest bid wins the asset and pays exactly their submitted bid amount.

The experiment consists of a total of 100 bidding decisions. For each decision, the agent's bid is applied to 10 consecutive, independent auctions. While the agent's bid and resale values remain constant across these 10 auctions, the computerized competitors draw new resale values and generate new bids for each individual auction.

The agent's resale value for the asset is assigned from the set $\{50, 60, 70, 80, 90\}$. The agent holds a constant valuation for 20 consecutive bidding decisions (i.e., 200 individual auctions) before it changes to a new value from the set, presented in a randomized order. The two computerized competitors' valuations are drawn independently from a discrete uniform distribution of integers between 1 and 100. The competitors are pre-programmed to bid in a manner that maximizes their expected earnings against similarly programmed opponents. ($2/3$ of their resale values, but the LLM agent would not know.)

The agent's profit in each individual auction is determined strictly by their bid ($b$) and their assigned resale value ($v$):
\begin{itemize}
    \item \textbf{Profit = $\boldsymbol{v - b}$}, if the agent's bid is the highest among all three bidders (i.e., the agent wins the auction).
    \item \textbf{Profit = 0}, if the agent does not place the highest bid. 
\end{itemize}
Bidding above one's resale value ($b > v$) results in a negative profit (loss) if the auction is won.

At the end of each block of 10 auctions, the agent receives detailed historical feedback for that block. This includes their own bid, the winning price and second-highest bid for each of the 10 auctions, and their realized profit. Additionally, under the `` Both regret " treatment, the agent receives two calculated metrics to evaluate their bidding efficiency:
\begin{enumerate}
    \item \textit{Money left on the table:} Calculated as the agent's bid minus the second-highest bid (if the agent won), representing how much less they could have bid while still winning. It is 0 if the agent lost.
    \item \textit{Missed opportunity to win:} Calculated as the agent's resale value minus the winning bid (if the agent lost, but the winning bid was lower than the agent's resale value). It is 0 if the agent won or if the winning bid exceeded their resale value.
\end{enumerate}
while under the ``Loser's regret", the LLM agent could only get the information of \textit{Missed opportunity to win}.
\subsubsection{Data Analysis}
 We compute bid/value for each subject and conduct one-sided t-test. LLaMA-70B ($t$ = 8.07, \textit{p}-value = 4.45e-10), Qwen-72B ($t$ = 30.86, \textit{p}-value = 8.15e-29), Qwen-32B ($t$ = 4, \textit{p}-value = 0.00013) are consistent with the conclusion; while GPT-3.5 (\textit{p}-value = 0.21), GPT-4o mini (\textit{p}-value = 0.92) are not consistent with the original conclusion.

\subsection{\cite{doi:10.1287/mnsc.1070.0788} }

\subsubsection{Core Hypothesis}
\cite{doi:10.1287/mnsc.1070.0788} investigate a supply chain contracting problem in which they test two-part tariffs as a way to increase supply chain efficiency. They vary the framing of the two-part tariff as either a fixed fee (TPT) or a quantity discount (QD). They find that supply chain efficiency is higher under the quantity discount framing. The core hypothesis we replicate is that supply chain efficiency is higher when a two-part tariff is framed as a quantity discount as opposed to a fixed fee.

\subsubsection{Task Structure}
We use an LLM to simulate 250 subjects per condition, each participating in 11 consecutive rounds. Half of the simulated subjects are assigned the role of Player A (the manufacturer), while the other half act as Player B (the retailer).

\begin{itemize}
    \item \textbf{In the TPT condition}, Player A determines both a per-unit wholesale price (\text{PRICE A}) and a lump-sum \text{FIXED FEE} (a nonnegative integer). Player B must first decide whether to accept the contract. If accepted, Player B commits to paying the fixed fee regardless of the sales volume and proceeds to choose the retail price (\text{PRICE B}) for end consumers. The point earnings are calculated as $(\text{PRICE A} - 2) \times \text{QUANTITY} + \text{FIXED FEE}$ for Player A, and $(\text{PRICE B} - \text{PRICE A}) \times \text{QUANTITY} - \text{FIXED FEE}$ for Player B. If Player B rejects the offer, the round ends and both players earn zero points.

    \item \textbf{In the QD condition}, Player A determines a quantity discount pricing scheme where the average unit price decreases as Player B buys more. Specifically, the scheme is formulated as $\text{PRICE A} = x + y / \text{QUANTITY}$, where Player A chooses an integer $x \in [0, 10]$ and a nonnegative integer $y$. If Player B accepts this pricing scheme, she chooses the retail price (\text{PRICE B}). The point earnings are $(\text{PRICE A} - 2) \times \text{QUANTITY}$ for Player A, and $(\text{PRICE B} - \text{PRICE A}) \times \text{QUANTITY}$ for Player B. Similar to the TPT condition, if the offer is rejected, both players earn zero points.
\end{itemize}

\subsubsection{Data Analysis}
 We conduct two-sample t-test to compare the efficiency between these two players. All the LLMs except for LLaMA-3B, Qwen-7B and Qwen-14B, show results that channel efficiency is higher in QD condition than in TPT condition with $\textit{p}$-value $< 0.01$.

\subsection{\cite{doi:10.1287/mnsc.2015.2264}}
\subsubsection{Core Hypothesis}
\cite{doi:10.1287/mnsc.2015.2264} simulate queue experiments and investigate the impact of waiting time on purchasing decisions of the uninformed consumers, who do not know the true value of a product under two conditions varied from the existence of informed consumers, who know the true value of the product. In a simulated queueing experiment, they find that uninformed consumers can infer product quality from observed waiting times if there are enough informed consumers. 

The core hypotheses could be divided into two parts. Relative to the setting with no informed consumers ($q = 0$), the presence of informed consumers ($q = 0.50$) makes uninformed consumers (a) less likely to purchase upon observing a short wait ($w = 1$) and (b) less sensitive to the purchase probability reduction associated with each marginal unit of wait time. We test these two findings in the setting with a high prior of quality ($p_0=0.5$)

\subsubsection{Task Structure}
We replicate the original queue experiment under two analogous conditions. 
\begin{enumerate}
    \item No informed consumers (i.e., $q00$): no one knows the true value of the product, and all consumers only have a prior belief that the product has a 50\% chance of being high profit. 
    \item 50\% informed consumers (i.e., $q50$): populations know with certainty whether the product is high or low profit with 50\% probability.
\end{enumerate}
We simulate LLM as 32 subjects under $q00$ treatment and 68 subjects under $q50$. In both conditions, participants will be put into cohorts of four, each subject in one cohort would simulate consumers 

For one cohort, we simulate a sequential market environment where LLMs act as potential consumers deciding whether to purchase a newly released product. The market consists of 4 potential consumers (all simulated by LLMs) who arrive sequentially in a randomly determined order. The experiment spans 26 independent product rounds. 

At the beginning of each round, the true value of the product is randomly drawn: it is either a high value of 3.50 (with 50\% probability, i.e., $p_0 = 0.5$) or a low value of 0 (with 50\% probability). Concurrently, each consumer is randomly assigned an information type:
In the $q00$ condition, $q = 0$; while in the $q50$ condition, $q = 0.5$;
\begin{itemize}
    \item \textbf{Informed Consumers (with probability of $q$):} With a $q$ probability, the consumer is informed and perfectly observes the true product value (3.50 or 0) before making a purchase decision.
    \item \textbf{Uninformed Consumers (with probability of $1-q$):} With a $q$ probability, the consumer is uninformed. They only know the prior probability distribution of the product's value and can observe the delivery time before deciding.
\end{itemize}

The firm can deliver exactly one unit of the product per week. When a consumer arrives, the firm assigns a delivery time (in weeks), which perfectly reveals the number of consumers who arrived earlier and chose to purchase. For example, a delivery time of 2 weeks indicates that exactly one prior consumer placed an order. However, the arriving consumer does not know their exact position in the arrival queue (e.g., whether they are the 2nd arrival or the 4th). Every week of wait time imposes a waiting cost of 1.00 on the consumer.

Each consumer begins the round with an initial endowment of 4.00. Upon observing their information type and the specified wait time $w$, the agent makes a binary decision: \textit{Order} or \textit{Do Not Order}.
\begin{itemize}
    \item If the agent chooses \textbf{Do Not Order}, they keep their endowment, yielding a final payoff of 4.00.
    \item If the agent chooses \textbf{Order}, their final payoff is calculated as: 
    \[ \text{Payoff} = 4.00 - (1.00 \times w) + \text{True Product Value} \]
\end{itemize}
Uninformed consumers must therefore weigh the guaranteed 4.00 from not ordering against the expected payoff of ordering, dynamically updating their belief about the product's true value based on the observed wait time (which serves as a signal of prior consumers' aggregate purchasing behavior).
\subsubsection{Data Analysis}
\cite{DavisFlickerHyndmanKatokKepplerLeiderLongTong23} adopted two probit regressions.\footnote{First probit regression : The consumer's purchase decision will be the dependent variable. The regression will consist of eight dummy variables representing all interactions of treatment \{$q00$, $q50$\} × waiting time $\{1, 2, 3, 4\}$. We will then conduct a Wald-test evaluating whether the coefficient on $q50$ interacted with waiting time of 1 is negative and significant.

Second probit regression: the consumer's purchase decision as the dependent variable. We will then include as independent variables the following: (1) the waiting time as a continuous variable, (2) a dummy variable for the $q50$ condition, (3) an interaction term between waiting time and the $q50$ dummy (4) and a constant. We will then determine whether the coefficient on (3) is positive and significant.}
Due to the multicollinearity generated by original data analysis methods, we focus on purchase probability of uniformed consumers with or without informed consumers. For the first hypothesis, we made comparison between purchase probability of uninformed consumers under two conditions when observing short waiting time; For the second hypothesis, we compare purchase probability decreasing rate as the waiting time increases.

By taking ``thinking" process of LLaMA-70B as reference, we could make inference of LLMs reasoning. Under $q00$ condition with no informed consumers, LLMs make rational decisions. For instance, when waiting time is one week, LLMs calculate and make comparison between expected earnings with initial endowment. \textit{Given the expected earnings when placing an order is 4.75 and the endowment if not placing an order is 4.00, with a delivery time of 1 week, it's beneficial to place an order since the expected earnings are higher than the endowment}. Same as $q00$, when LLaMA-70B face condition with informed consumers, diverged from human participants, they prone to ignore decisions made by former consumers and behave rationally. Thus, LLaMA-70B shows results misalignment with human response. 

\subsection{\cite{doi:10.1287/mnsc.1110.1382}}
\subsubsection{Core Hypothesis}
\cite{doi:10.1287/mnsc.1110.1382} study how individuals make forecasting decisions based on time-series data. They consider the effect of two kinds of random errors: temporary shocks and permanent shocks.

The core hypothesis is forecasters overreact to forecast error for low values of $W=c^2/n^2$ (more stable environment) and underreact for high values of $W=c^2/n^2$  (less stable environments). 

\subsubsection{Task Structure}
We generate true demands data by changing the random error parameters as follows.
$$
\begin{aligned}
& D_t=\mu_t+\varepsilon_t, \\
& \mu_t=\mu_{t-1}+v_t
\end{aligned}
$$
where $\varepsilon_t \sim \mathcal{N}\left(0, n^2\right)$ and $v_t \sim \mathcal{~N}\left(0, c^2\right)$. Under stable environment condition, $c=0$, $n=10$; Under less stable environment condition, we set $c=40$, $n=10$. 

We simulate LLMs as 67 managers. For each round, we let the LLMs know the true demand data for the most recent 30 weeks and let them output the predicted value for the next week. The session last for 50 rounds. 

\subsubsection{Data Analysis}
The smoothing factor, $\alpha$, is compared with the theoretical optimal value, $\alpha^*(W)$, to assess overreaction or underreaction. 
The value of $\alpha$ can be formalized as follows: 
$$
\alpha^*(W)=\frac{2}{1+\sqrt{1+4 / W}}
$$
where $W$=$c^2$/$n^2$, denoting the change-to-noise ratio. 
Note that $\alpha^*(W)$ equals 0 for stable environment, while for unstable environment it is 0.94. The original paper provides a behavioral model to estimate~$\alpha$:
$$
\begin{aligned}
\Delta F_{t+1(c, s, i)}=a_1^{s i} E_t+\tilde{a}_2 F_1+a_3^{s i} \Delta D_t+a_4 \Delta D_{t-1}+a_5 \Delta F_t +a_6 \Delta F_{t-1}+\text{constant}+v_{s(c)}+w_{i(s, c)}+\varepsilon_t
\end{aligned}
$$

The model uses a three-level nested structure to account for random effects and variability in behavioral parameters across individuals and conditions. Only coefficients $a_1^{s i}$ and $a_3^{s i}$ are treated as random slopes to avoid issues with non-convergence and ensure accurate estimation.

GPT-3.5, GPT-4 and GPT-4o mini showed that $\alpha$ value under stable environment is 0.084, 0.164, 0.079, which is higher than theoretical value, 0 and showed that managers tend to overreact to stable environment ; $\alpha$ value under less stable environment is 0.207, 0.331, 0.199, which is lower than 0.94 and showed that managers tend to underreact to less stable environment.

\subsection{\cite{doi:10.1287/mnsc.46.3.404.12070}}
\subsubsection{Core Hypothesis}

\cite{doi:10.1287/mnsc.46.3.404.12070} investigate newsvendor order quantity decisions for both high profit products (critical fractile of 75\%) and low-profit products (critical fractile of 25\%). They find that average order quantities are set too low for high profit products and too high for low-profit products. 

The core hypothesis is newsvendor order quantities are set too low for high-profit products and too high for low-profit products.

\subsubsection{Task Structure}
We simulate LLMs as 40 participants in a 30-round experiment, where 15 rounds involve high-profit products, and the other 15 rounds involve low-profit products.

For high-profit products, simulated participants buy each unit at 3 francs and sell it at 12 francs. Any unsold units are discarded with no salvage value; For low-profit products, simulated participants buy each unit at 9 francs and sell it at 12 francs, with no-salvage condition. Demand is uniformly distributed between 1 and 300 units per round. 
\[
\text{Profit} = (\text{selling price} \cdot \min(\text{demand}, \text{order})) - (\text{purchase price} \cdot \text{order})
\]
\subsubsection{Data Analysis}
We conduct two one-sample t test, comparing the averaged observed quantity to normative prediction. Under high profit condition, average order quantity of LLaMA-70B, Qwen-72B, Qwen-32B, GPT-3.5, GPT-4, GPT-4o mini are 163.69, 175.32, 150.18, 147.08, 183.80, 117.225, which are all lower than normative prediction for high profit products, 225; Under low profit, these models exhibit average order quantity of 149.94, 149.65, 146.19, 144.46, 171.29, 110.21, which are all higher than normative prediction for low profit products, 75.

\subsection{\cite{doi:10.1287/mnsc.1110.1334}}
\subsubsection{Core Hypothesis}
\cite{doi:10.1287/mnsc.1110.1334} study a supply chain setting where a supplier must make a capacity decision, and a manufacturer has private forecast information about market demand. The manufacturer can send a cheap talk message about the private forecast to the supplier. Standard theory predicts that the manufacturer's message should be uninformative (i.e. uncorrelated with the true private forecast), and that the supplier should ignore the message
(i.e. the capacity decision should be uncorrelated with the message). \cite{doi:10.1287/mnsc.1110.1334} find in human-subject experiments that there is informative communication, with manufacturers sending higher messages when the private forecast is higher, and suppliers choosing higher capacity when receiving higher messages.

The core hypothesis is that manufacturers' messages will be positively correlated with their private forecast, and suppliers' capacity decisions will be positively correlated with the messages received in the CHUH treatment.

\subsubsection{Task Structure}
We simulate LLMs as 40 manufacturers and 40 suppliers, prompting them to complete 30 rounds experiments.

The experiment is a two-player interaction between a manufacturer and a supplier under a wholesale price contract. In each round, the supplier chooses capacity before demand is realized, while the manufacturer privately observes a a forecast signal about market demand and can strategically communicate it to the supplier.

Market demand is determined by
\[
D = \mu + \xi + \varepsilon,
\]
where
\begin{itemize}
    \item \(\mu = 250\) is the common-knowledge baseline demand;
    \item \(\xi\) is the manufacturer's private forecast signal;
    \item \(\varepsilon \sim U[-75,75]\) is the market uncertainty.
\end{itemize}

The manufacturer observes the value of \(\xi\), while the supplier does not. The supplier knows \(\mu\), the distribution of \(\xi\), and the distribution of \(\varepsilon\).
\[
\xi \sim U[-150,150].
\]
Both players also observe the history of previous rounds.

Each round proceeds as follows:
\begin{enumerate}
    \item Nature determines the manufacturer's private forecast \(\xi\).
    \item The manufacturer observes \(\xi\) and sends a report \(\hat{\xi}\) to the supplier.
    \item The supplier observes \(\hat{\xi}\) and chooses capacity \(K\).
    \item Nature determines the market shock \(\varepsilon\).
    \item Demand \(D = \mu + \xi + \varepsilon\) is realized.
    \item Trade quantity is given by $q = \min(D,K)$.
    \item Payoffs are computed for both players.
\end{enumerate}

The manufacturer's payoff is
\[
\Pi^{M} = 25 \cdot \min(D,K).
\]

The supplier's payoff is
\[
\Pi^{S} = 75 \cdot \min(D,K) - 60K.
\]

\subsubsection{Data Analysis}

By conducting OLS regression to regress manufacturers' message on private information, and supplier's capacity on received message, we examine whether there exists trust and trustworthiness behavior. Among all the LLMs, LLaMA-70B, Qwen-72B, Qwen-32B, GPT-3.5, GPT-4, GPT-4o mini show coefficients on manufacturer's private information: 0.9623, 0.870, 0.988, 0.425, 0.394, 0.555 and coefficients on supplier's private information: 0.399, 0.634, 0.352,0.202, 0.229, 0.281.}

{

\section{Additional Analysis of Multi-agent Experiments}  \label{ec:multi-agent}

\subsection{Experiment of \citet{doi:10.1287/mnsc.1050.0436}}

\citet{doi:10.1287/mnsc.1050.0436} study a four-tier supply chain consisting of a Retailer, Wholesaler, Distributor, and Factory. The focal hypothesis is that sharing dynamic inventory information across the supply chain reduces order oscillation. We replicate the two treatments in the original study: the Beer Game without information sharing and the Beer Game with information sharing.

At the aggregate level, only GPT-3.5 and GPT-4o mini replicate the hypothesized reduction in order variance. The remaining models fail to pass the one-sided Mann-Whitney test comparing order variance between the base condition and the information-sharing condition. To understand the source of these failures, we conduct role-specific one-sided Mann-Whitney tests for each supply chain position. The results are reported in Table~\ref{tab:multi_agent_roles}.

\begin{table}[t]
\TABLE{{  Role-Specific Mann-Whitney Test Results for Order Variance} \label{tab:multi_agent_roles}}
{
{

\begin{tabular}{lcccc}
\toprule
\textbf{Model} & \textbf{Retailer} & \textbf{Wholesaler} & \textbf{Distributor} & \textbf{Factory} \\
\midrule
Deepseek & 0.008 (\Checkmark) & 0.300 (\ding{55}) & 0.578 (\ding{55}) & 0.803 (\ding{55}) \\
GPT-3.5 & 0.094 (\ding{55}) & 0.009 (\Checkmark) & 0.003 (\Checkmark) & 0.039 (\ding{55}) \\
GPT-4 & 0.002 (\Checkmark) & $<0.001$ (\Checkmark) & 0.972 (\ding{55}) & 0.998 (\ding{55}) \\
GPT-4o mini & $<0.001$ (\Checkmark) & $<0.001$ (\Checkmark) & $<0.001$ (\Checkmark) & 0.008 (\Checkmark) \\
LLaMA-3B & 0.050 (\ding{55}) & 0.991 (\ding{55}) & 0.653 (\ding{55}) & 0.794 (\ding{55}) \\
LLaMA-8B & 0.999 (\ding{55}) & 1.000 (\ding{55}) & 1.000 (\ding{55}) & 1.000 (\ding{55}) \\
LLaMA-70B & 0.999 (\ding{55}) & 1.000 (\ding{55}) & 0.976 (\ding{55}) & 0.212 (\ding{55}) \\
Qwen-7B & 1.000 (\ding{55}) & 1.000 (\ding{55}) & 0.998 (\ding{55}) & 1.000 (\ding{55}) \\
Qwen-14B & 0.994 (\ding{55}) & 0.998 (\ding{55}) & 0.500 (\ding{55}) & 0.862 (\ding{55}) \\
Qwen-32B & 0.972 (\ding{55}) & 0.991 (\ding{55}) & 0.994 (\ding{55}) & 0.969 (\ding{55}) \\
Qwen-72B & 0.972 (\ding{55}) & 0.277 (\ding{55}) & 0.018 (\ding{55}) & 0.323 (\ding{55}) \\
\bottomrule
\end{tabular}
}
}
{
{
\emph{Note.} The table reports $p$-values from one-sided Mann-Whitney tests comparing order variance between the base condition without information sharing and the condition with information sharing. A checkmark (\Checkmark) indicates statistical significance at the 1\% level; a cross (\ding{55}) indicates failure to pass the test.}
}
\end{table}

The role-level results reveal three patterns. First, the failures of the LLaMA and Qwen series are largely systemic: agents in nearly all supply chain positions fail to use shared inventory information to reduce order variance. Second, GPT-4o mini exhibits systemic success, passing the hypothesis in all four roles. Third, Deepseek, GPT-4, and GPT-3.5 display role-specific heterogeneity. Deepseek passes the hypothesis only for the Retailer, while GPT-4 passes for the Retailer and Wholesaler but fails for the Distributor and Factory. GPT-3.5 shows a different pattern, succeeding in the two intermediate roles but failing at the Retailer and Factory positions.

Figure~\ref{r2_croson_gpt_4} illustrates this heterogeneity using GPT-4 as an example. Although GPT-4's Retailer and Wholesaler reduce order variance under information sharing, the Distributor and Factory exhibit large variance spikes. This pattern suggests that the difficulty of multi-agent LLM simulations may depend on structural position. Upstream agents face accumulated demand distortions, delayed information, and compounded backlogs. These features increase the cognitive burden of interpreting shared inventory signals. When upstream agents fail, they can disproportionately destabilize the entire supply chain, even if downstream agents respond appropriately.

\begin{figure}[t]
\FIGURE{
\includegraphics[width=\textwidth]{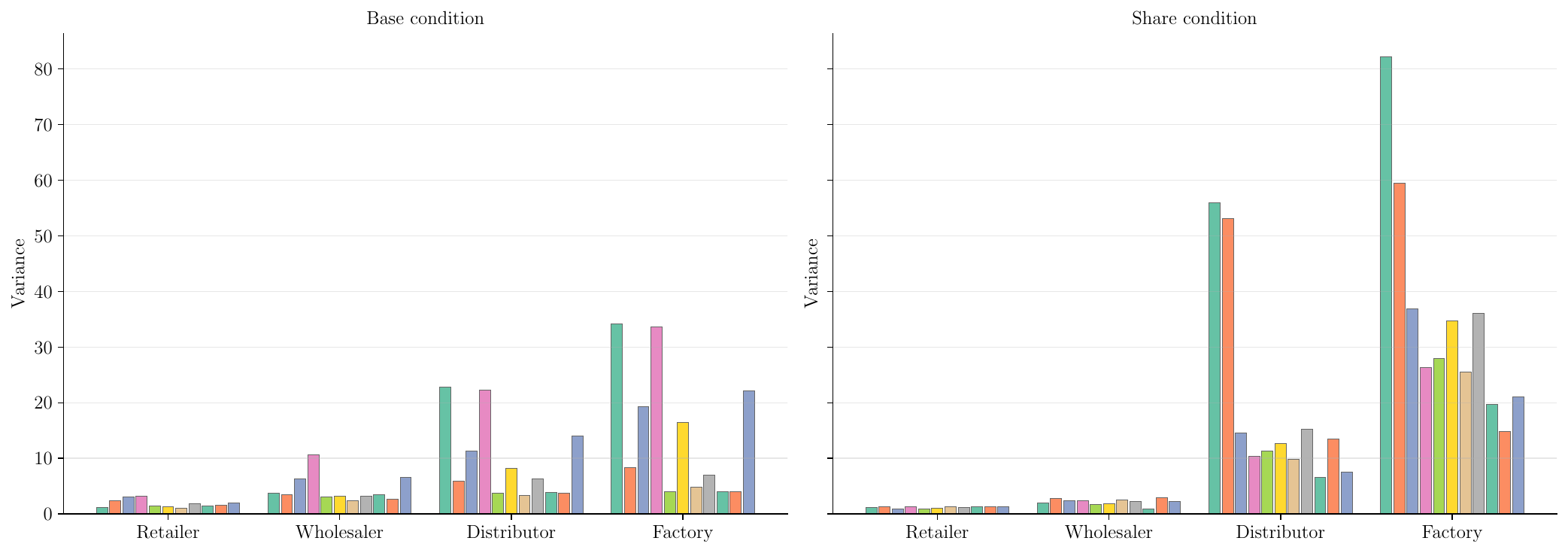}
}
{{ Group Variance in the Experiment of \citet{doi:10.1287/mnsc.1050.0436} for GPT-4 }\label{r2_croson_gpt_4}}
{{ The bar charts report order variance across 11 independent groups for the four supply chain roles: Retailer, Wholesaler, Distributor, and Factory. The left panel corresponds to the base condition without information sharing; the right panel corresponds to the information-sharing condition.}}
\end{figure}

\subsection{Experiment of \citet{doi:10.1287/mnsc.1070.0788}}

\citet{doi:10.1287/mnsc.1070.0788} study a supply chain contracting problem and examine whether two-part tariffs improve channel efficiency. The same contract can be framed either as a fixed-fee contract, denoted TPT, or as a quantity discount contract, denoted QD. The key hypothesis is that supply chain efficiency is higher under the QD framing than under the TPT framing.

The experiment has two roles: Manufacturer and Retailer. In the TPT condition, the Manufacturer chooses a per-unit wholesale price $w$ and a nonnegative fixed fee $F$. The Retailer then decides whether to accept the contract. If the contract is accepted, the Retailer chooses the retail price $p$; if it is rejected, both parties earn zero. The Manufacturer's payoff is $(w-2)\times \text{QUANTITY}+F$, and the Retailer's payoff is $(p-w)\times \text{QUANTITY}-F$.

In the QD condition, the Manufacturer chooses a quantity discount schedule of the form
\[
\text{PRICE A}=x+\frac{y}{\text{QUANTITY}},
\]
where $x\in[0,10]$ is an integer and $y$ is a nonnegative integer. This formulation is economically equivalent to the TPT condition, with $x$ corresponding to $w$ and $y$ corresponding to $F$. If the Retailer accepts the schedule, she chooses the retail price $p$; otherwise, both parties earn zero. We therefore continue to use $w$ and $F$ when discussing the QD condition.

The standard economic benchmark is that the Manufacturer should set the wholesale price close to marginal cost to reduce double marginalization, and then extract channel surplus through the fixed fee. In the human experiment, however, the TPT framing makes the fixed fee salient and induces higher rejection rates. The QD framing makes the fixed component less salient, mitigates this rejection tendency, and generates higher channel efficiency.

Several models fail to reproduce this framing effect. In the GPT series, GPT-3.5 and GPT-4o mini fail the hypothesis. The same is true for LLaMA-3B, LLaMA-70B, Qwen-7B, and Qwen-14B. Figure~\ref{r5_ho_analysis} reports the average channel efficiency, acceptance rate, wholesale price, fixed fee, retail price, and channel efficiency conditional on acceptance across the 11 LLMs.

Two mechanisms explain the failures. The first is a Retailer-side acceptance failure. Qwen-7B rejects all Manufacturer contracts, producing zero acceptance and hence zero efficiency under both TPT and QD. LLaMA-3B exhibits a similar, though less extreme, pattern, with acceptance rates of only 5\% under TPT and 3\% under QD. These models fail primarily because the Retailer role rejects contracts too often.

The second mechanism is a pricing and optimization failure among models with nontrivial acceptance rates. Under QD, several Manufacturer agents fail to implement the surplus-extraction logic. Instead of using the fixed component to extract surplus while keeping the wholesale price low, they raise the wholesale price. For example, GPT-4o mini increases $w$ from 4.18 under TPT to 6.45 under QD; GPT-3.5 increases $w$ from 4.26 to 6.31; and LLaMA-70B increases $w$ from 4.95 to 5.42 while reducing $F$ to 1.32. This pattern intensifies double marginalization, raises the Retailer's optimal price, and reduces total channel surplus.

Retailer pricing also becomes less accurate under the QD framing. Although the TPT and QD contracts are economically equivalent, the QD representation bundles the wholesale price and fixed fee into a nonlinear average-cost expression. This transformation appears to weaken the models' ability to compute the Retailer's optimal price. We measure this by the mean squared error between the chosen retail price $p$ and the theoretical optimum $p^*=(d+w)/2$. For GPT-4o mini, the pricing MSE increases from 2.99 under TPT to 5.96 under QD. For GPT-3.5, it increases from 2.59 to 5.35. For Qwen-14B, it increases from 1.09 to 1.55. Thus, some models fail not because they are insensitive to framing, but because the framing changes the computational representation of the decision problem and induces less accurate optimization.

\begin{figure}
\FIGURE{
\includegraphics[width=\textwidth]{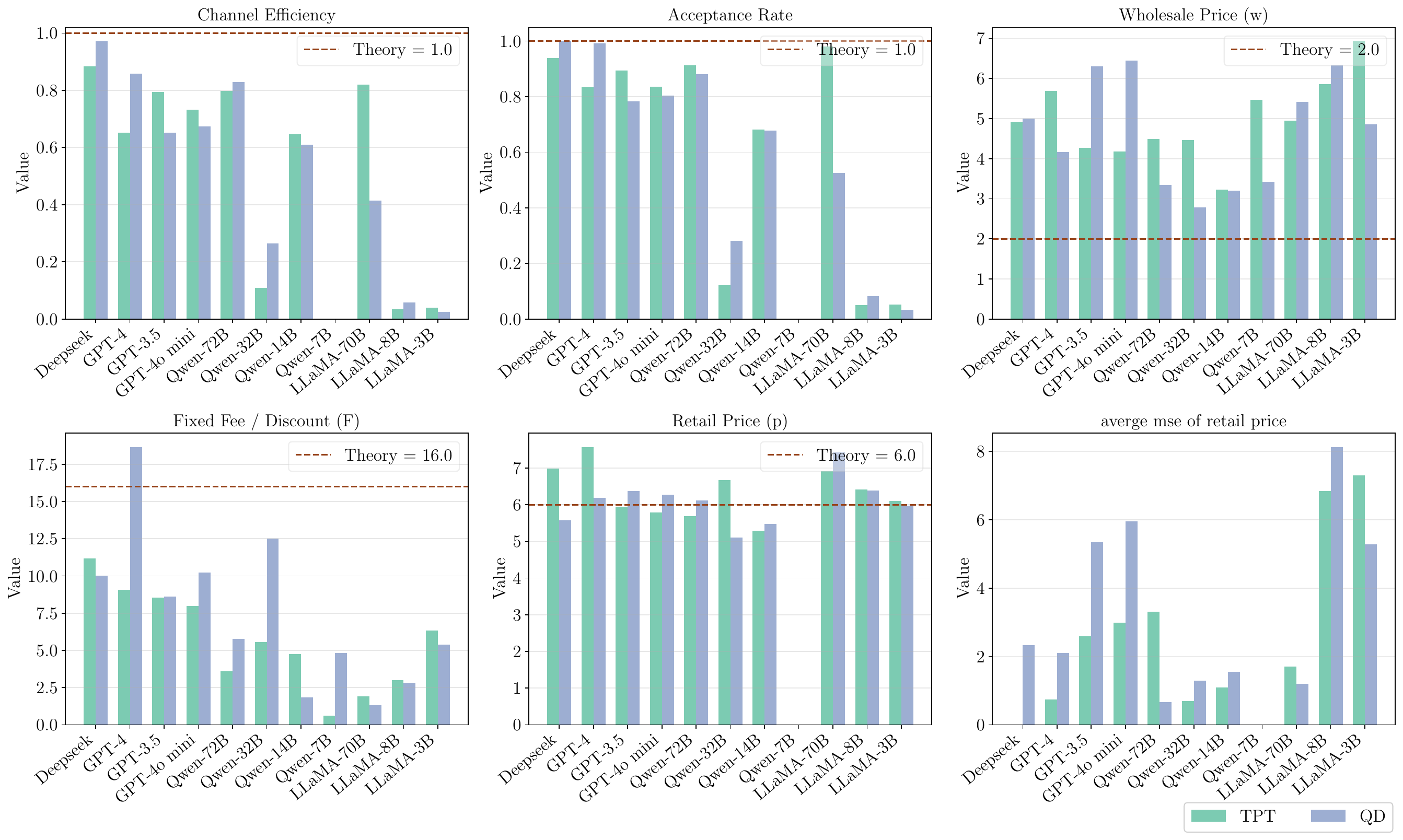}
}
{{ Analysis of the Experiment of \citet{doi:10.1287/mnsc.1070.0788} }\label{r5_ho_analysis}}
{{ The figure reports model-level averages for channel efficiency, acceptance rate, wholesale price, fixed fee, retail price, and channel efficiency conditional on contract acceptance under the TPT and QD conditions.}}
\end{figure}

\subsection{Experiment of \citet{doi:10.1287/mnsc.1110.1334}}

\citet{doi:10.1287/mnsc.1110.1334} study a supply chain with asymmetric demand information. We simulate LLMs as a Manufacturer and a Supplier interacting under a wholesale price contract. Demand is given by
\[
D=\mu+\xi+\epsilon,
\]
where $\mu=250$ is common knowledge, $\epsilon$ is market uncertainty uniformly distributed on $[-75,75]$, and $\xi$ is the Manufacturer's private forecast information. The Manufacturer observes $\xi$ exactly, whereas the Supplier only knows that $\xi$ is uniformly distributed on $[-150,150]$.

Each round proceeds as follows. First, the Manufacturer observes $\xi$ and sends a cheap-talk report $\hat{\xi}$ to the Supplier. Second, the Supplier observes $\hat{\xi}$ and chooses capacity $K$ at unit cost $c_K>0$. Third, demand is realized and the Manufacturer places an order. Fourth, the Supplier produces $\min(D,K)$ at unit cost $c>0$ and charges wholesale price $w$ per unit delivered. Finally, the Manufacturer sells the product at retail price $r>0$. Given $K$ and $\xi$, the Supplier's and Manufacturer's expected profits are
\[
\Pi^s(K,\xi)=(w-c)\,\mathbb{E}_{\epsilon}\!\left[\min(\mu+\xi+\epsilon,\,K)\right]-c_KK,
\]
and
\[
\Pi^m(K,\xi)=(r-w)\,\mathbb{E}_{\epsilon}\!\left[\min(\mu+\xi+\epsilon,\,K)\right].
\]
If the Supplier knew $\xi$, his optimal capacity would be
\[
K^s(\xi)=\mu+\xi+G^{-1}\left(\frac{w-c-c_K}{w-c}\right),
\]
where $G$ is the cumulative distribution function of $\epsilon$.

This setting separates the multi-agent interaction into two role-specific behavioral requirements. The Manufacturer must generate informative messages, and the Supplier must respond to those messages. Accordingly, we test two hypotheses:
\begin{itemize}
    \item Manufacturer behavior (Hyp OZ$_a$): the reported signal $\hat{\xi}$ is positively correlated with the true private forecast $\xi$.
    \item Supplier behavior (Hyp OZ$_b$): the capacity decision $K$ is positively correlated with the received signal $\hat{\xi}$.
\end{itemize}

The OLS regression results in Figure~\ref{r9_supplier_analysis} show that almost all models, including LLaMA-3B, pass Hyp OZ$_a$. Even LLaMA-3B generates cheap-talk messages that are positively correlated with the true forecast ($p<0.01$, slope $=0.27$). The larger divergence appears in Supplier behavior. More capable models such as Qwen-72B and LLaMA-70B exhibit strong positive responses to the received signal, with slopes of 0.72 and 0.62, respectively, both significant at the 1\% level. In contrast, the LLaMA-3B Supplier fails to respond meaningfully to the signal, with a near-zero slope of 0.0196 and $p=0.0296$. This suggests that the main driver of LLaMA-3B's multi-agent failure is not the absence of communication by the Manufacturer, but the Supplier's inability to condition capacity on the communicated information.

\begin{figure}[ht]
\FIGURE{
\includegraphics[width=\textwidth]{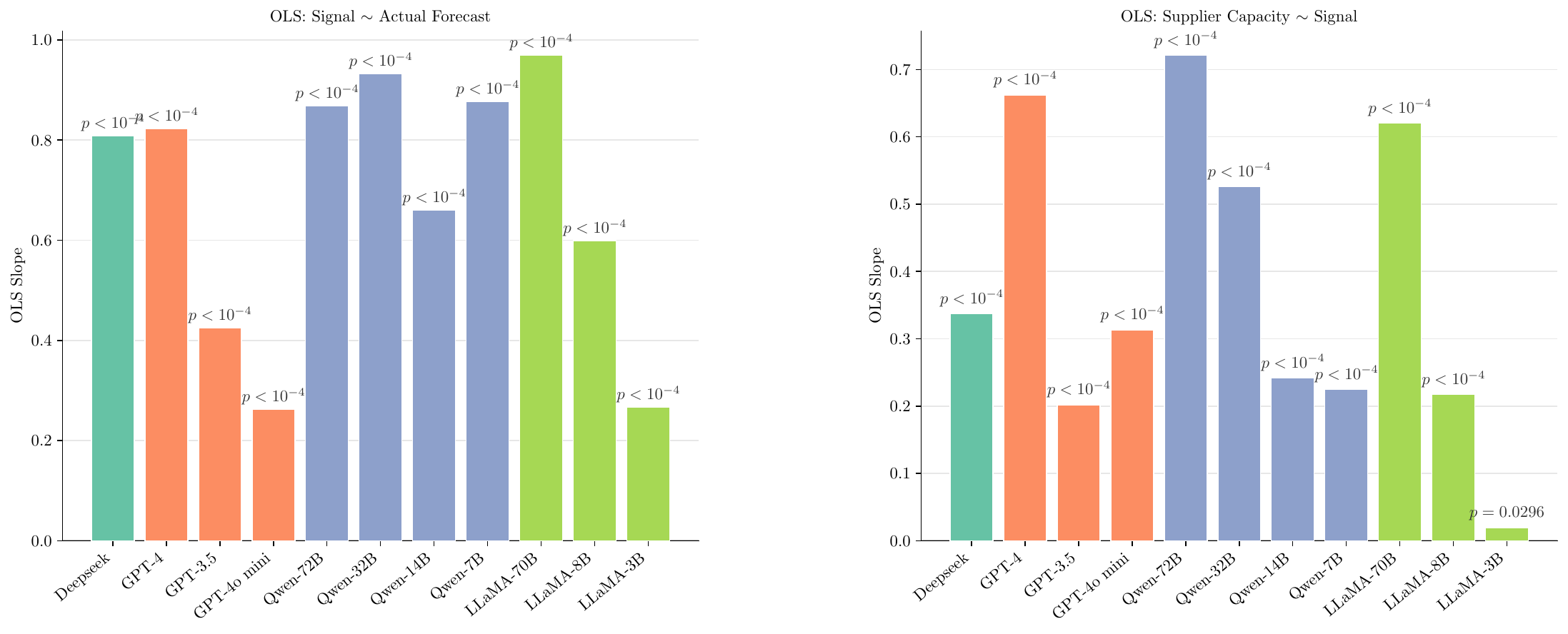}
}
{{ Regression Results in the Experiment of \citet{doi:10.1287/mnsc.1110.1334} }\label{r9_supplier_analysis}}
{{ The figure reports OLS slopes from two regressions. The left panel regresses the Manufacturer's reported signal $\hat{\xi}$ on the true private forecast $\xi$. The right panel regresses the Supplier's capacity decision $K$ on the received signal $\hat{\xi}$.}}
\end{figure}

To further illustrate the agent-level behavior, Figure~\ref{r9_scatter_plot} compares Qwen-72B, which passes both hypotheses, with LLaMA-3B, which fails on the Supplier-response dimension. Qwen-72B generates highly informative Manufacturer reports, with observations concentrated near the diagonal $\hat{\xi}=\xi$. LLaMA-3B also produces statistically informative reports, but the relationship is much noisier: similar forecast values can generate widely dispersed messages, and similar messages can correspond to very different forecast values. Thus, although the LLaMA-3B Manufacturer passes the correlation test, its communication is substantially less precise.

The Supplier-side contrast is sharper. Qwen-72B's capacity choices increase strongly with the received signal, although they often exceed the full-trust benchmark $K^s(\hat{\xi})$, indicating overreaction rather than calibrated partial trust. By contrast, the LLaMA-3B Supplier is nearly insensitive to the received signal and frequently chooses capacity below the no-trust benchmark $K^a$. This behavior suggests a conservative default policy: the Supplier largely discounts the Manufacturer's message and underinvests in capacity. The resulting breakdown is therefore a role-specific failure of signal use, rather than a complete failure of signal generation.

\begin{figure}[ht]
\FIGURE{
\includegraphics[width=0.7\textwidth]{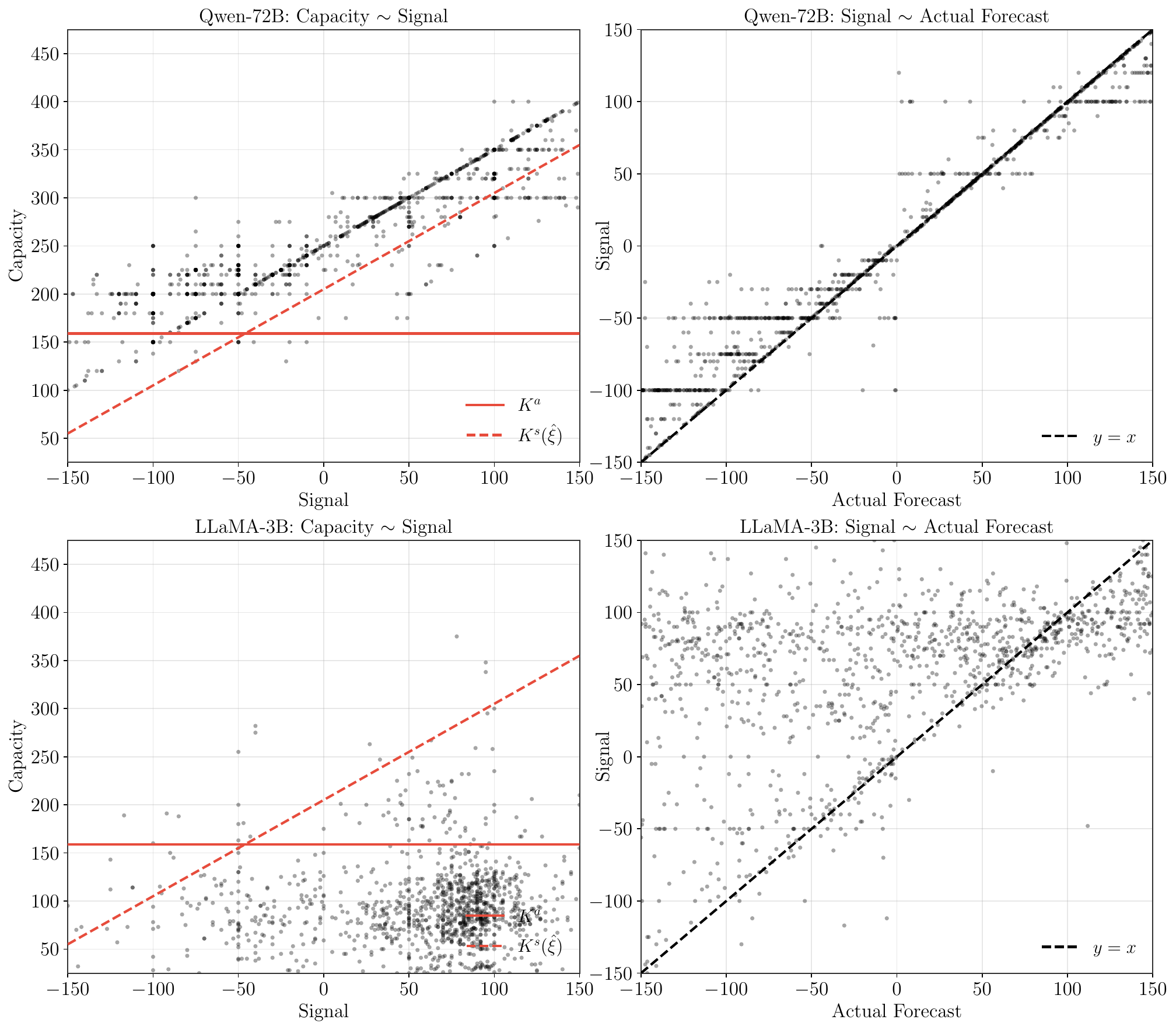}
}
{{ Agent-Level Behavior in the Experiment of \citet{doi:10.1287/mnsc.1110.1334}: Qwen-72B versus LLaMA-3B} \label{r9_scatter_plot}}
{{ The left panel shows the Supplier's capacity decision $K$ as a function of the received signal $\hat{\xi}$. The right panel shows the Manufacturer's reported signal $\hat{\xi}$ as a function of the true forecast $\xi$. The line $K^s(\hat{\xi})$ is the capacity the Supplier would choose under full trust in the Manufacturer's report. The line $K^a$ is the optimal capacity under an uninformative equilibrium, in which the Supplier places no trust in the Manufacturer's report. Specifically, $K^a=\mu+(F\circ G)^{-1}\left((w-c-c_K)/(w-c)\right)$, where $F\circ G$ is the cumulative distribution function of $\xi+\epsilon$.}}
\end{figure}

}

{

\section{Lightweight Behavioral Alignment Protocol} \label{intervention_process}

We clarify the lightweight intervention stage through a structured and reproducible protocol, which we refer to as the \emph{Lightweight Behavioral Alignment Protocol}. The protocol consists of five steps: (1) data processing and output standardization, (2) method selection for distributional alignment, (3) intervention design, and (4) robustness testing and ablation. This protocol applies to both CoT prompting and hyperparameter tuning, and is intended to make the intervention stage transparent, replicable, and comparable across experiments.

\subsection{Data Processing and Output Standardization}

To ensure that LLM responses can be processed consistently across experiments, we impose a maximum generation length and require structured outputs. Specifically, the model is instructed to prepend its final decision with the marker $\#\#\#$, which allows a customized parser to automatically identify and extract the final answer for downstream statistical analysis. For numerical decisions, such as order quantities, bids, prices, forecasts, or capacity choices, the parser extracts the numerical value following the marker and maps it to the corresponding decision variable in the original experiment.

If the parser fails to detect a valid numerical token after the marker, the system triggers a fallback procedure and re-prompts the model up to three times. If all three attempts fail, we treat the generation as unusable. In the hyperparameter-tuning analysis, such failures indicate that the current combination of temperature, sampling method, and sampling threshold induces unstable or excessively noisy outputs. We therefore exclude the configuration from the feasible evaluation set and, in Bayesian optimization, assign it a pessimistic penalty to discourage further exploration of unstable regions of the search space.

For decisions that are not naturally expressed as continuous numerical values, we convert responses into predefined machine-readable formats. For example, in the queueing experiment, where the agent decides whether to join the queue, the LLM is instructed to output $\#\#\# 1$ for joining and $\#\#\# 0$ for not joining. Similarly, in the contracting experiment, where the retailer decides whether to accept the manufacturer's offer, the LLM outputs the relevant numerical decision if the contract is accepted; if the contract is rejected, it outputs only the marker $\#\#\#$ followed by a blank field, which is recorded as a rejection in the processing pipeline. This procedure allows both numerical and non-numerical decisions to be converted into standardized representations while preserving the structure of the original experimental task.

\subsection{Method Selection for Distributional Alignment}

We use Wasserstein distance as the primary metric for distributional alignment because it respects the geometry of the decision space. Unlike information-theoretic measures such as KL divergence or cross-entropy, which typically require binning and treat response values as distinct categories, Wasserstein distance accounts for the magnitude of differences between decisions. For example, a simulated order quantity that differs from a human response by one unit is treated as closer than one that differs by ten units. This feature is especially important in our setting, where many behavioral outcomes are ordered numerical decisions, such as prices, quantities, bids, capacities, and forecasts.

At the same time, Wasserstein distance is not used as a black-box criterion. To diagnose the sources of misalignment, we also decompose the distributional gap into mean mismatch and dispersion mismatch. This decomposition allows us to distinguish whether an intervention improves alignment by shifting the center of the LLM response distribution, by changing its variability, or by affecting both. Such diagnostics are important because an intervention can reduce average bias while leaving the model unable to reproduce the dispersion of human behavior.

\subsection{Intervention Design}

Our protocol evaluates two lightweight intervention strategies: CoT prompting and hyperparameter tuning. We intentionally analyze these strategies separately rather than combining them in the main analysis, so that the marginal effect of each intervention can be identified without confounding from interaction effects.

CoT prompting modifies the reasoning process before the model generates a decision. We consider both simple CoT, which adds a generic reasoning instruction, and structured CoT, which imposes a stage-based reasoning framework tailored to the decision environment. Hyperparameter tuning, by contrast, modifies the sampling behavior of the model without changing the prompt structure or model weights. We tune generation parameters such as temperature, top-$p$, min-$p$, and top-$k$, which affect the concentration, dispersion, and extremity of generated responses.

\subsection{Robustness Testing and Ablation}

We conduct robustness and ablation analyses to assess whether the observed improvements are stable and interpretable. For CoT prompting, the analysis contains an internal ablation design. Beyond comparing the baseline, simple CoT, and structured CoT settings, we conduct stage-ablation analysis by removing or perturbing specific reasoning stages and then re-evaluating the resulting change in Wasserstein alignment, mean mismatch, and dispersion mismatch. This design identifies which components of the reasoning chain contribute positively or negatively to alignment with human behavior.

For hyperparameter tuning, robustness is supported through repeated evaluations and explicit treatment of unstable generations. In particular, Bayesian optimization evaluates each hyperparameter configuration over three independent runs and models both the sample mean and the associated noise. Failed configurations are penalized rather than silently discarded. This prevents the tuning procedure from favoring configurations that appear favorable only because they produce incomplete, invalid, or unstable outputs.

Together, these steps provide a transparent framework for implementing lightweight alignment interventions. The protocol clarifies how LLM outputs are generated, parsed, filtered, evaluated, and improved. It also separates two distinct mechanisms of intervention: CoT prompting changes the reasoning trace that precedes a decision, whereas hyperparameter tuning changes the sampling distribution from which the final response is drawn. Both approaches preserve the underlying model weights, making them feasible in behavioral OM settings where human data are limited and full fine-tuning is often impractical.

}

\clearpage
\section{Additional Results}
\subsection{Simulation with Different LLMs}\label{subsec:sim-diff-llms}

Table \ref{tab:wass_llms} showcases the Wasserstein distance by 11 different LLMs. For visual representation, please refer to Figure \ref{LLMs_wass_ratio}.

\vspace{0.5cm}
\begin{table}[htp!]
    \centering
    \setlength{\abovecaptionskip}{1pt}
    \setlength{\belowcaptionskip}{5pt}
    \vspace{-10.0pt}
    \TABLE
    {Wasserstein Distance with Different LLMs
    \label{tab:wass_llms}}
    {\setlength{\tabcolsep}{1.2mm}{
    \resizebox{\columnwidth}{0.3\textheight}{
    \begin{tabular}{lllcccccccccccccc}
    \toprule
& & \textbf{Deepseek}  & \multicolumn{3}{c}{\textbf{GPT}} & & \multicolumn{3}{c}{\textbf{LLaMA}} & & \multicolumn{4}{c}{\textbf{Qwen}}&  \\
\cline{4-6} \cline{8-10} \cline{12-16}
 Number & Paper/hypothesis & &  3.5 & 4 & 4o mini & & 3B & 8B & 70B  & & 7B & 14B & 32B & 72B \\
    \midrule
    1 & \cite{doi:10.1287/mnsc.1120.1638} &  &  &  & &  &   &   &   &  &   &   &   &  \\
& $\quad$  C & 2.17  & 1.91  & 1.07  & 1.25  &  & 1.59  & 2.28  & 2.15  &  & 1.26  & 0.99  & 2.48  & 1.26  \\ 
& $\quad$  O & 2.27  & 1.04  & 0.78  & 2.04  &  & 2.32  & 2.99  & 2.34  &  & 1.25  & 1.54  & 1.67  & 1.58  \\ 

2 & \cite{doi:10.1287/mnsc.1050.0436} &  &  &  & &  &   &   &   &  &   &   &   &  \\
& $\quad$  Base & 7.47   & 7.45   & 7.52   & 7.97   &  & 5.51   & 23.41   & 8.31   &  & 13.25   & 8.18   & 8.0   & 6.97   \\ 
& $\quad$  Share & 8.83   & 9.24   & 9.02   & 9.69   &  & 8.31   &  EF  & 9.7   &  &  EF  & 9.62   & 9.26   & 10.11   \\ 

3 & \cite{doi:10.1287/mnsc.1100.1258}  &  &  &  & &  &   &   &   &  &   &   &   & \\
& $\quad$ 1 & 11.64   & 38.6   & 13.51   & 13.47   &  &  EF  & 18.56   & 8.5   &  & 17.61   & 10.1   & 9.2   & 8.72   \\ 
& $\quad$ 4 & 10.52   & 34.69   & 10.11   & 11.02   &  &  EF  & 13.53   & 9.7   &  & 9.38   & 13.99   & 9.86   & 11.88   \\ 
& $\quad$ 7 & 17.84   & 24.6   & 14.81   & 13.98   &  &  EF  & 7.4   & 16.73   &  & 8.06   & 24.26   & 17.99   & 10.18   \\ 
& $\quad$ 10 & 24.95   & 17.37   & 21.09   & 20.54   &  &  EF  & 9.74   & 23.87   &  & 13.08   & 30.88   & 24.23   & 12.21   \\ 

4 & \hyperref[doi:10.1287/mnsc.1070.0806]{Engelbrecht-Wiggans} &  &  &  &   &   &   &   &  & &   &   &   & \\\n  & \hyperref[doi:10.1287/mnsc.1070.0806]{and Katok (2008)} & &  &  &  &   &   &   &  &   &   &   & &   \\
& $\quad$  both & 13.35   & 21.26   & 11.33   & 8.78   &  & 18.47   & 6.98   & 16.77   &  & 19.15   & 16.51   & 18.14   & 8.68   \\ 
& $\quad$  loser & 12.84   & 19.03   & 12.16   & 6.54   &  & 19.35   & 9.1   & 15.64   &  & 14.65   & 16.29   & 17.33   & 16.9   \\ 

5 & \cite{doi:10.1287/mnsc.1070.0788} &  &  &  &   &   &   &   &  &   &   &   & & \\
& $\quad$ qd x & 1.36   & 2.05   & 0.42   & 2.19   &  & 1.29   & 2.09   & 1.22   &  & 1.04   & 1.19   & 1.49   & 0.97   \\ 
& $\quad$ qd y & 5.17   & 3.51   & 13.28   & 4.84   &  & 0.73   & 2.79   & 4.31   &  & 1.86   & 3.8   & 7.26   & 0.62   \\ 
& $\quad$ qd price b & 1.59   & 0.51   & 1.08   & 0.56   &  & 0.62   & 1.0   & 1.04   &  &  Rej  & 1.41   & 2.06   & 0.88   \\ 
& $\quad$ tpt fixed fee & 6.9   & 4.3   & 4.92   & 3.72   &  & 2.24   & 1.48   & 2.36   &  & 3.72   & 0.67   & 1.3   & 0.78   \\ 
& $\quad$ tpt price a & 1.03   & 0.37   & 1.23   & 0.39   &  & 2.39   & 1.41   & 0.73   &  & 1.22   & 1.34   & 0.49   & 0.34   \\ 
& $\quad$ tpt price b & 1.17   & 0.98   & 0.98   & 1.13   &  & 0.78   & 0.72   & 0.77   &  &  Rej  & 2.04   & 0.68   & 1.31   \\ 

6 & \cite{doi:10.1287/mnsc.2015.2264} &  &  &  &   &   &   &  &   &   &   & &  &    \\
& $\quad$  q00 & 0.06   & 0.36   & 0.3   & 0.35   &  & 0.38   & 0.25   & 0.19   &  & 0.01   & 0.55   & 0.2   & 0.3   \\ & $\quad$  q50 & 0.18   & 0.19   & 0.22   & 0.18   &  & 0.48   & 0.05   & 0.04   &  & 0.28   & 0.08   & 0.16   & 0.04   \\ 

7 & \cite{doi:10.1287/mnsc.1110.1382} &  &  &  & &  &   &   &   &  &   &   &   & \\
& $\quad$  c1 & 4.7   & 5.47   & 5.37   & 5.4   &  & 11.92   & 4.18   & 4.56   &  & 4.52   & 5.22   & 3.75   & 3.63   \\ 
& $\quad$  c5 & 27.27   & 43.43   & 20.92   & 45.89   &  & 126.25   & 47.18   & 22.49   &  & 32.75   & 32.75   & 14.47   & 17.15   \\ 

8 & \cite{doi:10.1287/mnsc.46.3.404.12070} &  &  &  & &  &   &   &   &  &   &   &   & \\
& $\quad$ Low Profit   & 39.77   & 25.77   & 55.95   & 24.09   &  & 64.75   & 25.12   & 28.36   &  & 65.24   & 103.86   & 34.42   & 29.38   \\ 
& $\quad$ High Profit   & 58.9   & 48.89   & 84.85   & 71.64   &  & 107.98   & 28.4   & 35.05   &  & 43.29   & 59.66   & 58.27   & 39.88   \\ 

9 & \cite{doi:10.1287/mnsc.1110.1334} &  &  &  &   &   &   & &   &   &   &  &  & \\
& $\quad$  signal & 33.85   & 26.99   & 39.89   & 30.12   &  & 22.74   & 31.72   & 30.7   &  & 33.13   & 32.63   & 33.76   & 37.54   \\ 
& $\quad$  capacity & 84.36   & 46.32   & 42.17   & 44.01   &  & 114.73   & 253.59   & 36.75   &  & 38.68   & 77.98   & 39.83   & 47.05   \\ 

    \bottomrule

    \end{tabular}
    }
    }}{\emph{Note.} Rej denotes that all of the price given are rejected, so no numerical price provided. EF denotes experiment failure due to no valid result collected within the number of answer limit.}
    \end{table}

\newpage
\subsection{Simulation with and without CoT}\label{appendix_cot}
Beyond the visual illustration in Figures \ref{LLMs_cot_improve} and \ref{LLMs_change_r1}, the following provides a comprehensive comparison of the data between with and without CoT across three different temperatures (0.5, 1.0, and 1.5) using LLaMA-70B. It can be observed that, in general, the Wasserstein distance with CoT is smaller than that without it. This observation offers additional supporting evidence beyond the information provided in Section \ref{COT}.
\begin{table}[htp!]
\vspace{-5mm}
    \centering

    \TABLE{Wasserstein Distance by LLaMA-70B with and without CoT}{

    \resizebox*{\columnwidth}{0.7\textheight}{
    \begin{tabular}{lllcBcccBcccBcccc}
\toprule
& & \multicolumn{3}{c}{\textbf{LLaMA} 0.5} & & \multicolumn{3}{c}{\textbf{LLaMA} 1.0} & & \multicolumn{3}{c}{\textbf{LLaMA} 1.5} \\
\cline{3-5} \cline{7-9} \cline{11-13}
Number & Paper/hypothesis
& Without & CoT & \makecell[c]{CoT Stage\\[-2pt]Analysis}
& & Without & CoT & \makecell[c]{CoT Stage\\[-2pt]Analysis}
& & Without & CoT & \makecell[c]{CoT Stage\\[-2pt]Analysis} \\
\midrule
    1 & \cite{doi:10.1287/mnsc.1120.1638} &  &  &  & &  &   &   &   &   &   &   &   \\
& $\quad$  C & 1.07 & 0.69 & 0.67 &  & 1.97 & 0.62 & 0.72 &  & 0.91 & 0.54 & 0.54  \\ 
& $\quad$  O & 1.51 & 1.59 & 1.79 &  & 1.85 & 1.47 & 1.7 &  & 1.49 & 1.26 & 1.78  \\ 

2 & \cite{doi:10.1287/mnsc.1050.0436} &  &  &  & &  &   &   &   &   &   &   &   \\
& $\quad$  Base & 3.88   & 3.78 & 3.87   &  & 3.88   & 3.83 & 3.85   &  & 3.79   & 3.86 & 3.85   \\ 
& $\quad$  Share & 3.88   & 3.68 & 3.87   &  & 3.88   & 3.8 & 3.84   &  & 3.29   & 3.75 & 3.75   \\ 

3 & \cite{doi:10.1287/mnsc.1100.1258}  &  &  &  & &  &   &   &   &   &   &   &   \\
& $\quad$ 1 & 7.42 & 10.92 & 8.33 &  & 7.95 & 10.97 & 8.39 &  & \num{46514.07} & 9.49 & 8.03  \\ 
& $\quad$ 4 & 10.51 & 10.1 & 9.61 &  & 10.72 & 9.53 & 10.14 &  & \num{1241396.43} & 8.98 & 9.75  \\ 
& $\quad$ 7 & 17.15 & 17.24 & 16.54 &  & 17.4 & 16.67 & 16.82 &  & \num{5.99e+20}   & 16.24 & 16.56  \\ 
& $\quad$ 10 & 23.8 & 23.36 & 22.09 &  & 23.71 & 23.19 & 22.71 &  & \num{54330.64} & 23.23 & 22.5  \\ 

4 & \cite{doi:10.1287/mnsc.1070.0806} &  &  &  & &  &   &   &   &   &   &   &   \\

& $\quad$  both & 16.84   & 14.93 & 16.69   &  & 17.03   & 15.34 & 16.34   &  & 16.91   & 15.95 & 16.37  \\ 
& $\quad$  loser & 15.33   & 14.66 & 15.78   &  & 15.56   & 15.42 & 15.51   &  & 15.7   & 15.22 & 15.71  \\ 

5 & \cite{doi:10.1287/mnsc.1070.0788} &  &  &  & &  &   &   &   &   &   &   &   \\
& $\quad$ qd x & 1.35 & 0.23 & 2.31 &  & 1.14 & 0.24 & 1.84 &  & 0.89 & 0.2 & 1.34  \\ 
& $\quad$ qd y & 4.77 & 2.74 & 4.84 &  & 4.17 & 2.68 & 4.97 &  & 56.87 & 2.62 & 4.79  \\ 
& $\quad$ qd price b & 1.34 & 0.83 & 1.6 &  & 0.9 & 0.83 & 1.32 &  & 0.54 & 0.77 & 1.04  \\ 
& $\quad$ tpt fixed fee & 2.3 & 3.81 & 3.83 &  & 2.27 & 3.64 & 3.76 &  & \num{7819.15} & 3.43 & 3.65  \\ 
& $\quad$ tpt price a & 0.8 & 0.71 & 0.77 &  & 0.68 & 0.42 & 0.72 &  & 0.59 & 0.13 & 0.74  \\ 
& $\quad$ tpt price b & 1.02 & 1.34 & 1.35 &  & 0.73 & 1.13 & 1.36 &  & 0.55 & 0.98 & 1.16  \\  

6 & \cite{doi:10.1287/mnsc.2015.2264} &  &  &  & &  &   &   &   &   &   &   &   \\
& $\quad$  q00 & 0.22   & 0.26  & 0.04 &  & 0.17   & 0.26 & 0.16  &  & 0.17   & 0.21  & 0.13 \\ 
& $\quad$  q50 & 0.06   & 0.08  & 0.04 &  & 0.03   & 0.05  & 0.01 &  & 0.02   & 0.04  & 0.01 \\ 

7 & \cite{doi:10.1287/mnsc.1110.1382} &  &  &  & &  &   &   &   &   &   &   &   \\
& $\quad$  c1 & 4.42 & 4.29 & 4.44 &  & 4.52 & 4.39 & 4.44 &  & \num{1470779217490.35} & 4.3 & 4.64  \\ 
& $\quad$  c5 & 21.21 & 11.91 & 14.65 &  & 19.21 & 11.98 & 14.41 &  & \num{2556817554620.06} & 12.22 & 13.1  \\  

8 & \cite{doi:10.1287/mnsc.46.3.404.12070} &  &  &  & &  &   &   &   &   &   &   &   \\
& $\quad$ Low Profit   & 63.33 & 44.86 & 36.17 &  & 27.06 & 31.23 & 35.79 &  & 76.55 & 30.78 & 32.89  \\ 
& $\quad$ High Profit   & 33.34 & 26.91 & 48.73 &  & 34.3 & 46.05 & 28.56 &  & 37.23 & 29.72 & 35.3  \\  

9 & \cite{doi:10.1287/mnsc.1110.1334} &  &  &  & &  &   &   &   &   &   &   &   \\
& $\quad$  signal & 13.27 & 16.56 & 22.9 &  & 16.8 & 16.26 & 23.15 &  & 14.02 & 15.67 & 16.42  \\ 
& $\quad$  capacity & 31.36 & 28.49 & 26.59 &  & 25.68 & 24.21 & 28.38 &  & 23.16 & 26.45 & 28.45  \\ 
    \bottomrule
    \end{tabular}
    }
    }

    {\emph{Note.} LLaMA 0.5, LLaMA 1.0, and LLaMA 1.5 refer to LLaMA-70B evaluated at three different 
    temperature settings. ``Without'' denotes responses generated without CoT reasoning, 
    while ``CoT'' denotes responses generated with CoT reasoning and ``CoT Stage Analysis'' denotes the structured CoT setting with the stage-based analysis.
    }

    \end{table}

\newpage
\subsection{Simulation with Different Temperatures} \label{subsec:sim-diff-temp}
Tables \ref{tab:wass_qwen7b}, 

and \ref{tab:wass_qwen32b} 

showcase the Wasserstein distance using Qwen-7B and Qwen-32B respectively after temperature and sampling methods tuning, providing more detailed data support for Section \ref{temperature_tuning} of the main text. The sampling method (within the parentheses) corresponds to the sampling method that achieves the minimal Wasserstein distance under the same temperature setting. 

\begin{table}[htp!]
    \centering
    \caption{{Wasserstein Distance under Different Temperatures by Qwen-7B}}
    \label{tab:wass_qwen7b}
    \setlength{\tabcolsep}{1.2mm}{
    \resizebox*{\columnwidth}{!}{
    \begin{tabular}{lllccccccccc}
    \toprule
 Number & Paper/Feature & 0.5 & 1 & 1.5 & 1.7 & 1.9 & 2.5 & 3.0 & 3.5 & 4.0 \\
    \midrule
1 & \cite{doi:10.1287/mnsc.1120.1638} &  &  &  &  &  &  &  &  &  \\
 & $\quad$  C & 1.32 & 0.98 & 0.58 & 0.41 & 0.60 & 0.67 & 0.67 & 0.73 & 0.57 \\
 &  & (Top-$p$ 1.0) & (Top-$k$ 65) & (Top-$k$ 88) & (Top-$k$ 55) & (Top-$k$ 14) & (Min-$p$ 0.11) & (Min-$p$ 0.16) & (Min-$p$ 0.16) & (Min-$p$ 0.20) \\
 & $\quad$  O & 1.02 & 0.87 & 0.59 & 0.59 & 0.63 & 0.78 & 0.68 & 0.59 & 0.60 \\
 &  & (Top-$k$ 41) & (Top-$k$ 65) & (Top-$k$ 47) & (Top-$k$ 19) & (Top-$k$ 42) & (Min-$p$ 0.16) & (Min-$p$ 0.11) & (Min-$p$ 0.16) & (Min-$p$ 0.20) \\
2 & \cite{doi:10.1287/mnsc.1050.0436} &  &  &  &  &  &  &  &  &  \\
 & $\quad$  Base & 8.10 & 12.95 & 8.88 & 24.62 & 15.13 & 52.71 & 190.09 & 589.68 & 32282.48 \\
 &  & (Top-$k$ 29) & (Top-$k$ 68) & (Top-$p$ 0.50) & (Min-$p$ 0.16) & (Min-$p$ 0.16) & (Min-$p$ 0.16) & (Min-$p$ 0.16) & (Min-$p$ 0.16) & (Min-$p$ 0.18) \\
 & $\quad$  Share & 275.53 & 987.68 & 604.80 & 984.39 & 599.74 & 14.02 & 28709.36 & 138604.35 & 1459503.90 \\
 &  & (Top-$p$ 0.59) & (Top-$p$ 0.59) & (Top-$k$ 62) & (Top-$p$ 1.0) & (Top-$p$ 0.81) & (Top-$p$ 0.59) & (Min-$p$ 0.16) & (Min-$p$ 0.16) & (Min-$p$ 0.18) \\
3 & \cite{doi:10.1287/mnsc.1100.1258} &  &  &  &  &  &  &  &  &  \\
 & $\quad$  1 & 16.79 & 16.95 & 18.90 & 18.74 & 18.23 & 4.20 & 10.80 & 1411.31 & 20.50 \\
 &  & (Top-$p$ 0.76) & (Top-$k$ 13) & (Top-$p$ 0.57) & (Top-$k$ 10) & (Top-$p$ 0.50) & (Top-$p$ 0.75) & (Top-$p$ 0.50) & (Top-$k$ 23) & (Top-$k$ 32) \\
 & $\quad$  4 & 8.20 & 8.30 & 9.30 & 9.11 & 9.14 & 3.99 & 14.00 & 14.69 & 16.58 \\
 &  & (Min-$p$ 0.16) & (Min-$p$ 0.16) & (Top-$p$ 0.65) & (Top-$p$ 0.50) & (Top-$p$ 0.65) & (Top-$p$ 0.78) & (Min-$p$ 0.11) & (Min-$p$ 0.16) & (Min-$p$ 0.16) \\
 & $\quad$  7 & 7.46 & 7.04 & 6.76 & 5.81 & 5.37 & 5.78 & 6.59 & 5.29 & 8.37 \\
 &  & (Top-$k$ 13) & (Top-$k$ 47) & (Top-$k$ 51) & (Top-$k$ 61) & (Top-$p$ 1.0) & (Top-$p$ 0.59) & (Min-$p$ 0.11) & (Top-$k$ 63) & (Min-$p$ 0.16) \\
 & $\quad$  10 & 11.87 & 10.52 & 8.97 & 8.25 & 7.42 & 6.18 & 4.15 & 3.76 & 6.72 \\
 &  & (Top-$k$ 13) & (Top-$k$ 67) & (Top-$k$ 38) & (Top-$k$ 23) & (Top-$k$ 89) & (Top-$k$ 66) & (Top-$k$ 66) & (Top-$k$ 51) & (Top-$k$ 13) \\
4 & \cite{doi:10.1287/mnsc.1070.0806} &  &  &  &  &  &  &  &  &  \\
 & $\quad$  both & 17.77 & 18.00 & 17.94 & 16.72 & 16.93 & 14.33 & 17.97 & 12.78 & 13.78 \\
 &  & (Top-$p$ 0.77) & (Top-$p$ 0.65) & (Top-$p$ 1.0) & (Top-$k$ 55) & (Top-$k$ 14) & (Top-$k$ 26) & (Min-$p$ 0.11) & (Min-$p$ 0.05) & (Min-$p$ 0.11) \\
 & $\quad$  loser & 15.21 & 16.00 & 15.80 & 15.92 & 15.18 & 14.23 & 16.93 & 15.08 & 14.33 \\
 &  & (Min-$p$ 0.16) & (Top-$p$ 0.81) & (Min-$p$ 0.16) & (Top-$p$ 0.59) & (Top-$k$ 13) & (Top-$k$ 13) & (Min-$p$ 0.16) & (Min-$p$ 0.09) & (Min-$p$ 0.11) \\
5 & \cite{doi:10.1287/mnsc.1070.0788} &  &  &  &  &  &  &  &  &  \\
 & $\quad$  qd x & 1.43 & 1.12 & 0.69 & 0.58 & 0.53 & 0.58 & 0.63 & 0.67 & 0.69 \\
 &  & (Top-$k$ 40) & (Top-$k$ 66) & (Top-$k$ 88) & (Top-$k$ 19) & (Top-$p$ 0.94) & (Min-$p$ 0.11) & (Min-$p$ 0.16) & (Min-$p$ 0.20) & (Min-$p$ 0.20) \\
 & $\quad$  qd y & 2.22 & 1.60 & 1.01 & 0.84 & 0.70 & 0.68 & 0.71 & 0.70 & 0.70 \\
 &  & (Top-$k$ 40) & (Top-$k$ 32) & (Top-$k$ 43) & (Top-$k$ 25) & (Top-$k$ 43) & (Min-$p$ 0.05) & (Min-$p$ 0.05) & (Min-$p$ 0.16) & (Min-$p$ 0.20) \\
 & $\quad$  tpt fixed fee & 3.66 & 3.64 & 3.34 & 3.24 & 3.06 & 2.92 & 3.14 & 3.09 & 3.53 \\
 &  & (Top-$p$ 0.93) & (Top-$k$ 11) & (Top-$k$ 36) & (Top-$p$ 1.0) & (Top-$p$ 1.0) & (Top-$p$ 0.65) & (Min-$p$ 0.11) & (Min-$p$ 0.16) & (Min-$p$ 0.16) \\
 & $\quad$  tpt price a & 1.64 & 1.21 & 0.99 & 1.00 & 0.93 & 0.99 & 0.99 & 0.95 & 1.04 \\
 &  & (Top-$k$ 37) & (Top-$k$ 36) & (Top-$k$ 45) & (Top-$k$ 65) & (Min-$p$ 0.16) & (Top-$k$ 27) & (Top-$k$ 47) & (Min-$p$ 0.16) & (Min-$p$ 0.11) \\

6 & \cite{doi:10.1287/mnsc.2015.2264} &  &  &  &  &  &  &  &  &  \\
 & $\quad$  q00 & 0.14 & 0.02 & 0.02 & 0.05 & 0.11 & - & - & - & - \\
 &  & (Top-$p$ 1.0) & (Top-$k$ 27) & (Top-$k$ 16) & (Top-$p$ 1.0) & (Top-$p$ 0.65) &  &  &  &  \\
 & $\quad$  q50 & 0.21 & 0.22 & 0.22 & 0.22 & 0.22 & - & - & - & - \\
 &  & (Min-$p$ 0.16) & (Top-$p$ 0.50) & (Top-$p$ 0.59) & (Top-$p$ 0.50) & (Top-$p$ 0.51) &  &  &  &  \\
7 & \cite{doi:10.1287/mnsc.1110.1382} &  &  &  &  &  &  &  &  &  \\
 & $\quad$  c1 & 4.77 & 4.53 & 4.28 & 3.98 & 3.93 & 4.23 & 3.53 & 2.72 & 2.61 \\
 &  & (Top-$k$ 42) & (Top-$k$ 78) & (Top-$k$ 71) & (Top-$p$ 1.0) & (Top-$k$ 16) & (Top-$p$ 0.56) & (Min-$p$ 0.05) & (Min-$p$ 0.05) & (Min-$p$ 0.07) \\
 & $\quad$  c5 & 40.90 & 40.10 & 41.50 & 42.21 & 40.09 & 44.04 & 45.77 & 49.93 & 52.57 \\
 &  & (Top-$k$ 11) & (Min-$p$ 0.16) & (Top-$p$ 0.81) & (Min-$p$ 0.16) & (Top-$p$ 1.0) & (Top-$k$ 76) & (Min-$p$ 0.20) & (Min-$p$ 0.11) & (Min-$p$ 0.16) \\
8 & \cite{doi:10.1287/mnsc.46.3.404.12070} &  &  &  &  &  &  &  &  &  \\
 & $\quad$  Low Profit & 58.58 & 56.60 & 56.49 & 51.78 & 52.66 & 51.80 & 50.71 & 54.08 & 53.11 \\
 &  & (Top-$k$ 40) & (Min-$p$ 0.11) & (Top-$k$ 25) & (Top-$k$ 54) & (Top-$k$ 20) & (Min-$p$ 0.11) & (Top-$k$ 63) & (Min-$p$ 0.11) & (Min-$p$ 0.11) \\
 & $\quad$  High Profit & 35.45 & 36.75 & 32.53 & 32.29 & 22.76 & 24.08 & 17.34 & 21.27 & 17.41 \\
 &  & (Top-$k$ 40) & (Min-$p$ 0.11) & (Top-$p$ 0.81) & (Top-$p$ 1.0) & (Top-$p$ 1.0) & (Top-$k$ 27) & (Top-$k$ 73) & (Min-$p$ 0.11) & (Min-$p$ 0.11) \\
9 & \cite{doi:10.1287/mnsc.1110.1334} &  &  &  &  &  &  &  &  &  \\
 & $\quad$ signal & 28.19 & 27.03 & 28.21 & 30.84 & 30.82 & - & 48.95 & 58.30 & -\\
 &  & (Top-$k$ 39) & (Top-$k$ 74) & (Top-$p$ 0.65) & (Top-$p$ 0.65) & (Top-$p$ 0.65) & & (Min-$p$ 0.16) & (Min-$p$ 0.16) &  \\
 & $\quad$  capacity & 44.32 & 35.88 & 28.46 & 23.03 & 20.72 & - & 23.45 & 13.20 & - \\
 &  & (Top-$k$ 67) & (Top-$k$ 19) & (Top-$k$ 10) & (Top-$k$ 33) & (Top-$k$ 32) &  & (Min-$p$ 0.16) & (Min-$p$ 0.16) &  \\
    \bottomrule
    \end{tabular}
    }
    }
    \end{table}

\begin{table}[htp!]
    \centering
    \caption{{Wasserstein Distance under Different Temperatures by Qwen-32B}}
    \label{tab:wass_qwen32b}
    \setlength{\tabcolsep}{1.2mm}{
    \resizebox*{\columnwidth}{!}{
    \begin{tabular}{lllccccccccccc}
    \toprule
 Number & Paper/Feature & 0.5 & 1 & 1.5 & 1.7 & 1.9 & 2.5 & 3.0 & 3.5 & 4.0 \\ 
    \midrule
1 & \cite{doi:10.1287/mnsc.1120.1638} &  &  &  &  &  &  &  &  &  \\
 & $\quad$  C & 3.06 & 2.30 & 2.27 & 2.15 & 2.07 & 2.22 & 2.09 & 1.75 & 1.83 \\
 &  & (Top-$k$ 32) & (Top-$k$ 62) & (Top-$k$ 80) & (Top-$k$ 69) & (Top-$k$ 70) & (Min-$p$ 0.16) & (Min-$p$ 0.05) & (Min-$p$ 0.05) & (Min-$p$ 0.16) \\
 & $\quad$  O & 1.88 & 1.89 & 1.94 & 1.95 & 1.89 & 2.12 & 2.12 & - & - \\
 &  & (Top-$p$ 0.59) & (Top-$p$ 0.65) & (Top-$k$ 74) & (Top-$p$ 0.50) & (Top-$p$ 0.59) & (Top-$k$ 12) & (Min-$p$ 0.16) &  &  \\
2 & \cite{doi:10.1287/mnsc.1050.0436} &  &  &  &  &  &  &  &  &  \\
 & $\quad$  Base & 8.04 & 7.95 & 7.77 & 7.61 & 7.47 & 6.64 & 6.98 & 5.93 & 5.28 \\
 &  & (Top-$k$ 32) & (Top-$k$ 64) & (Top-$k$ 85) & (Top-$k$ 18) & (Top-$k$ 66) & (Top-$k$ 12) & (Min-$p$ 0.07) & (Min-$p$ 0.07) & (Min-$p$ 0.11) \\
 & $\quad$  Share & 9.24 & 8.87 & 7.26 & 6.69 & 6.10 & 6.97 & 6.93 & 27.69 & 134.67 \\
 &  & (Top-$k$ 32) & (Top-$k$ 66) & (Top-$k$ 90) & (Top-$k$ 20) & (Top-$k$ 67) & (Min-$p$ 0.16) & (Min-$p$ 0.16) & (Min-$p$ 0.16) & (Min-$p$ 0.20) \\
3 & \cite{doi:10.1287/mnsc.1100.1258} &  &  &  &  &  &  &  &  &  \\
 & $\quad$  1 & 7.68 & 7.99 & 7.33 & 7.22 & 6.88 & 5.64 & 4.84 & 3.74 & 3.16 \\
 &  & (Min-$p$ 0.11) & (Min-$p$ 0.05) & (Top-$p$ 1.0) & (Top-$k$ 55) & (Top-$k$ 17) & (Top-$k$ 39) & (Top-$k$ 28) & (Top-$k$ 32) & (Top-$k$ 16) \\
 & $\quad$  4 & 10.04 & 9.63 & 9.15 & 9.14 & 8.83 & 8.24 & 7.24 & 6.48 & 5.27 \\
 &  & (Top-$k$ 40) & (Top-$p$ 0.81) & (Top-$p$ 1.0) & (Top-$k$ 66) & (Top-$k$ 16) & (Top-$k$ 49) & (Top-$k$ 41) & (Top-$k$ 11) & (Top-$k$ 66) \\
 & $\quad$  7 & 17.69 & 17.22 & 16.49 & 16.69 & 16.33 & 15.50 & 13.78 & 12.69 & 11.34 \\
 &  & (Top-$p$ 0.81) & (Top-$k$ 16) & (Top-$k$ 40) & (Top-$k$ 90) & (Top-$k$ 32) & (Top-$k$ 65) & (Top-$k$ 19) & (Top-$k$ 23) & (Top-$k$ 21) \\
 & $\quad$  10 & 23.56 & 23.60 & 22.53 & 22.80 & 22.32 & 21.09 & 19.23 & 17.74 & 16.01 \\
 &  & (Top-$p$ 0.81) & (Min-$p$ 0.11) & (Top-$k$ 60) & (Top-$k$ 46) & (Top-$k$ 18) & (Top-$k$ 73) & (Top-$k$ 26) & (Top-$k$ 23) & (Top-$k$ 12) \\
4 & \cite{doi:10.1287/mnsc.1070.0806} &  &  &  &  &  &  &  &  &  \\
 & $\quad$  both & 17.65 & 17.65 & 17.72 & 17.36 & 17.08 & 16.64 & 16.37 & 16.21 & 16.15 \\
 &  & (Min-$p$ 0.16) & (Top-$p$ 0.50) & (Top-$p$ 1.0) & (Top-$p$ 1.0) & (Top-$p$ 1.0) & (Top-$k$ 33) & (Top-$k$ 65) & (Top-$k$ 25) & (Min-$p$ 0.05) \\
 & $\quad$  loser & 17.08 & 17.03 & 16.99 & 16.81 & 16.76 & 16.57 & 14.82 & 15.67 & 15.34 \\
 &  & (Top-$p$ 0.59) & (Top-$p$ 0.59) & (Top-$k$ 85) & (Top-$k$ 19) & (Top-$k$ 66) & (Top-$k$ 40) & (Top-$p$ 0.84) & (Top-$k$ 88) & (Top-$k$ 38) \\

5 & \cite{doi:10.1287/mnsc.1070.0788} &  &  &  &  &  &  &  &  &  \\
 & $\quad$  qd x & 1.53 & 1.43 & 1.33 & 1.28 & 1.23 & 1.08 & 0.87 & 0.76 & 0.65 \\
 &  & (Top-$k$ 10) & (Top-$p$ 1.0) & (Top-$p$ 1.0) & (Top-$k$ 11) & (Top-$k$ 66) & (Top-$k$ 42) & (Top-$k$ 65) & (Top-$k$ 67) & (Top-$k$ 26) \\
 & $\quad$  qd y & 6.79 & 7.91 & 7.42 & 7.13 & 7.07 & 4.98 & 1.50 & 1.41 & 7.89 \\
 &  & (Top-$p$ 0.59) & (Top-$k$ 26) & (Top-$k$ 89) & (Top-$k$ 56) & (Top-$k$ 67) & (Top-$p$ 0.97) & (Top-$p$ 1.0) & (Top-$p$ 0.81) & (Min-$p$ 0.16) \\
 & $\quad$  tpt fixed fee & 0.81 & 0.83 & 0.81 & 0.88 & 0.93 & 0.70 & 0.45 & 1.80 & 1.58 \\
 &  & (Top-$p$ 0.67) & (Top-$p$ 0.59) & (Top-$p$ 0.60) & (Top-$p$ 0.56) & (Top-$p$ 0.56) & (Top-$p$ 1.0) & (Top-$p$ 1.0) & (Top-$k$ 46) & (Top-$k$ 87) \\
 & $\quad$  tpt price a & 0.61 & 0.46 & 0.39 & 0.38 & 0.37 & 0.35 & 0.33 & 0.37 & 0.37 \\
 &  & (Top-$k$ 40) & (Top-$k$ 20) & (Top-$p$ 0.75) & (Top-$k$ 55) & (Top-$k$ 40) & (Top-$k$ 11) & (Top-$k$ 14) & (Top-$k$ 17) & (Min-$p$ 0.11) \\
6 & \cite{doi:10.1287/mnsc.2015.2264} &  &  &  &  &  &  &  &  &  \\
 & $\quad$  q00 & 0.22 & 0.19 & 0.16 & 0.15 & 0.14 & 0.12 & 0.12 & 0.12 & 0.10 \\
 &  & (Top-$k$ 10) & (Top-$k$ 29) & (Top-$k$ 44) & (Top-$k$ 67) & (Top-$k$ 10) & (Top-$k$ 43) & (Min-$p$ 0.05) & (Min-$p$ 0.11) & (Min-$p$ 0.05) \\
 & $\quad$  q50 & 0.14 & 0.15 & 0.14 & 0.15 & 0.14 & 0.15 & 0.14 & 0.14 & 0.14 \\
 &  & (Top-$p$ 0.59) & (Top-$k$ 27) & (Min-$p$ 0.11) & (Top-$k$ 65) & (Top-$k$ 65) & (Top-$p$ 1.0) & (Top-$p$ 0.50) & (Min-$p$ 0.16) & (Min-$p$ 0.05) \\
7 & \cite{doi:10.1287/mnsc.1110.1382} &  &  &  &  &  &  &  &  &  \\
 & $\quad$  c1 & 3.86 & 3.78 & 3.63 & 3.55 & 3.51 & 3.23 & 2.89 & 3.21 & 3.22 \\
 &  & (Top-$k$ 40) & (Top-$k$ 20) & (Top-$p$ 1.0) & (Top-$k$ 90) & (Top-$p$ 1.0) & (Top-$k$ 21) & (Top-$k$ 31) & (Min-$p$ 0.05) & (Min-$p$ 0.16) \\
 & $\quad$  c5 & 13.64 & 13.55 & 13.47 & 13.23 & 13.10 & 12.69 & 11.81 & 10.62 & 12.76 \\
 &  & (Top-$k$ 40) & (Top-$k$ 44) & (Min-$p$ 0.16) & (Top-$k$ 51) & (Top-$k$ 56) & (Top-$k$ 15) & (Top-$k$ 36) & (Top-$k$ 74) & (Min-$p$ 0.16) \\
8 & \cite{doi:10.1287/mnsc.46.3.404.12070} &  &  &  &  &  &  &  &  &  \\
 & $\quad$  Low Profit & 33.93 & 29.68 & 22.75 & 28.80 & 23.23 & 22.71 & 22.06 & 19.31 & 20.83 \\
 &  & (Top-$k$ 40) & (Top-$p$ 0.52) & (Top-$k$ 45) & (Top-$k$ 19) & (Top-$k$ 16) & (Top-$k$ 10) & (Top-$k$ 67) & (Top-$k$ 16) & (Top-$k$ 10) \\
 & $\quad$  High Profit & 44.90 & 45.33 & 45.51 & 44.71 & 43.49 & 43.37 & 31.97 & 39.11 & 37.85 \\
 &  & (Top-$k$ 32) & (Top-$k$ 64) & (Top-$k$ 14) & (Top-$k$ 13) & (Top-$k$ 67) & (Top-$k$ 14) & (Top-$k$ 26) & (Top-$k$ 45) & (Min-$p$ 0.16) \\
9 & \cite{doi:10.1287/mnsc.1110.1334} &  &  &  &  &  &  &  &  &  \\
 & $\quad$  signal & 32.32 & 31.03 & 28.58 & 29.04 & 26.63 & 24.56 & 24.34 & 23.22 & 20.87 \\
 &  & (Top-$k$ 32) & (Top-$k$ 84) & (Top-$k$ 32) & (Top-$k$ 55) & (Top-$k$ 32) & (Top-$k$ 33) & (Top-$k$ 40) & (Min-$p$ 0.16) & (Min-$p$ 0.11) \\
 & $\quad$  capacity & 31.86 & 35.57 & 36.36 & 36.78 & 35.19 & 40.86 & 49.68 & 49.82 & 55.58 \\
 &  & (Top-$k$ 77) & (Top-$p$ 0.65) & (Min-$p$ 0.05) & (Top-$p$ 0.65) & (Top-$p$ 0.72) & (Top-$p$ 0.50) & (Min-$p$ 0.16) & (Min-$p$ 0.16) & (Min-$p$ 0.16) \\
    \bottomrule
    \end{tabular}
    }
    }
    \end{table}

\clearpage

\clearpage
\subsection{LLaMA-70B Word Output Distribution}
\begin{figure}[htbp!] 
\vspace{-3mm}
\centering
\caption{LLaMA-70B Word Probability Distribution: Replication Results of Conditions from \cite{doi:10.1287/mnsc.1120.1638}, \cite{doi:10.1287/mnsc.1050.0436},
\cite{doi:10.1287/mnsc.1100.1258} and  \cite{doi:10.1287/mnsc.1070.0806}}
\label{fig:ec2:LLaMA70b_distribution}
 \begin{subfigure}[b]{0.5\textwidth}
         \centering
         \includegraphics[width=0.95\textwidth, height=0.3\textwidth]{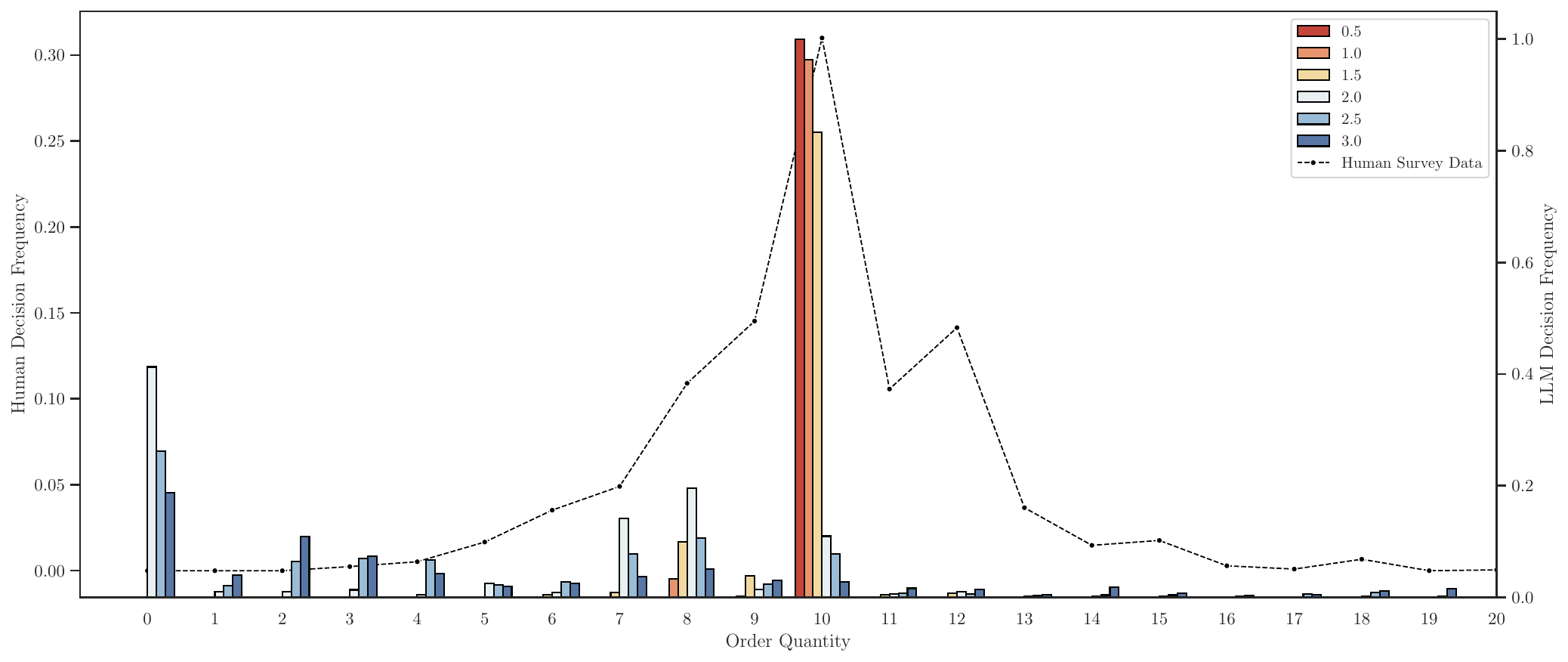}
         \vspace{-3mm}
         \caption{\cite{doi:10.1287/mnsc.1120.1638} -- C}
     \end{subfigure}
     \hspace{-5mm}
     \begin{subfigure}[b]{0.5\textwidth}
         \centering
         \includegraphics[width=0.95\textwidth, height=0.3\textwidth]{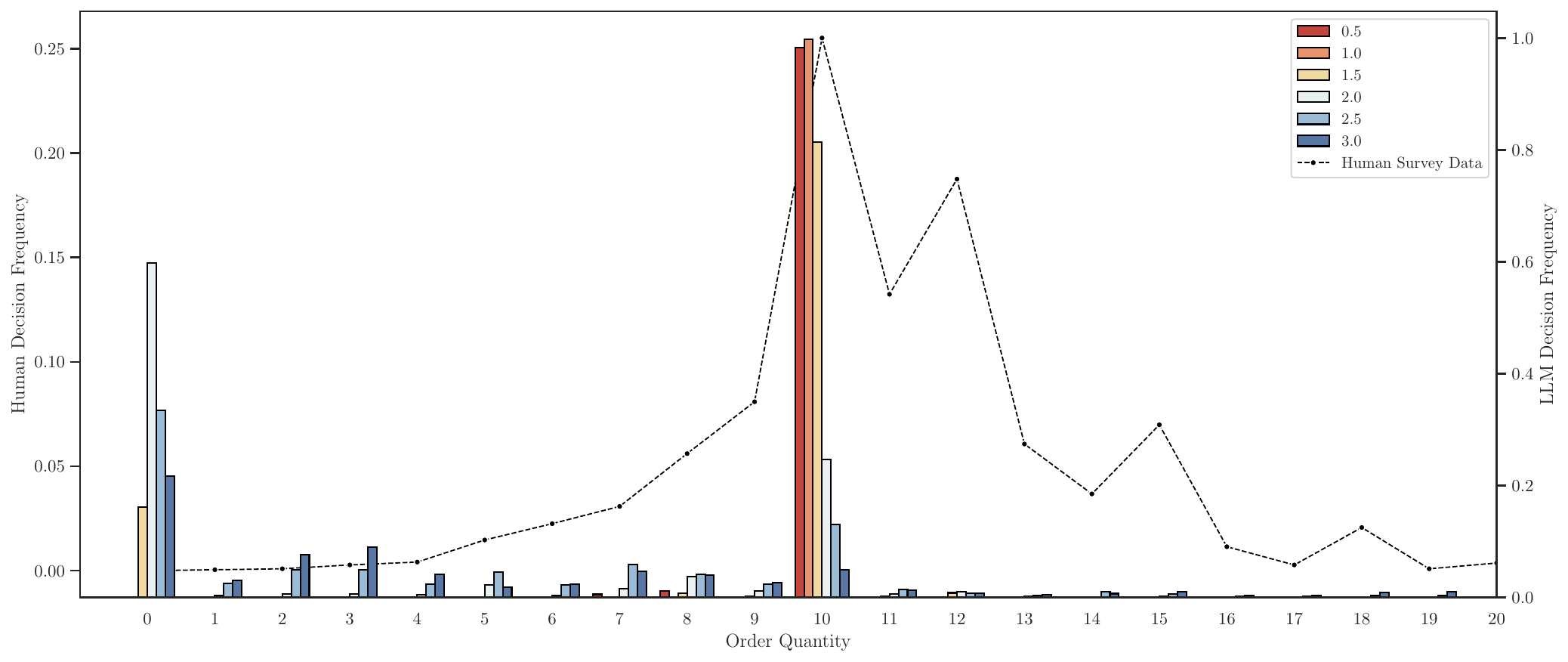}
         \vspace{-3mm}
         \caption{\cite{doi:10.1287/mnsc.1120.1638} -- O}
     \end{subfigure} 
     \vspace{-1mm}
 \begin{subfigure}[b]{0.5\textwidth}
         \centering
         \includegraphics[width=0.95\textwidth, height=0.29\textwidth]{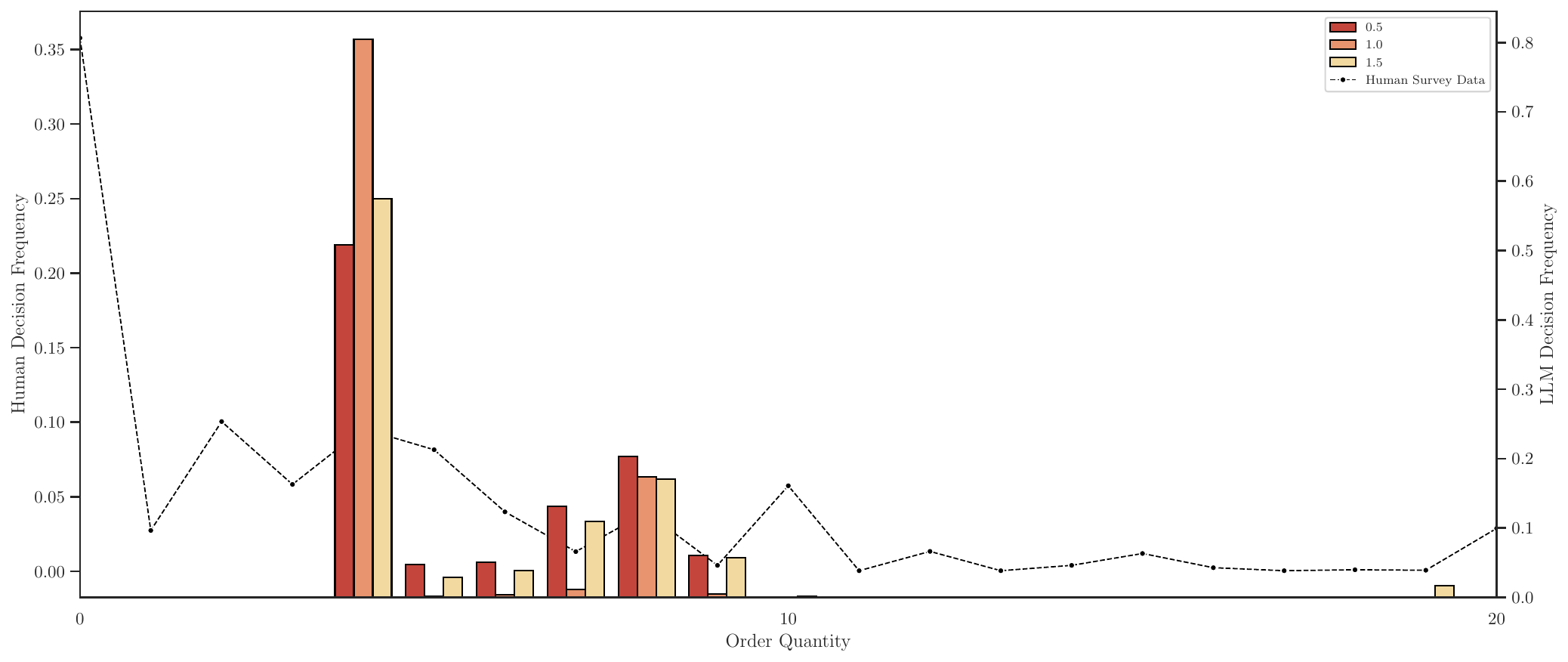}
         \vspace{-3mm}
         \caption{\cite{doi:10.1287/mnsc.1050.0436} -- Base}
     \end{subfigure}
     \hspace{-5mm}
     \vspace{-1mm}
     \begin{subfigure}[b]{0.5\textwidth}
         \centering
         \includegraphics[width=0.95\textwidth, height=0.29\textwidth]{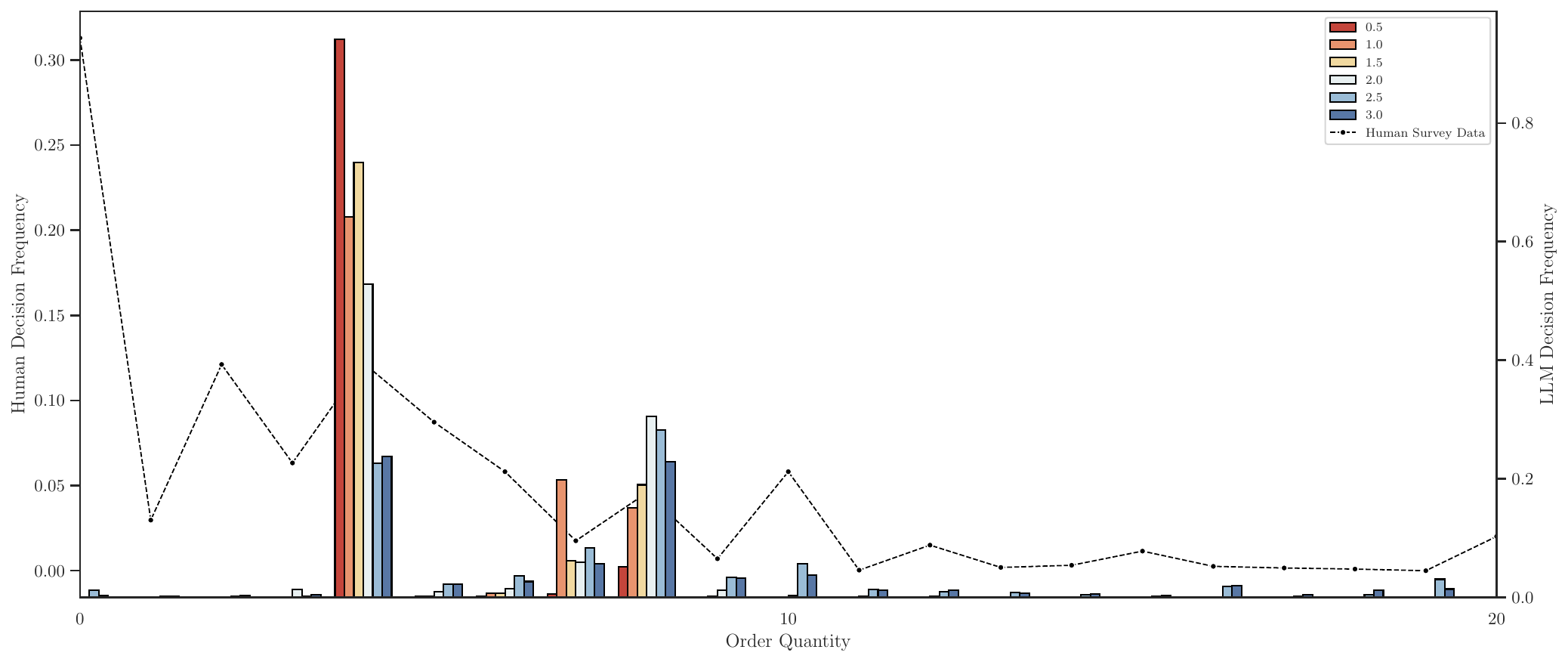}
         \vspace{-3mm}
         \caption{\cite{doi:10.1287/mnsc.1050.0436} -- Share}
     \end{subfigure}
     \vspace{-1mm}
 \begin{subfigure}[b]{0.5\textwidth}
         \centering
         \includegraphics[width=0.95\textwidth, height=0.300\textwidth]{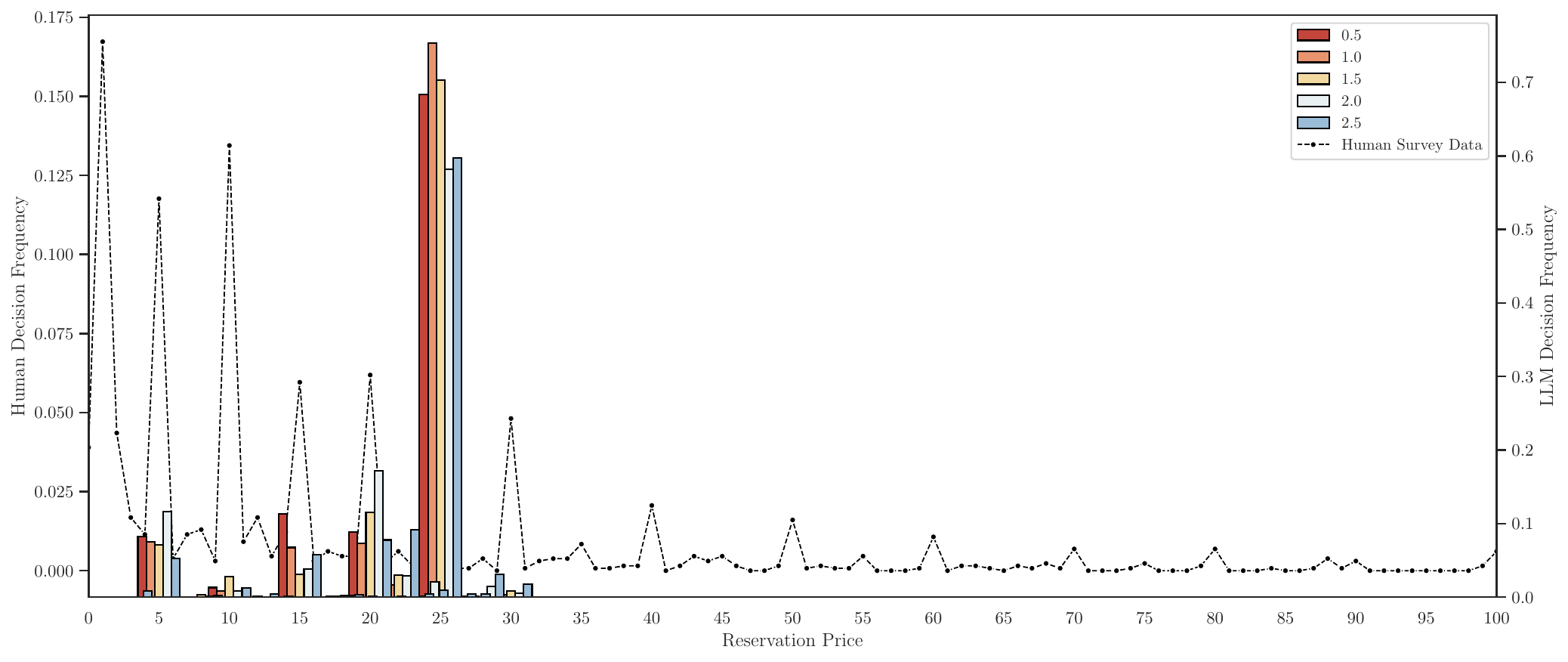}
         \vspace{-3mm}
         \caption{\cite{doi:10.1287/mnsc.1100.1258} -- 1}
     \end{subfigure}
     \hspace{-5mm}
     \vspace{-1mm}
     \begin{subfigure}[b]{0.5\textwidth}
         \centering
         \includegraphics[width=0.95\textwidth, height=0.300\textwidth]{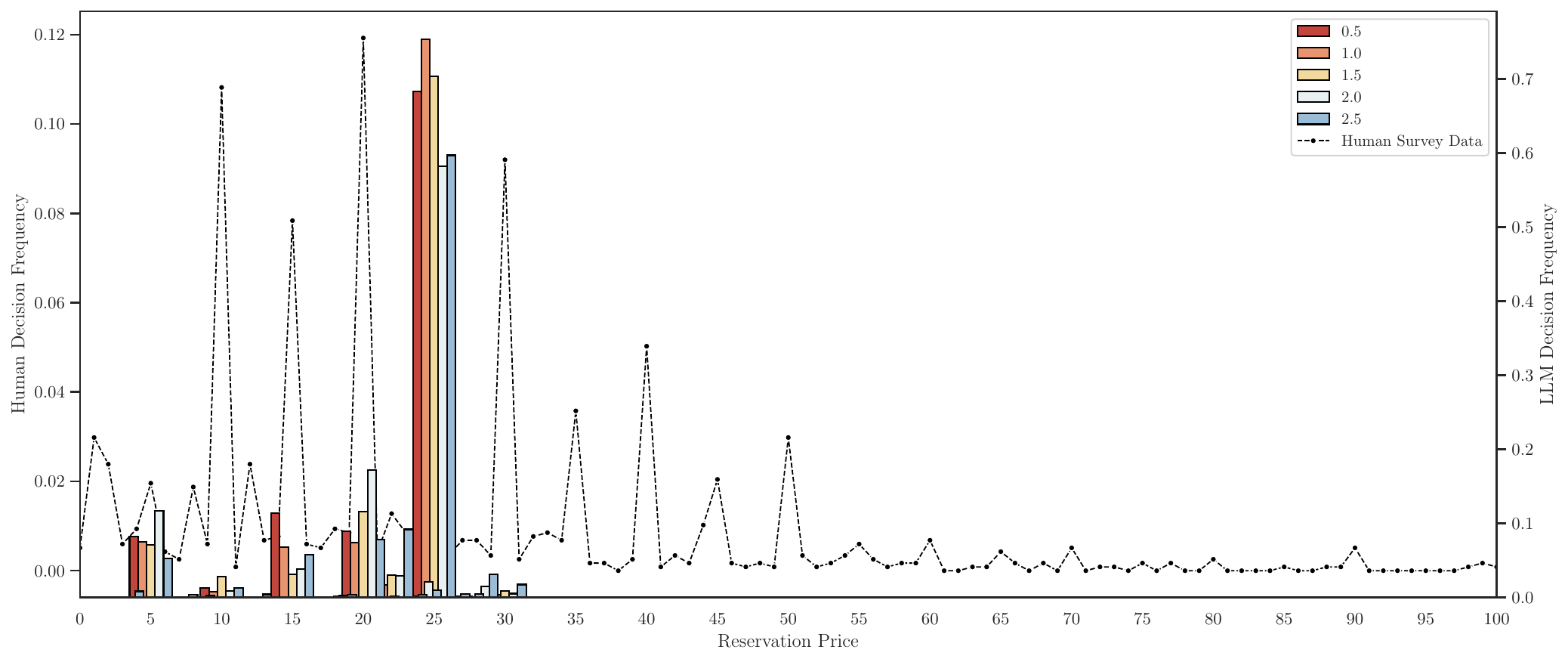}
         \vspace{-3mm}
         \caption{\cite{doi:10.1287/mnsc.1100.1258} -- 4}
     \end{subfigure}
     \vspace{-1mm}
     \begin{subfigure}[b]{0.5\textwidth}
         \centering
         \includegraphics[width=0.95\textwidth, height=0.300\textwidth]{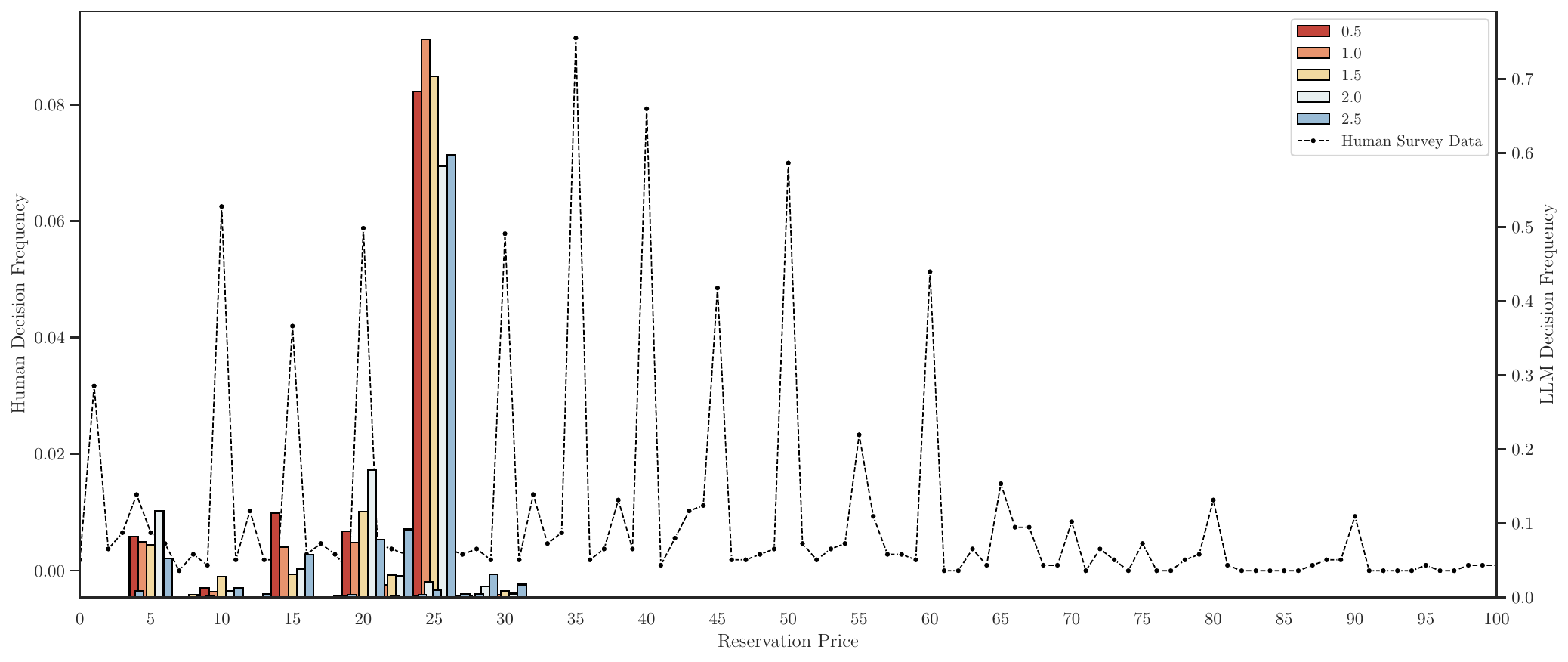}
         \vspace{-3mm}
         \caption{\cite{doi:10.1287/mnsc.1100.1258} -- 7}
     \end{subfigure}
     \hspace{-5mm}
     \vspace{-1mm}
     \begin{subfigure}[b]{0.5\textwidth}
         \centering
         \includegraphics[width=0.95\textwidth, height=0.300\textwidth]{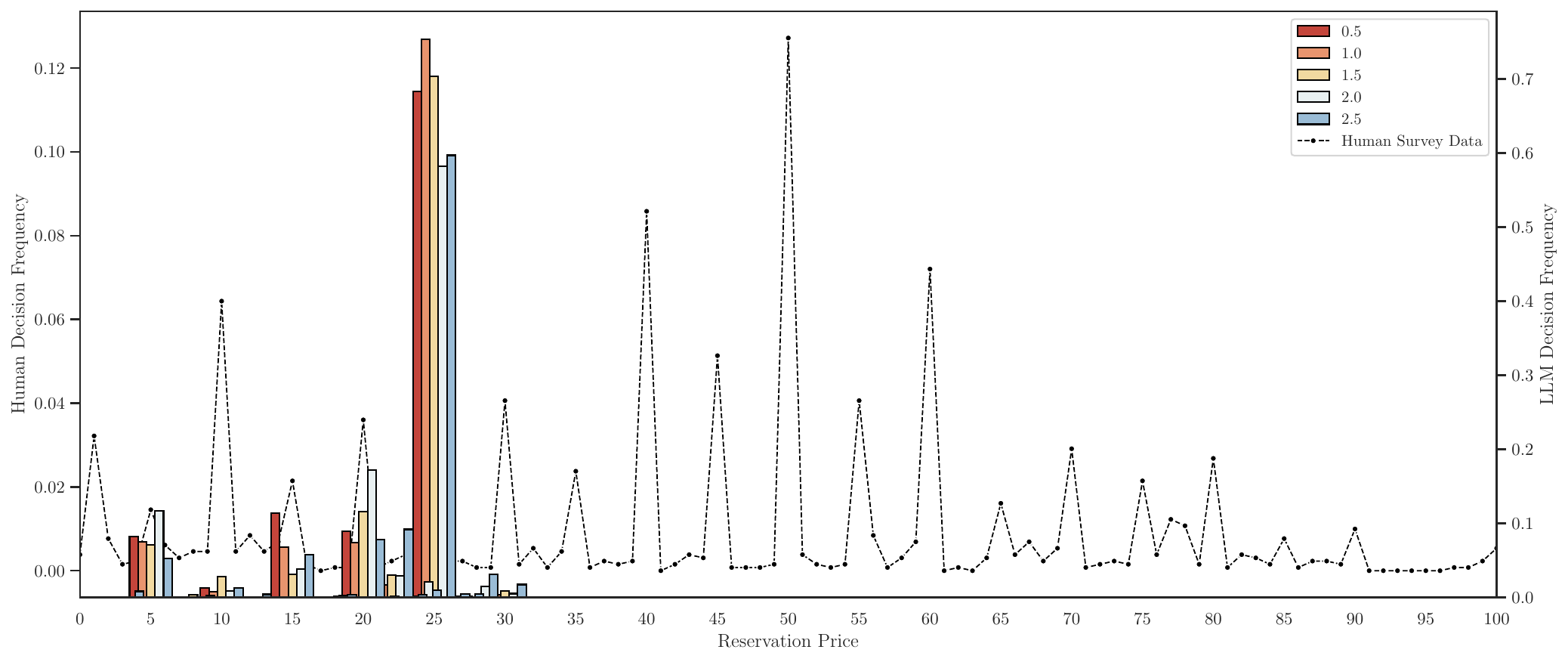}
         \vspace{-3mm}
         \caption{\cite{doi:10.1287/mnsc.1100.1258} -- 10}
     \end{subfigure}
     \vspace{-1mm}
     \begin{subfigure}[b]{0.5\textwidth}
         \centering
         \includegraphics[width=0.95\textwidth, height=0.300\textwidth]{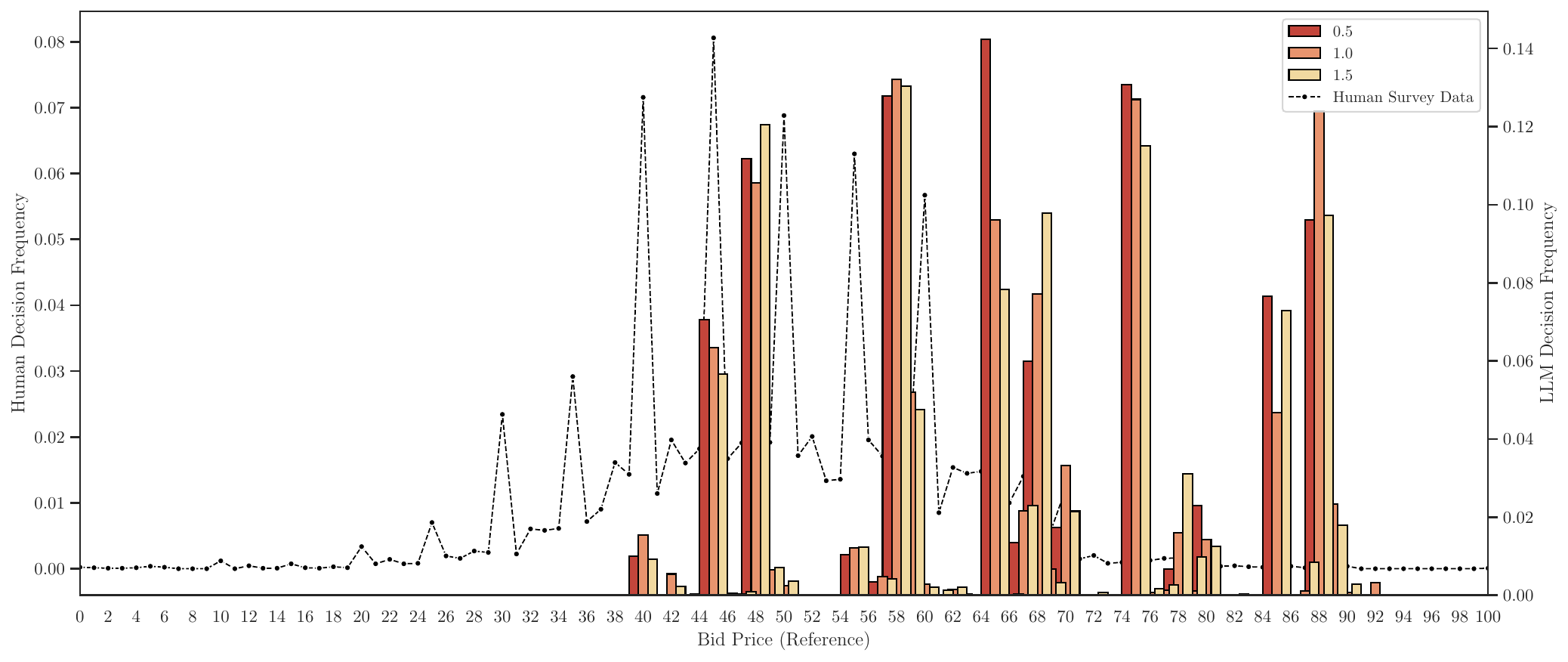}
         \vspace{-3mm}
         \caption{\cite{doi:10.1287/mnsc.1070.0806} \\ -- Both}
     \end{subfigure}
     \hspace{-5mm}
     \vspace{-1mm}
     \begin{subfigure}[b]{0.5\textwidth}
         \centering
         \includegraphics[width=0.95\textwidth, height=0.300\textwidth]{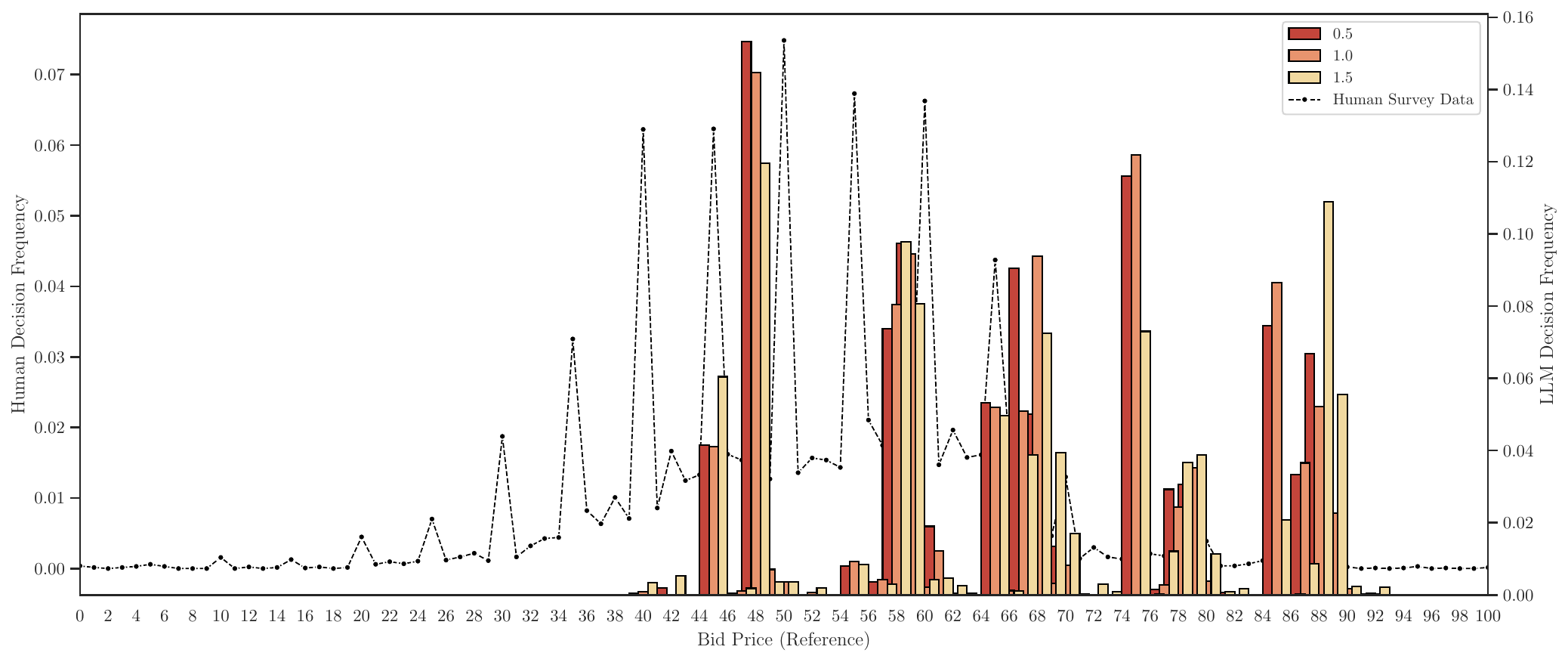}
         \vspace{-3mm}
         \caption{\cite{doi:10.1287/mnsc.1070.0806} \\ -- Loser}
     \end{subfigure}
    \end{figure}

\begin{figure}[htbp!]

  \centering
\caption{LLaMA-70B Word Probability Distribution: Replication Results of Conditions from \cite{doi:10.1287/mnsc.1070.0788}, \cite{doi:10.1287/mnsc.2015.2264}, \cite{doi:10.1287/mnsc.1110.1382}, \cite{doi:10.1287/mnsc.46.3.404.12070} and \cite{doi:10.1287/mnsc.1110.1334}}
\label{fig:ec3:LLaMA70b_distribution}

\begin{subfigure}[b]{0.5\textwidth}
         \centering
         \includegraphics[width=0.95\textwidth, height=0.300\textwidth]{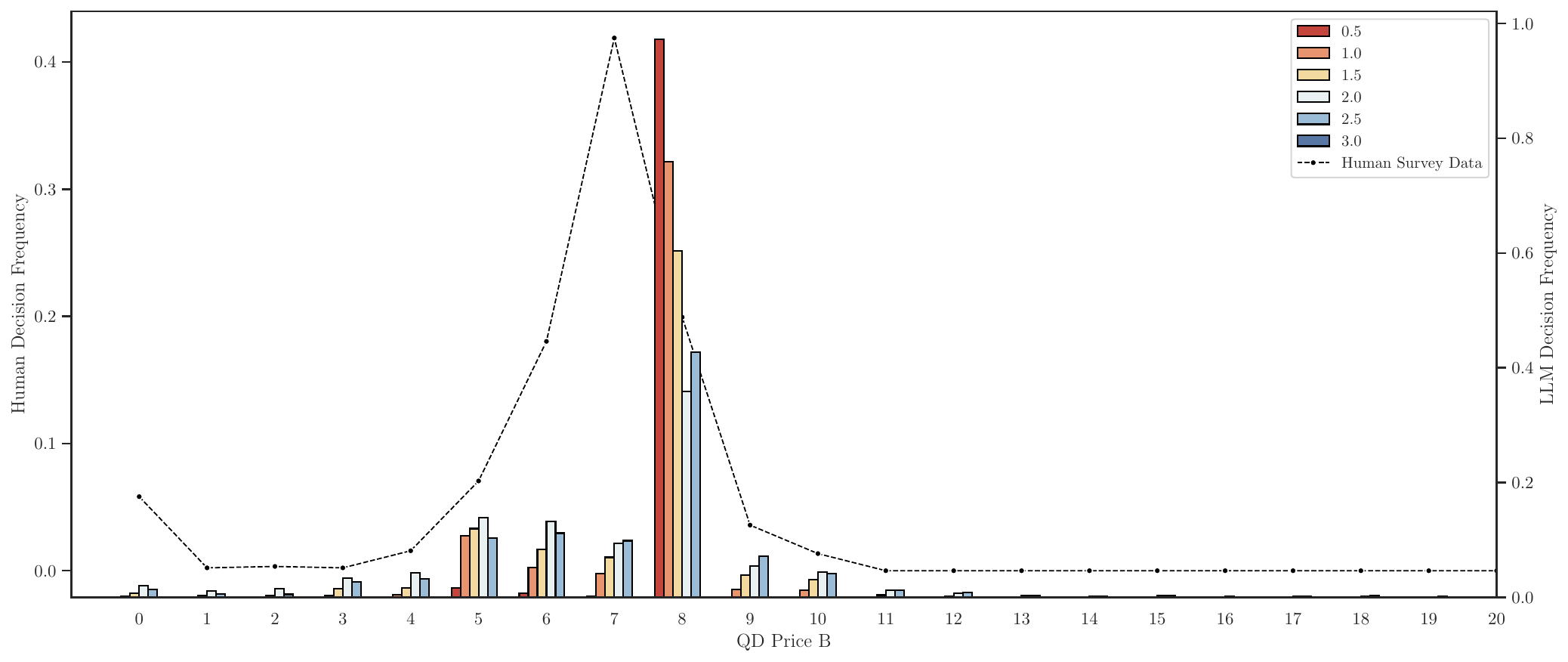}
         \vspace{-3mm}
         \caption{\cite{doi:10.1287/mnsc.1070.0788} -- QD price B}
     \end{subfigure}
     \hspace{-5mm}
     \begin{subfigure}[b]{0.5\textwidth}
         \centering
         \includegraphics[width=0.95\textwidth, height=0.300\textwidth]{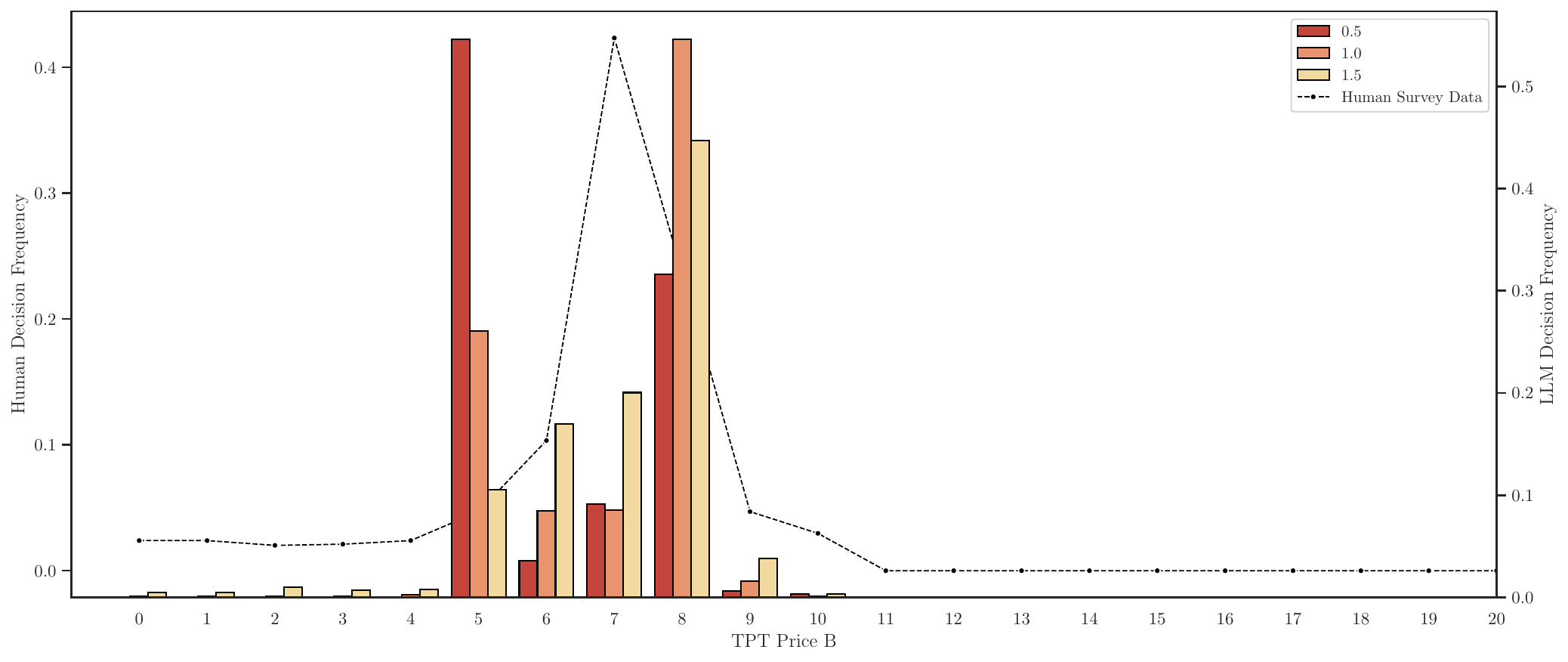}
         \vspace{-3mm}
         \caption{\cite{doi:10.1287/mnsc.1070.0788} -- TPT price B}
     \end{subfigure}
     \vspace{-1mm}
    \begin{subfigure}[b]{0.5\textwidth}
         \centering
         \includegraphics[width=0.95\textwidth, height=0.300\textwidth]{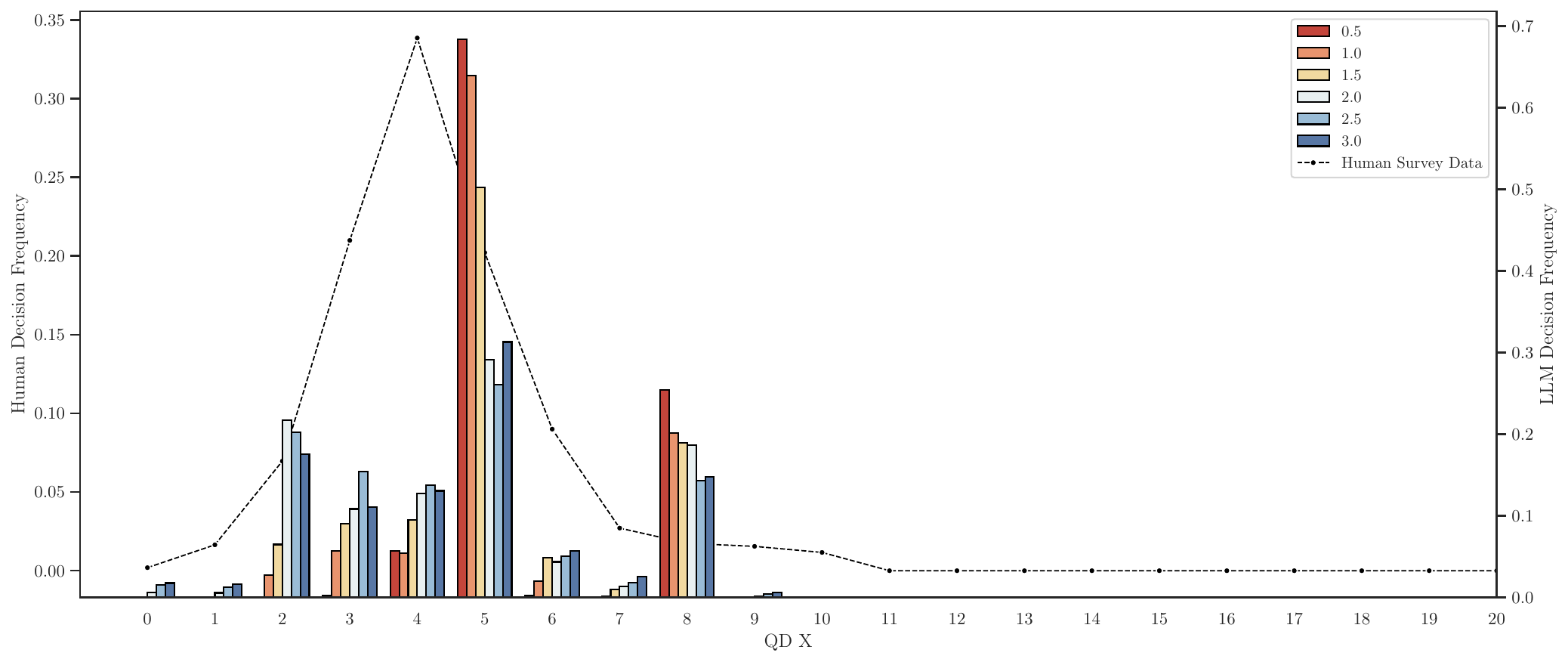}
         \vspace{-3mm}
         \caption{\cite{doi:10.1287/mnsc.1070.0788} -- QD x}
     \end{subfigure}
     \hspace{-5mm}
     \vspace{-1mm}
    \begin{subfigure}[b]{0.5\textwidth}
         \centering
         \includegraphics[width=0.95\textwidth, height=0.300\textwidth]{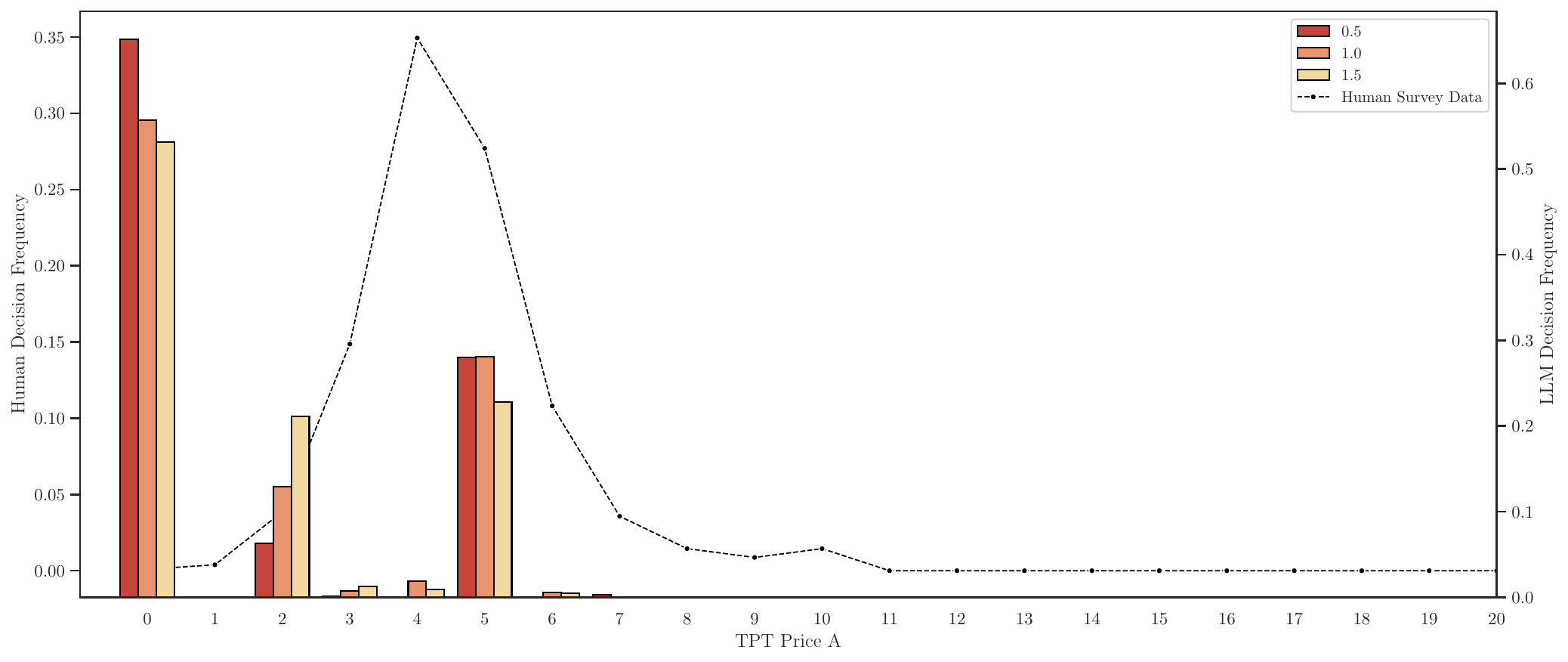}
         \vspace{-3mm}
         \caption{\cite{doi:10.1287/mnsc.1070.0788} -- TPT price A}
     \end{subfigure}
     \vspace{-1mm}
     \begin{subfigure}[b]{0.5\textwidth}
         \centering
         \includegraphics[width=0.95\textwidth, height=0.300\textwidth]{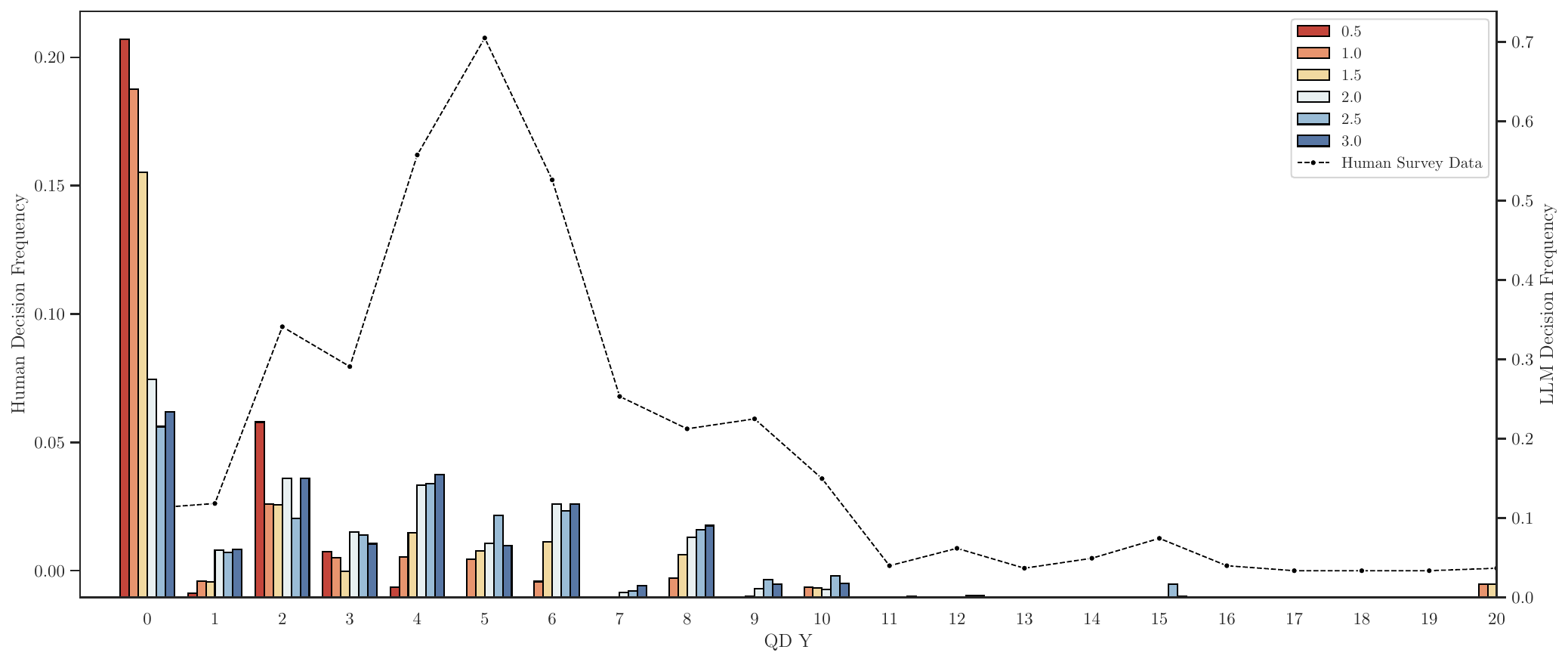}
         \vspace{-3mm}
         \caption{\cite{doi:10.1287/mnsc.1070.0788} -- QD y}
     \end{subfigure}
     \hspace{-5mm}
     \vspace{-1mm}
     \begin{subfigure}[b]{0.5\textwidth}
         \centering
         \includegraphics[width=0.95\textwidth, height=0.300\textwidth]{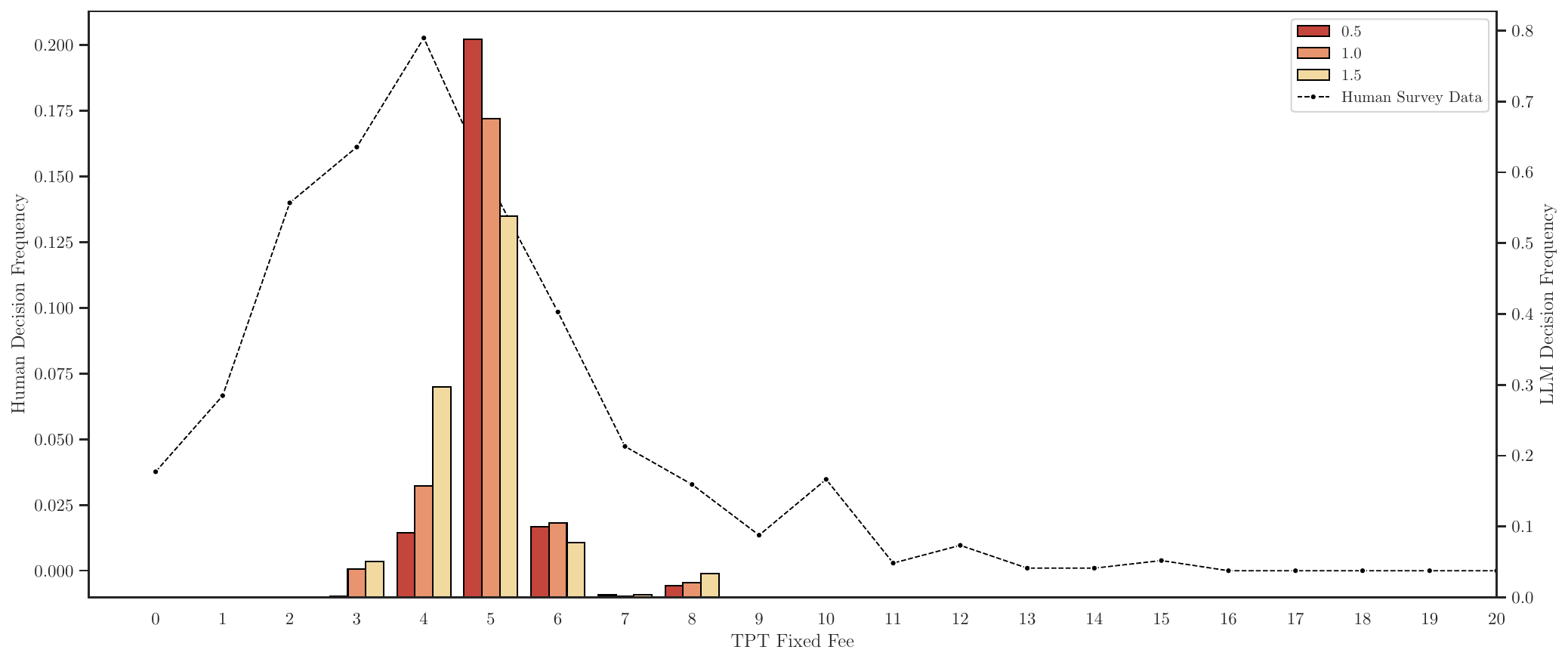}
         \vspace{-3mm}
         \caption{\cite{doi:10.1287/mnsc.1070.0788} -- TPT fixed fee}
     \end{subfigure}

\vspace{-1mm}
 \begin{subfigure}[b]{0.5\textwidth}
         \centering
         \includegraphics[width=0.95\textwidth, height=0.300\textwidth]{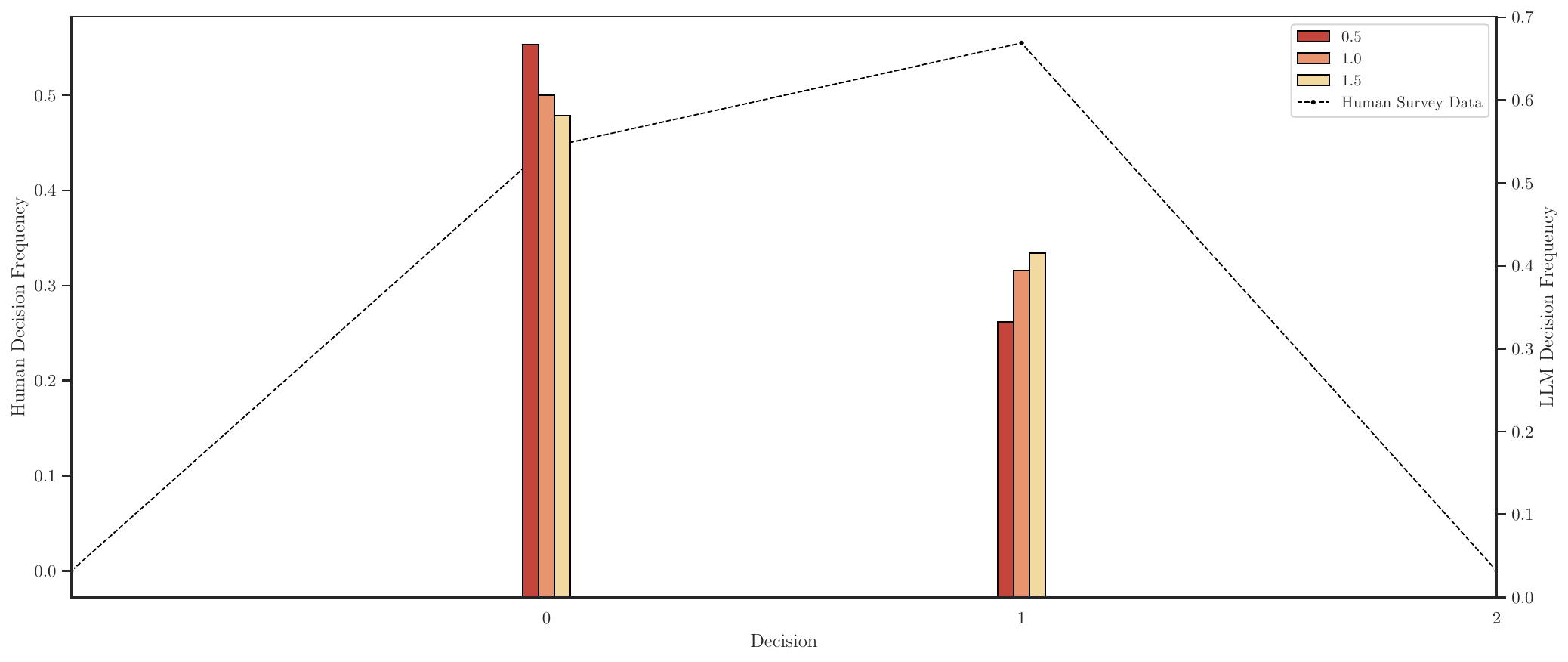}
         \vspace{-3mm}
         \caption{\cite{doi:10.1287/mnsc.2015.2264} -- $q00$}
     \end{subfigure}
     \hspace{-5mm}
     \vspace{-1mm}
     \begin{subfigure}[b]{0.5\textwidth}
         \centering
         \includegraphics[width=0.95\textwidth, height=0.300\textwidth]{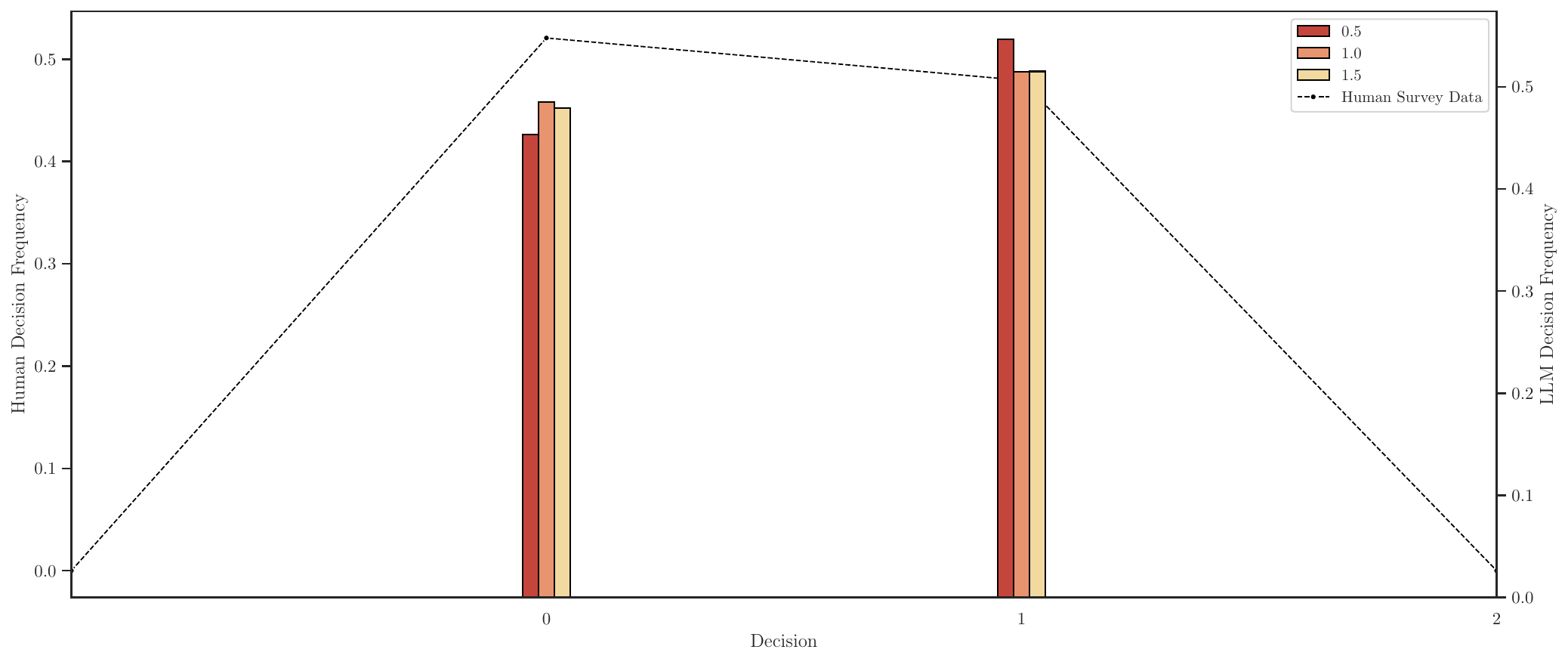}
         \vspace{-3mm}
         \caption{\cite{doi:10.1287/mnsc.2015.2264} -- $q50$}
     \end{subfigure}

\vspace{-1mm}
     \begin{subfigure}[b]{0.5\textwidth}
         \centering
         \includegraphics[width=0.95\textwidth, height=0.300\textwidth]{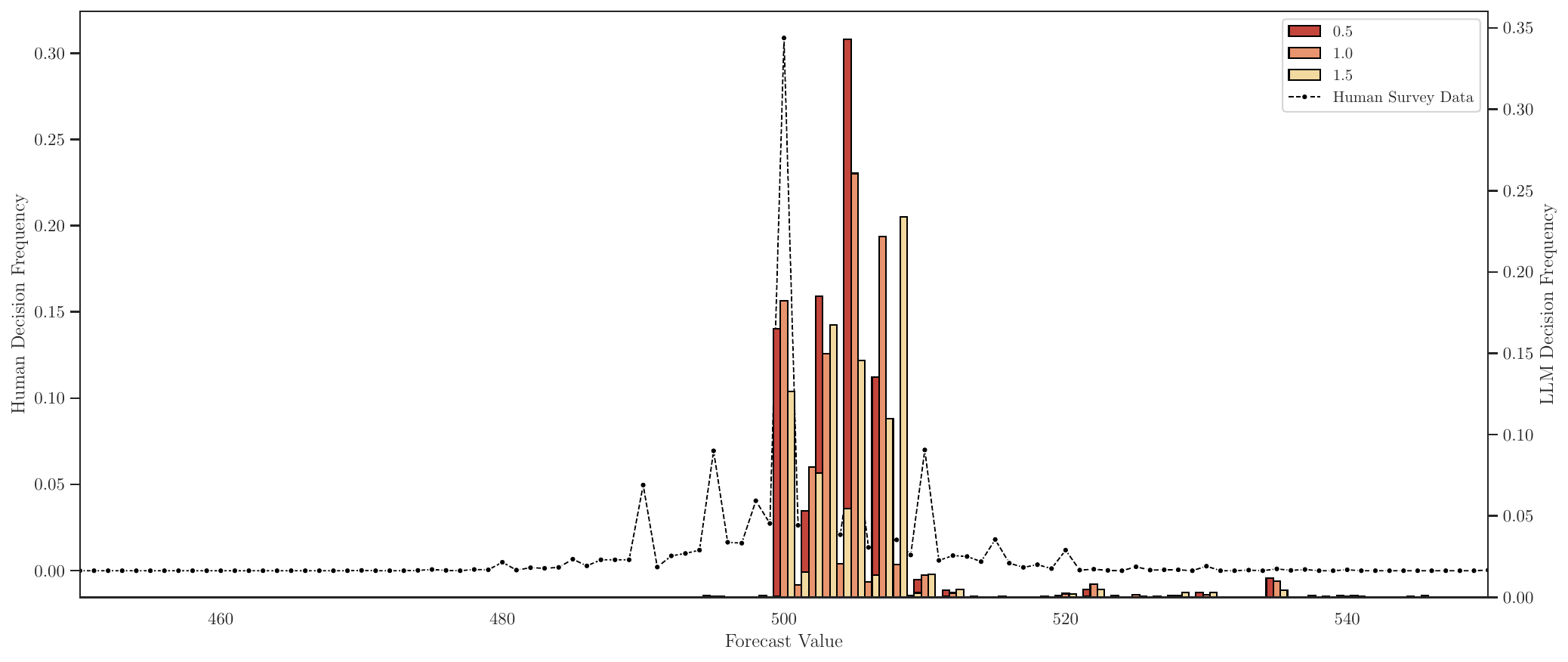}
         \vspace{-3mm}
         \caption{\cite{doi:10.1287/mnsc.1110.1382} -- c1}
     \end{subfigure}
     \hspace{-5mm}
     \vspace{-1mm}
     \begin{subfigure}[b]{0.5\textwidth}
         \centering
         \includegraphics[width=0.95\textwidth, height=0.300\textwidth]{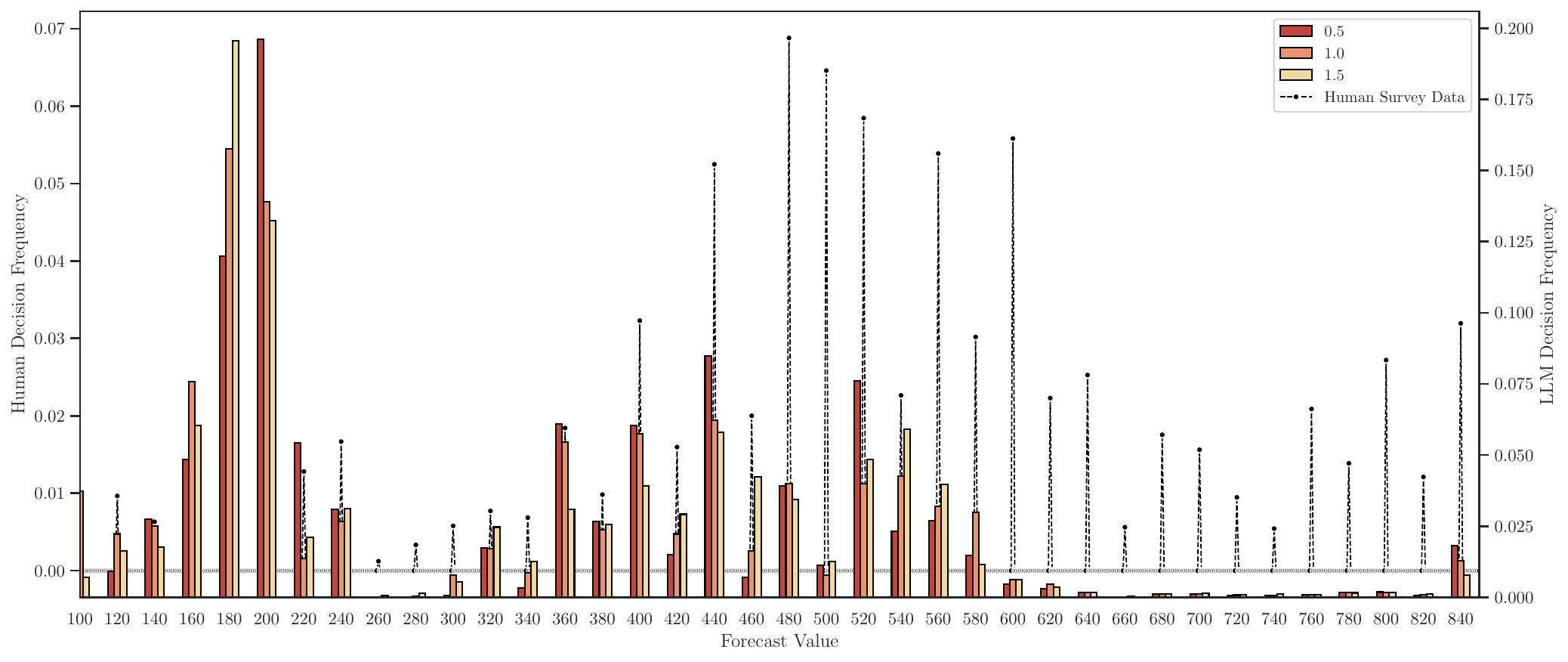}
         \vspace{-3mm}
         \caption{\cite{doi:10.1287/mnsc.1110.1382} -- c5}
     \end{subfigure}

\vspace{-1mm}
\begin{subfigure}[b]{0.5\textwidth}
         \centering
         \includegraphics[width=0.95\textwidth, height=0.300\textwidth]{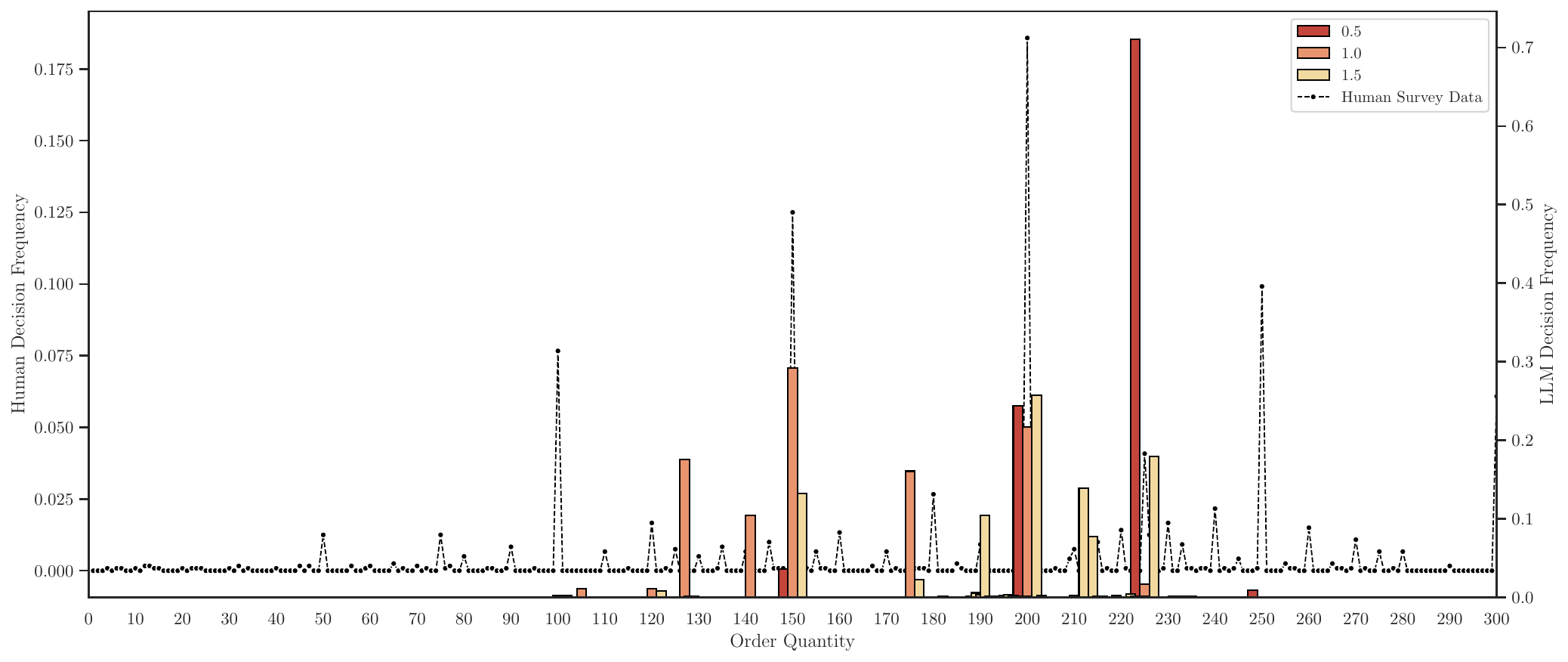}
         \vspace{-3mm}
         \caption{\cite{doi:10.1287/mnsc.46.3.404.12070} 
         \vspace{-2mm}
         \\ -- High Profit}
     \end{subfigure}
     \hspace{-5mm}
     \vspace{-1mm}
     \begin{subfigure}[b]{0.5\textwidth}
         \centering
         \includegraphics[width=0.95\textwidth, height=0.300\textwidth]{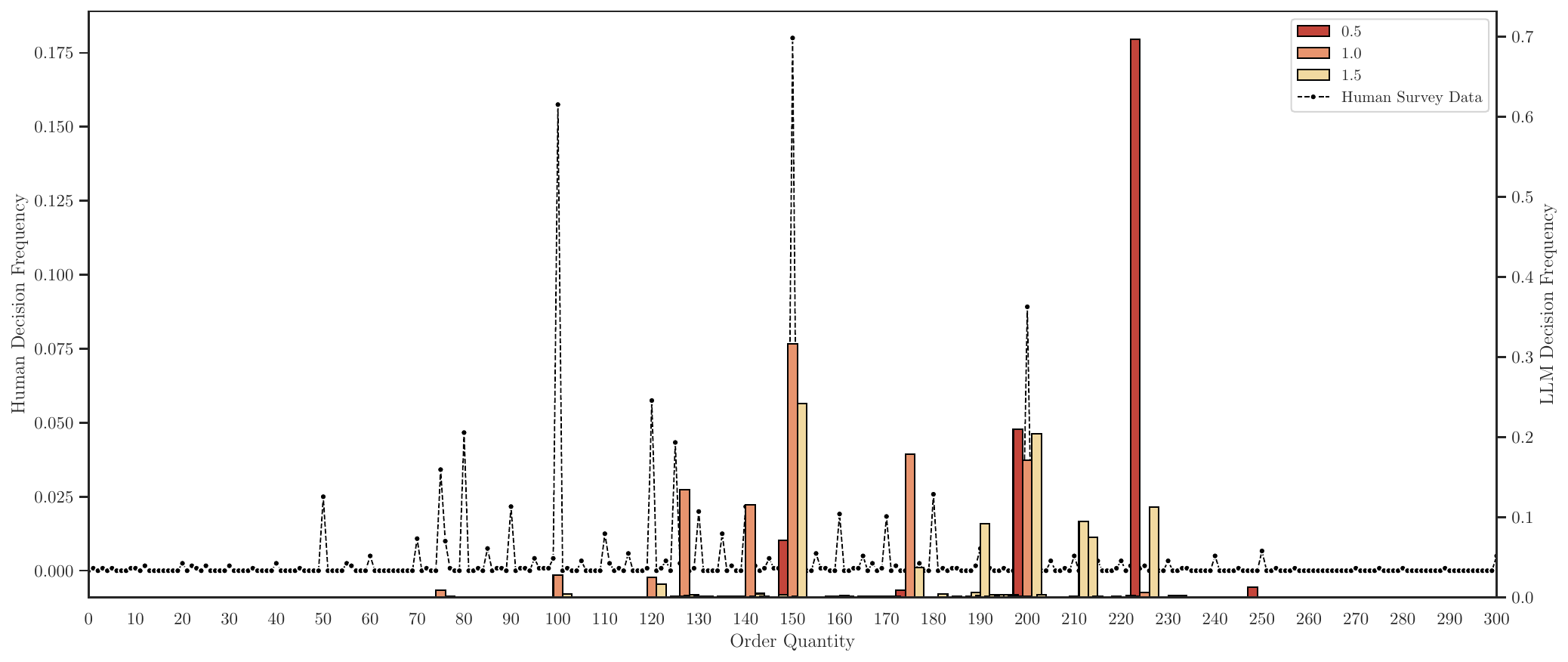}
         \vspace{-3mm}
         \caption{\cite{doi:10.1287/mnsc.46.3.404.12070}
         \vspace{-2mm}
         \\ -- Low Profit}
     \end{subfigure}

\vspace{-1mm}
     \begin{subfigure}[b]{0.5\textwidth}
         \centering
         \includegraphics[width=0.95\textwidth, height=0.300\textwidth]{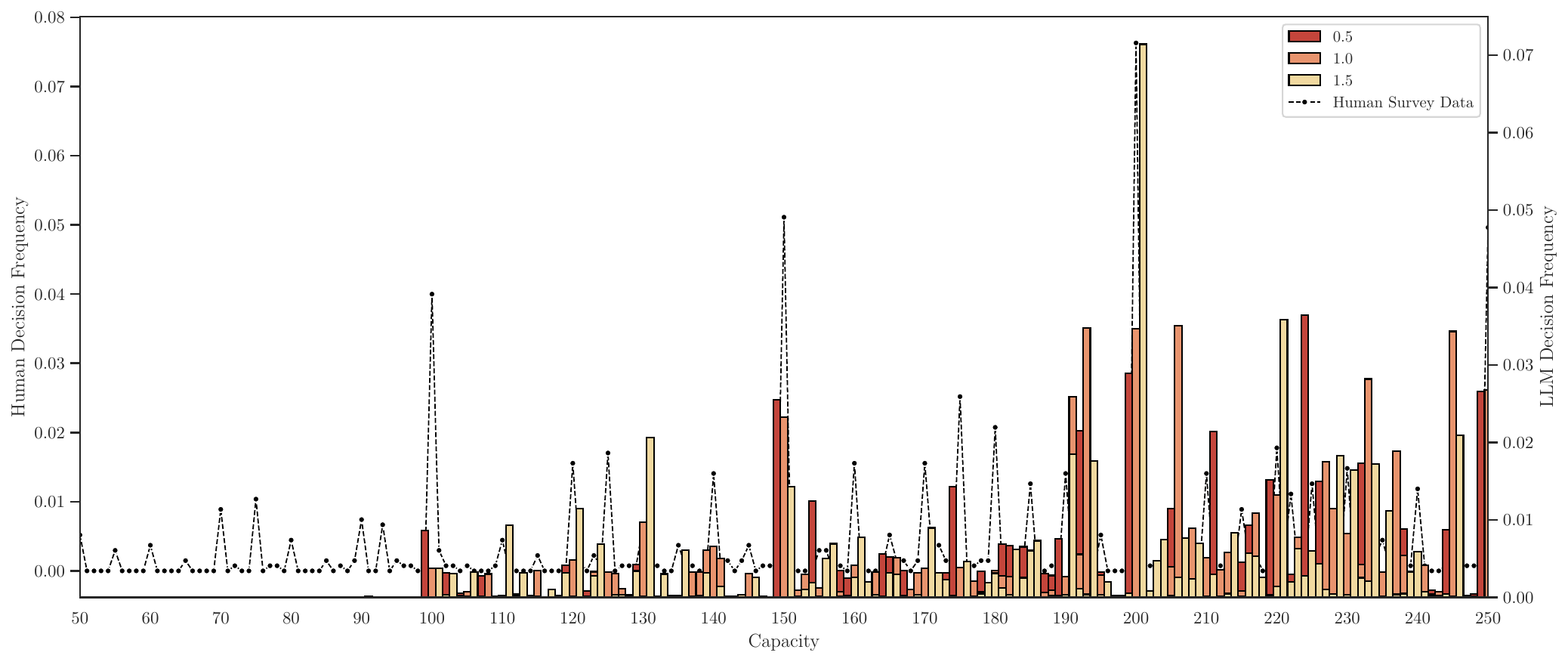}
         \vspace{-3mm}
         \caption{\cite{doi:10.1287/mnsc.1110.1334} -- Capacity}
     \end{subfigure}
     \hspace{-5mm}
     \vspace{-1mm}
     \begin{subfigure}[b]{0.5\textwidth}
         \centering
         \includegraphics[width=0.95\textwidth, height=0.300\textwidth]{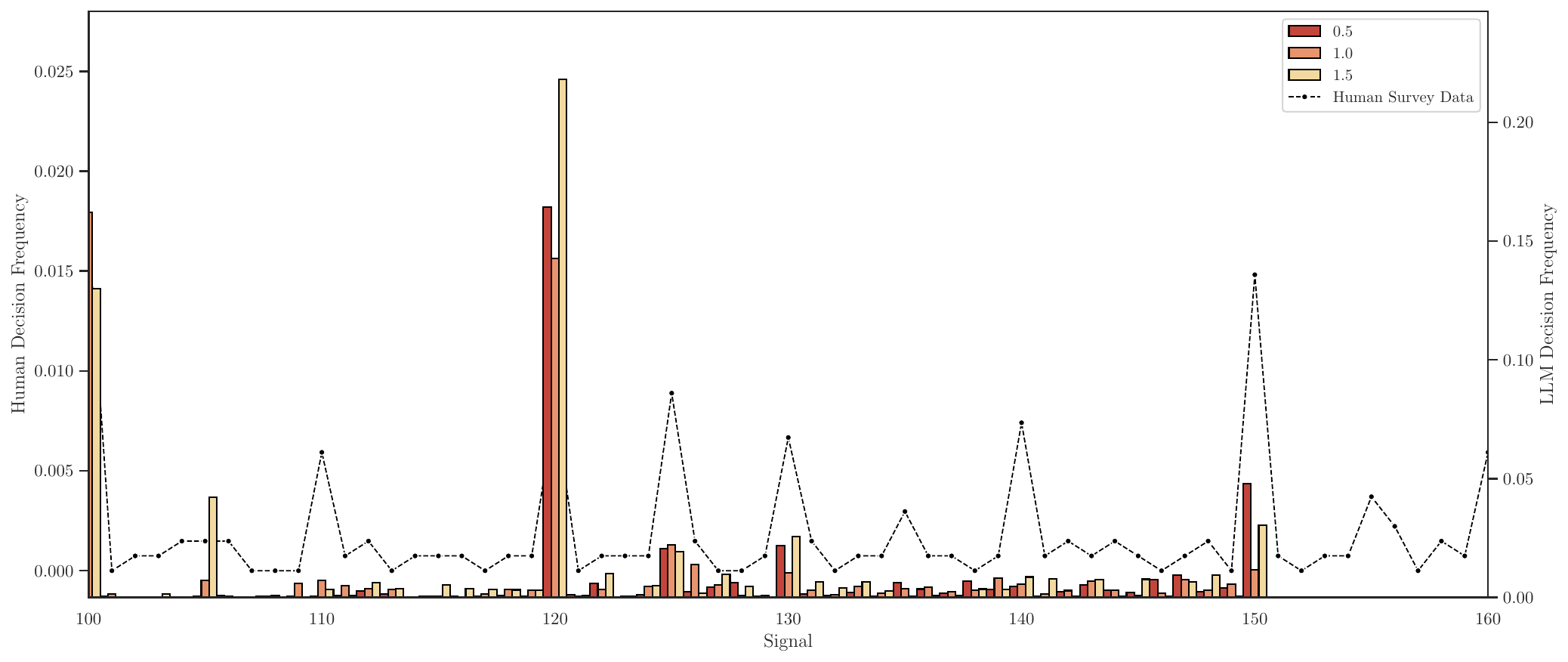}
         \vspace{-3mm}
         \caption{\cite{doi:10.1287/mnsc.1110.1334} -- Signal}

     \end{subfigure}

\end{figure}

We employ the CoT prompting of LLaMA-70B; accordingly, Figures \ref{fig:ec2:LLaMA70b_distribution} and \ref{fig:ec3:LLaMA70b_distribution} depict the distributions of human and LLM decision frequencies across six temperature settings in nine BOM experiments. These results primarily highlight the distributional differences between human- and LLM-generated data. While the outputs of LLMs tend to concentrate around a limited set of values, human responses exhibit a more dispersed pattern. Furthermore, we observe that the output distribution varies with temperature, which motivates our quantitative analysis using the Wasserstein distance and subsequent calibration through temperature tuning. For some experiments (e.g., \cite{doi:10.1287/mnsc.2015.2264} in \ref{fig:ec2:LLaMA70b_distribution}), we only report distributions under three temperature settings (0.5, 1.0, and 1.5) due to inappropriate outputs in the remaining settings within the three available attempts.

\end{document}